\RequirePackage{etoolbox}
\csdef{input@path}{%
 {sty/}
 {img/}
}%

\documentclass[preprint]{imsart} 

\usepackage[mathscr]{euscript}
\usepackage{amsthm}
\usepackage{amsmath}
\usepackage{amsfonts}       
\usepackage{natbib}
\usepackage{minibox}
\usepackage[colorlinks,citecolor=blue,urlcolor=blue,filecolor=blue,backref=page]{hyperref}
\usepackage{graphicx}
\usepackage{bm}

\usepackage{algorithm}
\usepackage[noend]{algpseudocode}
\usepackage{todonotes}
\usepackage{amssymb}

\usepackage[font=small,labelfont=bf]{caption}
\usepackage[font=footnotesize]{subcaption}
\usepackage{url}            

\usepackage{comment}
\usepackage{pstricks}
\usepackage{graphicx}
\usepackage{pstricks, pst-node}

\usepackage{tikz}
\usepackage{tikz-cd}
\usepackage{tkz-graph}
\usetikzlibrary{fit,positioning,bayesnet}

\startlocaldefs

\usepackage[page,title]{appendix}

\let\stdappendixpage\appendixpage
\renewcommand*\appendixpage{{%
   \let\markboth\relax\stdappendixpage}}

\numberwithin{equation}{section}
\theoremstyle{plain}

\newtheorem{rem}{Remark}[section]
\newtheorem{ex}{Example}[section]

\newcommand{\pa}[1]{{\mathop{\textup{pa}{(#1)}}}}
\newcommand{\nfa}[1]{{\mathop{\neg{\textup{fa}}{(#1)}}}}
\newcommand{\fa}[1]{{\mathop{\textup{fa}{(#1)}}}}
\newcommand{\vi}{V}
\newcommand{\vibar}{{\bar{V}}}
\newcommand{\mycolon}{:}
\newcommand{\Polya}{P{\'o}lya~}
\newcommand{\PolyaBayes}{P{\'o}lya-Bayes~}
\newcommand{\SMC}{SMC~}
\newcommand{\E}[2][]{\mathsf{E}_{#1}\left\{{#2}\right\}}

\newcommand{\ind}[1]{\mathbb{I}\left\{{#1}\right\}}
\newcommand{\Cliques}{\mathcal{C}_{\mathcal{G}}}
\newcommand{\Separators}{\mathcal{S}_{\mathcal{G}}}

\usepackage{cancel}
\usepackage{array}
\usepackage{tikz}
\usetikzlibrary{
  shapes.multipart,
  matrix,
  positioning,
  shapes.callouts,
  shapes.arrows,
  calc}

\newcommand{\DrawCube}[6]{%
\pgfmathsetmacro{\cubex}{#2}
\pgfmathsetmacro{\cubey}{#1}
\pgfmathsetmacro{\cubez}{#3}
\pgfmathsetmacro{\xo}{#5}
\pgfmathsetmacro{\yo}{-#4}
\pgfmathsetmacro{\zo}{#6}
\draw[black,fill=white] (\xo,\yo,\zo) -- ++(\cubex,0,0) -- ++(0,-\cubey,0) -- ++(-\cubex,0,0) -- cycle;
\draw[black,fill=white] (\xo+\cubex,\yo,\zo) -- ++(0,0,-\cubez) -- ++(0,-\cubey,0) -- ++(0,0,\cubez) -- cycle;
\draw[black,fill=white] (\xo,\yo,\zo) -- ++(\cubex,0,0) -- ++(0,0,-\cubez) -- ++(-\cubex,0,0) -- cycle;
}

\newcommand{\DrawCubeLabels}[9]{%
\pgfmathsetmacro{\cubex}{#2}
\pgfmathsetmacro{\cubey}{#1}
\pgfmathsetmacro{\cubez}{#3}
\pgfmathsetmacro{\xo}{#5}
\pgfmathsetmacro{\yo}{-#4}
\pgfmathsetmacro{\zo}{#6}
\draw[thick,black,fill=white,->] (\xo,\yo,\zo) -- ++(\cubex,0,0);
\node [below] at (\xo+\cubex/2,\yo,\zo) {#8};
\node [left] at (\xo,\yo-\cubey/2,\zo) {#7};
\node [above] at (\xo,\yo,\zo) {#9};
\draw[thick,black,fill=white,->] (\xo,\yo,\zo) -- ++(0,0,-\cubez);
\draw[thick,black,fill=white,->] (\xo,\yo,\zo) -- ++(0,-\cubey,0); 
}

\endlocaldefs

\begin{document}

\begin{frontmatter}
\title{Bayesian Allocation Model: \\Inference by Sequential Monte Carlo for Nonnegative Tensor Factorizations and Topic Models using P{\'o}lya Urns}
\runtitle{Bayesian Allocation Model}

\begin{aug}
\author{\fnms{Ali Taylan} \snm{Cemgil}\thanksref{boun},
\ead[label=e1]{taylan.cemgil@boun.edu.tr}
}
\author{\fnms{Mehmet Burak} \snm{Kurutmaz}\thanksref{boun},
\ead[label=e2]{burak.kurutmaz@boun.edu.tr}
}
\author{\fnms{Sinan} \snm{Y{\i}ld{\i}r{\i}m}\thanksref{su},
\ead[label=e3]{sinanyildirim@sabanciuniv.edu}
}
\author{\fnms{Melih} \snm{Barsbey}\thanksref{boun},
\ead[label=e4]{melih.barsbey@boun.edu.tr}
}
\author{\fnms{Umut} \snm{\c{S}im\c{s}ekli}\thanksref{paristech}
\ead[label=e5]{umut.simsekli@telecom-paristech.fr}
}

\address[boun]{Departmant of Computer Engineering, Bo\u{g}azi\c{c}i University, \.{I}stanbul, Turkey
, \printead{e1,e2,e4}
}
\address[su]{Faculty of Engineering and Natural Sciences, Sabanc{\i} University, \.{I}stanbul, Turkey
, \printead{e3}
}
\address[paristech]{LTCI, T\'el\'ecom ParisTech, Universit\'{e} Paris-Saclay, 75013, Paris, France
, \printead{e5}
}

\runauthor{Cemgil, Kurutmaz, Y{\i}ld{\i}r{\i}m, Barsbey, \c{S}im\c{s}ekli}
\end{aug}
 
\begin{abstract}
We introduce a dynamic generative model, Bayesian allocation model (BAM), which establishes explicit connections between nonnegative tensor factorization (NTF), graphical models of discrete probability distributions and their Bayesian extensions, and the topic models such as the latent Dirichlet allocation. BAM is based on a Poisson process, whose events are marked by using a Bayesian network, where the conditional probability tables of this network are then integrated out analytically. We show that the resulting marginal process turns out to be a \Polya urn, an integer valued self-reinforcing process. This urn processes, which we name a \PolyaBayes process, obey certain conditional independence properties that provide further insight about the nature of NTF. These insights also let us develop space efficient simulation algorithms that respect the potential sparsity of data: we propose a class of sequential importance sampling algorithms for computing NTF and approximating their marginal likelihood, which would be useful for model selection. The resulting methods can also be viewed as a model scoring method for topic models and discrete Bayesian networks with hidden variables. The new algorithms have favourable properties in the sparse data regime when contrasted with variational algorithms that become more accurate when the total sum of the elements of the observed tensor goes to infinity. We illustrate the performance on several examples and numerically study the behaviour of the algorithms for various data regimes.
\end{abstract}


\end{frontmatter}

\section{Introduction} \label{sec: Introduction}
Matrices and tensors are natural candidates for the representation of  relations between two or more entities \citep{cichocki09, Kolda2009}. In matrix factorization, the goal is computing an exact or approximate decomposition of a matrix $X$ of form $X \approx WH$, possibly subject to various constraints on factor matrices $W$, $H$ or both. The decomposition, also called a factorization due to its algebraic structure, provides a latent representation of data that can be used for prediction, completion, anomaly detection, denoising or separation tasks to name a few. 

The difficulty of the matrix decomposition problem may depend on the nature of the constraints on factor matrices $W, H$ or the nature of the error function used for measuring the approximation quality. For many matrix decomposition problems, exact algorithms are not known. An important exception is the ubiquitous singular value decomposition where efficient and numerically stable algorithms are known for the computation of a decomposition of form $X=U^\top \Sigma V$ with orthonormal $U$ and $V$ and nonnegative diagonal $\Sigma$ for any matrix \citep{Golub2013}. Instead, when both factors $W$ and $H$ are required to be nonnegative, we obtain the nonnegative matrix factorization (NMF) \citep{Cohen1993, Paatero1994, Lee2001} and the problem becomes intractable in general \citep{Vavasis2007, Gillis2017}. However, under further conditions on the matrices $W$ and $H$, such as separability \citep{Donoho_and_Stodden_2004} or orthogonality \citep{Asteris_et_al_2015}, it becomes possible to obtain provable algorithms with certain optimality guarantees \citep{Gillis2012, Arora2016, Gillis2014, NIPS2015_5998}. Despite the lack of practical general exact algorithms, NMF and its various variants have been extensively investigated in the last decades \citep{Gillis2017}.

In domains where it is natural to represent data as a multi-way array (e.g., a three-way array is denoted as $X \in \mathbb{R}^{I \times J \times K}$, where $I,J,K \in \mathbb{N}_+$) involving more then two entities, tensor decomposition methods are more natural \citep{Acar2009, Kolda2009,cichocki09}, along with several structured extensions such as coupled/collective factorizations \citep{Paatero1999,yilmazGTF}, tensor trains and tensor networks \citep{Cichocki2016,Cichocki2017}. Unfortunately, in contrast to matrix decompositions, computing exact or approximate tensor factorizations is a much harder task that is known to be an NP-hard problem in general \citep{Hastad1990,Hillar2013}. Algorithms used in practice mostly rely on non-convex optimization, and depend on iterative minimization of a discrepancy measure between the target data and the desired decomposition. Despite their lack of optimality guarantees, tensor decomposition algorithms have found many modern applications as surveyed in \citet{Sidiropoulos2017, Papalexakis2016}, psychometrics \citep{tucker, harshman1970fpp}, knowledge bases \citep{Nickel2016}, graph analysis, social networks, \citep{Papalexakis2016}, music, audio, source separation  \citep{Virtanen2015, Simsekli2015}, communications, channel estimation \citep{Kofidis2001, Sidiropoulos2017}, bioinformatics \citep{Morup2008,Acar2011}, link prediction \citep{Ermis_2015}, and computational social science \citep{Schein16}.

Another closely related set of methods for relational data are the probabilistic topic models \citep{Blei2012}. Topic models have emerged even back in the early 1990's primarily from the need of understanding vector space models for the analysis and organizing large text document corpora \citep{Deerwester1990, Papadimitriou2000, Arora2015}, but quickly applied to other data modalities such as images, video, audio \citep{ Signoretto,Liu2013,Abdallah2007, Virtanen2008}. Research in this direction is mostly influenced by the seminal works \citet{Blei2003, Minka2002} and shortly thereafter many related models have been proposed \citep{Canny2004, Li2006, Airoldi2008, ScheinPBW15}. The empirical success of topic models has also triggered further research in  several alternative inference strategies \citep{Griffiths2004a, Teh2007}. 

It has been a folkloric knowledge that factorizing nonnegative matrices and fitting topic models to document corpora are closely related problems where both problems can be viewed as inference and learning in graphical models \citep{gopalan2013scalable}. In this paper, we formalize this observation and illustrate that nonnegative tensor factorization (NTF) models and probabilistic topic models have the same algebraic form as an undirected graphical model, which `factorizes' a joint distribution as a product of local compatibility functions. Based on this observation, we develop a general modeling framework that allows us computing tensor factorizations as well as approximating the marginal likelihood of a tensor factorization model for Bayesian model selection. 

In our construction, the central object is a dynamic model that we coin as the \emph{Bayesian allocation model} (BAM). The model first defines a homogeneous Poisson process on the real line \citep{Kingman1993, Daley2007} and we imagine the increments of this Poisson process as `tokens', drawing from the topic modeling literature. Each token is then marked in an independently random fashion using mark probabilities $\Theta$. The mark probabilities $\Theta$ are assumed to obey a certain factorization that respects a Bayesian network of discrete random variables, with the directed graph\footnote{For developing a dynamical process view, we find working with directed graphical models more convenient as compared to undirected graphs. A directed graph-based factorization of a joint distribution, also known as a Bayesian network, is a standard tool in modeling domains containing several interacting variables \citep{Lauritzen1996}.} $\mathcal{G}$ \citep{Lauritzen1996, Barber2012} where the probability tables corresponding to $\mathcal{G}$ are also random with a Dirichlet prior. Thanks to the conjugacy properties of the Poisson process and the Dirichlet-Gamma distributions, we can integrate out the random probability tables to arrive at a marginal model that turns out to be a \Polya urn process. The Markov properties of the directed graph $\mathcal{G}$ allows us to understand the nature of this \Polya urn model.

For general graphs, exact inference, i.e., the computation of certain marginal probabilities conditioned on the observations, can be accomplished using the junction tree algorithm \citep{Lauritzen1988, Lauritzen1992, Spiegelhalter1993}. For graphical models with discrete probability tables, under suitable assumptions, it is also possible to integrate out the probability tables to calculate the marginal likelihood. In this paper, we will show that computing the marginal likelihood of BAM is equivalent to computing the probability that the aforementioned \Polya urn process hits a certain set in its state space. This dynamic view leads to a class of novel sequential algorithms for the calculation of the marginal likelihood.

Note that hierarchical probability models, including topic models and tensor factorizations can be always expressed as Bayesian networks that include both the involved random variables as well as the probability tables. Indeed in the topic modeling literature, it is a common practice to express a joint distribution induced by a topic model using a plate notation. We argue that this detailed description may sometime blur the direct connection of topic models to Bayesian networks. We will illustrate that many known topic models, including Latent Dirichlet allocation, mixed membership stochastic blockmodels, or various NTF models such as nonnegative Tucker or canonical Polyadic decompositions can be viewed in a much simpler way, by just focusing on the discrete random variables and analytically integrating out the probability tables. 

The key contribution of the present paper is a surprisingly simple generic sequential Monte Carlo (SMC) algorithm that exploits and respects the sparsity of observed tensors. The computational complexity of the algorithm scales with the total sum of the elements in the observed tensor to be decomposed and the treewidth of the graph $\mathcal G$ that defines the mark distribution. Unlike other approaches that are based on standard techniques such as non-convex optimization, variational Bayes, or Markov chain Monte Carlo \citep{Acar2009, csimcsekli2012markov, ermis2014variational, nguyen2018efficient}, the proposed algorithm does not depend on the size of the observed tensor. This is a direct consequence of the junction tree factorization that implies also a factorized representation of the \Polya urn. Moreover, since the tokens are allocated one by one to the urn (due to the Poisson process formulation), the representation is typically sparse and this allows the use of efficient data structures for implementation. 

The paper is organized as follows. In Section \ref{sec:background}, we give a review of NTF models and setup our notation to highlight the connections to topic models and Bayesian networks. Then, Section \ref{sec: Allocation Model} introduces BAM as a generative model for tensors and discusses the properties of the resulting model. In the subsequent Section \ref{sec: Inference by Sequential Monte Carlo}, we show how the generic model can be viewed as a \Polya urn model and derive an SMC algorithm for unbiased estimation of the marginal likelihood. Finally, Section \ref{sec: Simulation Results} contains a simulation study to illustrate the performance and practical utility of the derived algorithms when compared with variational method. We illustrate with examples where our algorithm is practical in the sparse data regime. To make the paper self contained, we also provide a generic derivation of VB \citep{beal2006variational} and related optimization techniques in the same section. We conclude with the Section \ref{sec: Discussion and Conclusions}. 

\section{Background}\label{sec:background}
In this section, we will give a short review of matrix and tensor factorizations, topic models and probabilistic graphical models. All of these subjects are quite mature where excellent and extensive reviews are available, so our purpose will be setting up a common notation and building motivation for the rest of the paper.  

\subsection{Nonnegative Matrix and Nonnegative Tensor Factorizations}\label{sec: Nonnegative Matrix and Nonnegative Tensor Factorizations}
First, we provide a review of nonnegative matrix and tensor factorizations, the latter being a generalization of the former.

\subsubsection{Nonnegative matrix factorization} \label{sec: Nonnegative matrix factorization}
Nonnegative matrix factorization (NMF) is a statistical model that has been a widely adopted method for data analysis \citep{Paatero1994, Cohen1993, Lee2001, Gillis2017}. In its original formulation, the NMF problem is typically cast as a minimization problem:
\begin{eqnarray}
W^*, H^* & = & \arg\min_{W,H} D(X||WH), \label{eq:nmf-original}
\end{eqnarray}
where $X$ is an $I\times J$ observed matrix and $W$ and $H$ are factor matrices of sizes $I\times K$ and $K \times J$ respectively with nonnegative entries such that a divergence $D$ is minimized. One geometric interpretation is that the model seeks $K$ nonnegative basis vectors $W_{:k}$ for $k\in [K] \equiv\{1,2,\dots, K\} $ represent each data vector $X_{:j}$ by a conic (nonnegative) combination of $\sum_k W_{:k} H_{kj}$. Here, replacing an index with a $:$ denotes a vector as in $W_{:k} = [W_{1k}, W_{2k},\dots, W_{Ik}]^\top$.  As nonnegativity prevents cancellation, typical solutions of the matrices $W$ and $H$ are empirically known to have qualitatively different behaviour than other matrix decomposition models such as principal component analysis, that can be obtained by singular value decomposition.  

One possible divergence choice for $D$ is the information divergence, 
\begin{eqnarray*}
D(p||q) & = & \sum_i d_{\text{KL}}(p_i|| q_i),
\end{eqnarray*}
where
\begin{eqnarray*}
d_{\text{KL}}(p_i|| q_i) & = & p_i \log p_i - p_i \log q_i - p_i + q_i.
\end{eqnarray*}
We will refer to the corresponding model as \emph{KL-NMF}, following the established terminology in the literature. We note that the abbreviation KL refers to the Kullback-Leibler divergence that is actually the information divergence between suitably normalized measures $p$ and $q$ where $\sum_i p_i = \sum_i q_i$. In the sequel we will assume that the matrix $X$ has nonnegative integer entries, i.e., it is a count matrix. For practical considerations, this is not a major restriction as one can always scale finite precision rational numbers to integers by appropriate scaling.

When $X$ is a count matrix, the model is sometimes referred as the Poisson factorization model: the negative log-likelihood of the Poisson intensity is equal, up to constant terms, to the information divergence \citep{Cemgil2009,gopalan2013scalable}. More precisely, if
\begin{eqnarray}
X_{ij}|W, H  & \stackrel{\text{i.i.d.}}{\sim} & \mathcal{PO} \left(\sum_{k = 1}^{K} W_{ik}H_{kj} \right),  \quad i\in[I], j\in[J], \label{eq:KLNMFx}
\end{eqnarray}
the log-likelihood is
\begin{eqnarray*}
\ell_X(W, H) & = & - D(X||WH) + \text{const}.
\end{eqnarray*}
Here, the statistical interpretation of NMF is estimating a low rank intensity matrix that maximizes the likelihood of observed data.

\subsubsection{Nonnegative tensor factorization} \label{sec: Nonnegative tensor factorization}
Nonnegative tensor factorization (NTF) models can be viewed as structured extensions of NMF, in a sense we will now describe. Note that the matrix product $WH$ is a particular matrix valued bilinear function $\mathcal{T} = \mathcal{T}[W, H]$, the element-wise products $W$ and $H$ followed by a contraction on index $k$ as $\mathcal{T}(i,j) = \sum_k W_{ik}H_{kj}$. To express general models, we will introduce a multi-index notation where for a set of integers
$U=\{m,n,r,\dots\}$ we let $a_U = (a_m, a_n, a_r,\dots)$ and for any $m<n$ we let $m:n$ denote $\{m,m+1,\dots, n\}$. In the sequel, we will use NMF as a running example.

In the NMF model, we have $L=2$ factor matrices and $N=3$ indices. We redefine the factor matrices $W_1 = W$, $W_2 = H$ and rename the indices as $i_1=i, i_2=j, i_3=k$ and define the sets $U_1 = \{1,3\}, U_2 = \{2,3\}$, $V = \{1,2\}$ and $\bar{V} = \{3\}$. We let $W_1(i_{U_1})$ denote $W_1(i_1, i_3)$ (that is $W_{ik}$), and $W_2(i_{U_2})$ denote $W_2(i_2, i_3)$ (which is $H_{kj}$). The contraction index set is $\bar{V}$ so the bilinear function can be expressed as $\mathcal{T}[W_1, W_2](i_V) = \sum_{i_{\bar{V}}} W_1(i_{U_1})W_2(i_{U_2})$ (which is just $\mathcal{T}[W, H](i,j) = \sum_k W_{ik}H_{kj}$).

In tensor factorization, the goal is finding an approximate decomposition of a target tensor $X$ as a product of $L$ factor tensors $W_1, W_2, \dots, W_L \equiv W_{1:L}$, on a total of $N$ indices. The target tensor has indices $i_V$ where $V \subset [N]$, where each element is denoted as $X(i_V)$ and the order of this target tensor is the cardinality of $V$, denoted by $|V|$. Each of the factor tensors $W_l$ have indices $i_{U_l}$ where $U_l \subset [N]$ and each element is denoted as $W_l(i_{U_l})$. 
As such, a tensor factorization 
model is 
\begin{eqnarray}
W_{1:L}^* & = & \arg\min_{W_{1:L}} D(X|| \mathcal{T}[W_{1:L}] ) \nonumber \\
& = & \arg\min_{W_{1:L}} \sum_{i_V} d_{\text{KL}}\left(X(i_V) || \mathcal{T}[W_{1:L}](i_V) \right), \label{eq:ntf-original}
\end{eqnarray}
where 
\begin{eqnarray}
\mathcal{T}[W_{1:L}](i_V) = \sum_{i_{\bar{V}}} \prod_{l=1}^L W_l(i_{U_l}) \label{eq:ntf-model}
\end{eqnarray}
and we use the bar notation to denote the complement of an index set as $\bar{V} = [N] \setminus V$. 
Note that this is simply a particular marginal sum over the indices $i_{\bar{V}}$ that are not members of the target tensor. For any set $U \subset [N]$ we will refer to the corresponding marginal using the same notation $\mathcal{T}[W_{1:L}](i_U) = \sum_{i_{\bar{U}}} \prod_{l=1}^L W_l(i_{U_l})$.

Using the notation above, we can define some known models easily. For example, a canonical Polyadic decomposition (also known as a PARAFAC) model is defined by
\begin{eqnarray}
\mathcal{T}[W_1, W_2, W_3](i_1,i_2,i_3) = \sum_{i_4} W_1(i_1, i_4) W_2(i_2, i_4) W_3(i_3, i_4) \label{eq:parafac}
\end{eqnarray}
and can be encoded as $U_1 = \{1,4\}$, $U_2 = \{2,4\}$, $U_3 = \{3,4\}$, $V=\{1,2,3\}$, $\bar{V} = \{4\}$, where we decompose an order $|V| = 3$ tensor as a product of $L = 3$ factor matrices. Another common model is the Tucker model
\begin{eqnarray}
\mathcal{T}[W_1, W_2, W_3, W_4](i_1,i_2,i_3) = \sum_{i_4, i_5, i_6} W_1(i_1, i_4) W_2(i_2, i_5) W_3(i_3, i_6) W_4(i_4, i_5, i_6) \label{eq:tucker}
\end{eqnarray}
that corresponds to $U_1 = \{1,4\}$, $U_2 = \{2,5\}$, $U_3 = \{3,6\}$, $U_4 = \{4,5,6\}$, $V=\{1,2,3\}$, $\bar{V} = \{4,5,6\}$,  
where we decompose an order $|V| = 3$ tensor as a product of $L = 4$ tensors, three factor matrices and a so-called core-tensor. Many other tensor models, such as tensor trains \citep{Oseledets_2011}, tensor networks \citep{Orus2013, Cichocki2016, Cichocki2017} can also be expressed in this framework. 

There exist a plethora of divergences that can be used for measuring the quality of an approximate tensor factorization \citep{Fevotte2010, Finesso2006, Cichocki2006}. In this paper, we focus on Poisson tensor factorization, that has a particularly convenient form due to the properties of the information divergence and is a natural choice in fitting probability models. The derivative of the scalar information divergence with respect to a scalar parameter $w$ is ${\partial} d_{\text{KL}}(p|| q(w))/{\partial w}  =   \left(1 - {p}/{q(w)}\right){\partial q(w)}/{\partial w}$. Thanks to the multilinear structure of a tensor model, the partial derivatives of the divergence $D$ can be written explicitly as the difference of two positive terms
\begin{eqnarray}
\frac{\partial D(X||\mathcal{T}[W_{1:L}])}{\partial W_l(i_{U_l})} & = & \frac{\partial}{\partial W_l(i_{U_l})} \sum_{i_V} d_{\text{KL}} (X(i_V)|| \mathcal{T}[W_{1:L}](i_V)) \nonumber \\
& = & \sum_{i_{\bar{U}_l}} \left(1 - \frac{X(i_V)}{\mathcal{T}[W_{1:L}](i_V)}\right) \frac{\partial \mathcal{T}[W_{1:L}](i_V) }{\partial W_l(i_{U_l})} \nonumber \\
 & = &  - \sum_{i_{\bar{U}_l}}\frac{X(i_V)}{\mathcal{T}[W_{1:L}](i_V)} \prod_{l'\neq l} W_{l'}(i_{U_{l'}}) + \sum_{i_{\bar{U}_l}}  \prod_{l'\neq l} W_{l'}(i_{U_{l'}})  \nonumber \\
 & \equiv & -(\nabla^-_l D)(i_{U_l}) + (\nabla^+_l D)(i_{U_l}) \equiv (\nabla_l D)(i_{U_l}). \nonumber
\end{eqnarray}
After writing the Lagrangian of the objective function \eqref{eq:ntf-original} for all factors $W_l$ for $l=1\dots L$, necessary optimality criteria can be derived for each $W_l$ from the Karusch-Kuhn-Tucker (KKT) conditions (see, e.g., \citet{Kazemipour_et_al_2017}) as
\begin{eqnarray*}
0 = (\nabla_l D)(i_{U_l}) W_l(i_{U_l}),
\end{eqnarray*}
or, explicitly,
\begin{eqnarray}
\sum_{i_{\bar{U}_l}}\frac{X(i_V)}{\mathcal{T}[W_{1:L}](i_V)} \prod_{l'} W_{l'}(i_{U_{l'}}) =  \sum_{i_{\bar{U}_l}}  \prod_{l'} W_{l'}(i_{U_{l'}}) \label{eq:KKT_cond}
\end{eqnarray}
for all $i_{U_{l}}$, $l = 1, \ldots, L$. For solving the KKT conditions, one can derive a multiplicative algorithm for each factor with the ratio of the negative and positive parts of the gradient  
\begin{equation}
W_l(i_{U_l}) \leftarrow W_l(i_{U_l}) \frac{\sum_{i_{\bar{U}_l}}\frac{X(i_V)}{\mathcal{T}(i_V)} \prod_{l'\neq l}^L W_{l'}(i_{U_{l'}})}{\sum_{i_{\bar{U}_l}}  \prod_{l'\neq l}^L W_{l'}(i_{U_{l'}})}  = W_l(i_{U_l}) \frac{(\nabla^-_l D)(i_{U_l})}{(\nabla^+_l D)(i_{U_l})}. \label{eq:mult_update}
\end{equation}
It can be verified that this algorithm is a local search algorithm that chooses a descent direction as $W_l(i_{U_l})$ increases when $(\nabla_l D)(i_{U_l})<0$ and decreases when $(\nabla_l D)(i_{U_l})>0$. In the KL case, the same algorithm can also be derived as an expectation-maximization (EM) algorithm by data augmentation \citep{Cemgil2009, yilmazGTF}. For the matrix (NMF) case, the convergence properties of the multiplicative updates are established in \citet{Finesso2006, Chih-JenLin2007}. 

An alternative interpretation of the KKT optimality conditions is a balance condition where the marginals on $i_{U_l}$ must match on both sides for all $l \in [L]$
\begin{eqnarray}
\sum_{i_{\bar{U}_l}} \mathcal{P}[W_{1:L}](i_{\bar{V}}|i_V) X(i_V) & = &  \sum_{i_{\bar{U}_l}} \mathcal{T}[W_{1:L}](i_V, i_{\bar{V}}), \label{eq:KKT_cond-alt}
\end{eqnarray}
where we let 
\[
\mathcal{P}[W_{1:L}](i_{\bar{V}}|i_V) \equiv \frac{\prod_{l'} W_{l'}(i_{U_{l'}})}{\mathcal{T}[W_{1:L}](i_V)} = \frac{\mathcal{T}[W_{1:L}](i_V, i_{\bar{V}})}{\mathcal{T}[W_{1:L}](i_V)}
\]
and use the letter $\mathcal{P}$ to highlight the fact that this quantity can be interpreted as a conditional probability measure. 

For a general tensor model, calculating these marginals, hence the gradient required for learning can be easily intractable. The theory of probabilistic graphical models \citep{Lauritzen1988, Maathuis2018} provides a precise characterization of the difficulty of this computation. In fact, the NTF model in \eqref{eq:ntf-model} is equivalent to an undirected graphical model with hidden random variables. Here, the undirected graph is defined with a node for each index, and for each pair of nodes there is an undirected edge in the edge set if two indices appear together in an index set $U_l$ for $l = 1\dots L$.  Exact inference turns out to be exponential in the treewidth of this undirected graph. The relations between tensor networks and graphical models have also been noted \citep{Robeva2018}. Due to this close connection, we will review graphical models and closely related topic models, that have a well understood probabilistic interpretation and highlight their connections to NTF models. It will turn out, that the concept of conditional independence translates directly to a `low rank' structure often employed in construction of tensor models.

\subsection{Probabilistic Graphical Models}\label{sec: Probabilistic Graphical Models}
Probabilistic graphical models are a graph based formalism for specifying joint distributions of several random variables $\xi_1, \dots, \xi_N$ and for developing computation algorithms \citep{Maathuis2018}. For discrete random variables, when each $\xi_n$ for $n \in [N]$ takes values in a finite set with $I_n$ elements, one natural parametrization of the distribution is via an $N$-way table $\theta$, where each element $\theta(i_1,\dots,i_N)$ for $i_n \in [I_n]$ specifies the probability of the event $\xi_1 = i_1 \wedge \xi_2 = i_2 \wedge \dots \wedge \xi_N = i_N$, expressed as $\Pr\{\xi_{1:N} = i_{1:N}\}$. More formally, we have 
\[
\theta(i_{1:N}) = \E{\prod_{n = 1}^{N} \ind{\xi_n = i_n} } = \Pr\{\xi_{1:N} = i_{1:N}\},
\]
where $\E{\cdot}$ denotes the expectation with respect to the joint distribution $\Pr$ and $\ind{\text{cond}}$ is the indicator of the condition, that is $1$ if the condition inside the bracket is true and $0$ otherwise. 

For large $N$, explicitly storing the table $\theta$ is not feasible due to the fact that its size is growing exponentially with $N$, so alternative representations are of great practical interest. One such representation is the undirected graphical models briefly mentioned in the previous section. Another alternative representation is a Bayesian network \citep{Pearl1988}. In this representation, the probability model is specified with a reference to a graph ${\mathcal G} = (V_{\mathcal G}, E_{\mathcal G})$ with the set of vertices $V_{\mathcal{G}} = [N]$ and directed edges $E_{\mathcal G}$. This graph is a directed acyclic graph (DAG) meaning that the set of directed edges $E_{\mathcal{G}} \subset V_{\mathcal{G}} \times V_{\mathcal{G}}$ are chosen such that there are no directed cycles in $\mathcal{G}$. Each vertex $n$ of the graph corresponds to a random variable and missing edges represent conditional independence relations. 

Given a directed acyclic graph $\mathcal{G}$, for each vertex $n$, we define the set of {\emph{parents}} $\pa{n}$ as the set of vertices $v$ from which there is a directed edge incident to node $n$, defined by $\pa{n} \equiv \{v: (v,n) \in E_{\mathcal{G}}\}$. We will also define the \emph{family} of the node $n$ as $\fa{n} \equiv \{ n\} \cup \pa{n}$. With these definitions, the Bayesian network encodes nothing but a factored representation of the original probability table as:
\begin{equation}
\theta({i_{1:N}}) = \prod_{n = 1}^{N} \theta_{n|\pa{n}} (i_n, i_\pa{n}). \label{eq:bn1}
\end{equation}

The factors are typically specified during model construction or need to be estimated from data once a structure is fixed. In theory, to obtain such a factorized representation, we can associate each factor $\theta_{n|\pa{n}}$ with a conditional distribution of form $\Pr(\xi_n| \xi_{\pa{n}})$ defined by the ratio of two marginals
\[
\Pr(\xi_n = i_{n}| \xi_{\pa{n}} = i_{\pa{n}}) = \frac{\Pr(\xi_{\fa{n}} = i_{\fa{n}})}{\Pr(\xi_{\pa{n}} = i_{\pa{n}})},
\]
where the notation $\xi_U$ refers to a collection of random variables, indexed by $U$ as $\xi_U = \{\xi_u: u \in U\}$ for any $U \subset [N]$. Formally, we define the (moment) parameters of a marginal distribution $\Pr(\xi_U)$ as 
\[
\theta_{U}(i_U) = \E{\prod_{u \in U}\ind{\xi_u = i_u}} = \sum_{i'_{1:N}} \theta(i'_{1:N}) \prod_{u \in U}\ind{i'_u = i_u} = \sum_{i_{\bar{U}}} \theta(i_{1:N})
\]
as such a marginal is, in the tensor terminology introduced in the previous section, a contraction on $i_{\bar{U}}$ where we have $\theta_{U}(i_U) =\sum_{i_{\bar{U}}} \theta(i_{1:N})$. We can define a conditional probability table as:
\[
\theta_{n|\pa{n}}(i_n, i_\pa{n}) = \frac{\theta_{\fa{n}}(i_n, i_{\pa{n}})}{\theta_{\pa{n}}(i_{\pa{n}})} = \Pr\{\xi_n = i_n| \xi_\pa{n}=i_\pa{n} \}.
\]
We use the notation $n|\pa{n}$ just to highlight the fact that $\sum_{i_n} \theta_{n|\pa{n}}(i_n, i_\pa{n}) = 1$, for all $i_\pa{n}$.

One key problem in probabilistic graphical models is the inference problem, that is the computation of marginal probabilities conditioned on a subset of random variables $\xi_V$ observed to be in a specific state, where $V \subset [N]$ is the set of \emph{visible indices}. Exact inference is feasible, but this depends critically on the structure of the graph $\mathcal{G}$ as well as the particular set of visible indices $V$. To highlight the relation to tensor factorization, we will describe the conditioning using tensor operations and define an \emph{observation} as a tensor with a single nonzero entry
\[
s_V(i_V) \equiv \ind{\xi_{V} = i_{V}} = \prod_{v \in V} \ind{\xi_v = i_v}.
\]
The inferential goal is, given $\xi_{V} = i^{\ast}_{V}$ (so that $s_{V}(i^{\ast}_{V}) = 1$) and $U \subset [N]$, computing posterior marginals of form
\begin{align*}
\Pr\{\xi_{U} = i_{U} | \xi_{V} = i^{\ast}_V \} &=  \sum_{i_{\bar{U}}} \frac{ \Pr\{\xi_{\bar{V}}= i_{\bar{V}}, \xi_{V}=i_{V}\} }{\Pr\{\xi_{V} = i_V\}} \ind{\xi_{V} = i_{V}} \\
&= \sum_{i_{\bar{U}}} \frac{ \theta(i_{1:N}) }{\sum_{i'_{\bar{V}}} \theta((i_{V}, i'_{\bar{V}})) } s_{V}(i_{V}).
\end{align*}

A general method for exact inference in graphical models is the junction tree algorithm (see, e.g., \citet{Lauritzen1988}) that proceeds by combining groups of nodes into cliques by moralization and triangulation steps to arrive at a representation as
\begin{equation}
\Pr(\xi_{1:N} = i_{1:N}) = \frac{\prod_{C \in \Cliques} \Pr(\xi_{C} = i_{C}) }{\prod_{D \in \Separators} \Pr(\xi_{D} = i_{D})}, \label{eq:jtfact}
\end{equation}
where $\Cliques$ and $\Separators$ are collection of sets named \emph{cliques} and \emph{separators} satisfying the running intersection property. This factorization allows a propagation algorithm on a tree for efficiently computing desired marginals. In a sense, the junction tree can be viewed as a compact representation of the joint distribution from which desired posterior marginals can still be computed efficiently. In general, each clique $C \in \Cliques$ will contain at least one of the families of $\fa{n}$ so the junction tree algorithm provides a practical method that facilitates the computation of $\Pr\{\xi_{\fa{n}} = i_{\fa{n}} | \xi_{V} = i_V\}$ for all $n$, including cases where some variables are observed, i.e., when $\fa{n} \cap V$ is not empty.

We note that these posterior marginals are in fact closely related to the required gradients when solving the KKT conditions in \eqref{eq:KKT_cond} for the NTF model. In other words, the gradients required when fitting a tensor model to data can be computed in principle by the junction tree algorithm. To see this, we assume that the observed tensor $X$ is in fact a contingency table of $T$ observations, $\xi_{V}^{1}, \ldots, \xi_{V}^{T}$ (with corresponding tensors $s^{1}_{V}, \ldots, s^{T}_{V}$), where each cell of $X$ gives the total count of observing the index $i_V$
\[
X(i_V) \equiv \sum_{\tau = 1}^{T} s^{\tau}_{V}(i_{V}) = \sum_{\tau=1}^{T} \ind{\xi_V^{\tau} = i_V}.
\]
In this case, given $\xi_{V}^{\tau} = i_{V}^{\tau}$ (so that $s^{\tau}(i_{V}^{\tau}) = 1$) for $\tau \in [T]$, the total gradient, also known as the expected sufficient statistics, is given by 
\begin{align*}
\sum_{\tau=1}^{T} \Pr\{\xi_{U} = i_{U} | \xi_{V} = i_V^\tau \} &= \sum_{\tau=1}^{T} \sum_{i_{\bar{U}}} \frac{\theta(i_{1:N})}{\sum_{i'_{\bar{V}}} \theta((i_{V}, i'_{\bar{V}})) } s_{V}^{\tau}(i_{V}) \\
&= \sum_{i_{\bar{U}}} \frac{\theta(i_{1:N})}{\sum_{i'_{\bar{V}}} \theta((i_{V}, i'_{\bar{V}})) } X(i_V).
\end{align*}
We will not further delve into the technical details of the junction tree algorithm here but refer the reader to the literature \citep{Lauritzen1996, Cowell2003, Barber2012}.

\subsection{Topic Models}\label{sec: Topic Models}
Topic models are a class of hierarchical probabilistic generative models for relational data with the latent Dirichlet allocation (LDA) as the prototypical example \citep{Blei2003}. The LDA is introduced as a generative model for document collections and can be viewed as a full Bayesian treatment of a closely related model, probabilistic latent semantic indexing (PLSI) \citep{Hofmann_1999}. Suppose we are given a corpus of $J$ documents with a total of $S_+$ words from a dictionary of size $I$. To avoid confusion, we will refer to the individual instances of words as tokens. Formally, for each token $\tau$ where $\tau \in [S_+]$, we are given a data set of document labels $j \in [J]$ and word labels $i \in [I]$ from a fixed dictionary of size $I$. This information is encoded as a product of two indicators $d^{\tau}$ and $w^{\tau}$: $d^\tau_j w^\tau_i = 1$ if and only if token $\tau$ comes from document $j$ and it is assigned to word $i$.

\begin{figure}
\begin{psmatrix}
\begin{tikzpicture}[>=stealth',auto,scale=0.7, every node/.style={transform shape}]        

        \node[latent]                       (j)     {$d$};
        \node[latent, right=of j]        (k)     {$z$};
        \node[latent, right=of k]           (i)     {$w$};

        \edge {j} {k};
        \edge {k} {i};
\end{tikzpicture}
&
\begin{tikzpicture}[>=stealth',auto,scale=0.7, every node/.style={transform shape}]        

        \node[obs]                       (j)     {$d$};
        \node[latent, right=of j]        (k)     {$z$};
        \node[obs, right=of k]           (i)     {$w$};

        \node[latent, above=of k]        (t2)     {$\vartheta$};
        \node[latent, above=of i]           (t3)     {$\beta$};

        \node[obs, above=of t2]        (eta2)     {$\eta_\vartheta$};
        \node[obs, above=of t3]           (eta3)     {$\eta_\beta$};
           
        \edge {j} {k};
        \edge {k} {i};
        \edge {t2} {k};
        \edge {t3} {i};
        \edge {eta2}{t2} ;
        \edge {eta3}{t3} ;
\end{tikzpicture}
&
\begin{tikzpicture}[>=stealth',auto,scale=0.7, every node/.style={transform shape}]        

        \node[obs]                       (j)     {$d$};
        \node[latent, right=of j]        (k)     {$z$};
        \node[obs, right=of k]           (i)     {$w$};

        \node[latent,above=of j]         (t1)     {$\theta_d$};
        \node[latent, above=of k]        (t2)     {$\theta_{z|d}$};
        \node[latent, above=of i]           (t3)     {$\theta_{w|z}$};

        \node[obs, above=of t2]           (a)     {$\alpha$};
           
        \edge {j} {k};
        \edge {k} {i};
        \edge {t1} {j};
        \edge {t2} {k};
        \edge {t3} {i};
        \edge {a}{t1} ;
        \edge {a}{t2} ;
        \edge {a}{t3} ;
\end{tikzpicture}
\end{psmatrix}
\caption{(Left) A Bayesian network representation of the probability model $p(d)p(z|d)p(w|z)$, also known as the PLSI model. $d, z, w$ correspond to documents, topics and words. (Middle) Bayesian network representation of LDA for a single token (see text), LDA can be viewed as a full Bayesian treatment for the discrete graphical model on the left; (Right)  The equivalent model for KL-NMF and LDA derived from BAM. We extend this connection via graphical models to other NTF models. Consistency requires that all factors share a common base measure $\alpha$.}
\label{fig:gm}
\end{figure}
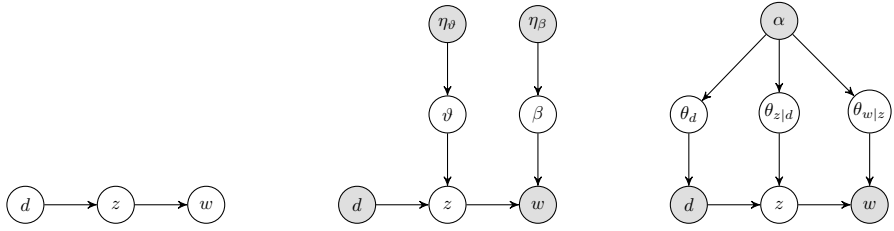

LDA is typically presented as a mixture model where one assumes that, conditioned on the document indicator $d^\tau$, the token $\tau$ first chooses a topic $k \in [K]$ among $K$ different topics,  with document specific topic probability $\vartheta_{kj}$,  then chooses the word $w^\tau$ conditioned on the topic with probability $\beta_{ik}$. This model can be summarized by the following hierarchical model for all $\tau$  
\begin{align}
\vartheta_{:j} & \sim  \mathcal{D}(\eta_\vartheta) & \beta_{:k} & \sim \mathcal{D}(\eta_\beta) \label{eq:lda_prior}\\
z_{: \tau}|d_{:\tau} & \sim \prod_{j = 1}^{J} \mathcal{M}(\vartheta_{:j}, 1)^{d_{j\tau}} & w_{: \tau}|z_{: \tau} & \sim \prod_{k = 1}^{K} \mathcal{M}(\beta_{:k},1)^{z_{k\tau}}. \nonumber
\end{align}
The inferential goal of LDA is, given $d$ and $w$ estimating the posterior of $z$ as well as the tables $\beta, \vartheta$. The corresponding simplified graphical model is illustrated in Figure \ref{fig:gm} Following the seminal work of \citet{Blei2003}, many more variations have been proposed \citep{Li2006, Airoldi2008}. One can view LDA as a hierarchical Bayesian model for the conditional distribution $p(w|d) = \sum_z p(w|z)p(z|d)$. As such, the approach can be viewed naturally as a Bayesian treatment of the graphical model $d \rightarrow z \rightarrow w$ with observed $d$ and $w$ and latent $z$; see Figure \ref{fig:gm}. In this paper, we will view topic models and structured tensor decompositions from the lens of discrete graphical models. 

In topic modeling, it is common to select independent priors on factor parameters, for example in \eqref{eq:lda_prior} the hyper-parameters could be specified freely. While this choice seems to be intuitive, it turns out having a quite dramatic effect on posterior inference and may even lead to possibly misleading conclusions. The choice of a Dirichlet may also seem to be arbitrary and merely due to convenience, but \citet{Geiger_and_Heckerman_1995} prove that under certain plausible assumptions the Dirichlet choice is inevitable. If the parameters of a Bayesian network are assumed to be \textit{globally and locally independent}, and there are no extra assumptions about the conditional independence structure of the model other than the ones directly encoded by the graph, i.e., any Markov equivalent graph structures could not be discriminated, then the only possible choice of priors happens to be a collection of Dirichlet distributions satisfying \textit{equivalent sample size} principle. In plain terms, this requires that all Dirichlet hyper parameters should be chosen consistently as marginal pseudo-counts from a fixed, common imaginary data set \citep{Heckerman1995}. These results, initially stated only for Bayesian structure learning problems in discrete graphical models with observable nodes directly apply to the much more general setting of topic models and tensor factorizations. Hence, in the following section, we will establish the link of topic models with tensor decomposition models and Bayesian networks. Our strategy will be introducing a dynamic model, that we coin as BAM and showing that this model can be used for constructing all the related models. 

\section{Allocation Model} \label{sec: Allocation Model}
We first define an {\emph{allocation process}} as a collection of homogeneous Poisson processes $\mathcal{S}^t(i_{1:N}), t \in [0,1)$ where each index $i_n$ for $n \in [N]$ has the cardinality of $I_n$, i.e., $i_n \in [I_n]$, so we have $\prod_n I_n$ processes. All processes are obtained by {\emph{marking}} an homogeneous Poisson process $\{\mathcal{S}^t, t \in [0,1) \}$ with constant intensity $\lambda$ that we name as the \emph{base process}. The marking is done by randomly assigning each event (\emph{token}) generated by the base process independently to the processes $\mathcal{S}^t(i_{1:N})$ with probability $\theta({i_{1:N}})$. Let the event times of the base process be $0<t_1 <  \dots < t_T < 1$ and define $\mathbf{S}^\tau(i_{1:N}) = \mathcal{S}^{t_\tau}(i_{1:N})$ for $\tau \in [T]$. Note that $T$ is random also. We will also define the increments 
\[
\mathbf{s}^\tau(i_{1:N}) = \mathbf{S}^\tau(i_{1:N}) - \mathbf{S}^{\tau-1}(i_{1:N}), \quad \tau = 1, \ldots, T
\]
with the convention $\mathbf{S}^{0}(i_{1:N})=0$. When $\tau=T$, we let $\mathbf{S} = \mathbf{S}^{T}$ and refer to this object as the \emph{allocation tensor}. 
It is useful to view the allocation tensor as a collection of cells where cell $i_{1:N}$ contains $\mathbf{S}({i_{1:N}})$ tokens at time $t=1$ and the increments $\mathbf{s}^\tau(i_{1:N})$ as indicators of the allocation of the $\tau$'th token to cell $i_{1:N}$. Later, it will be also useful to think of $i_{1:N}$ as a color, where each index $i_n$ encodes a color component.

As the assignments of tokens to cells are random, we can define the sequence of random variables $\mathbf{c}^{\tau}$, $\tau = 1, \ldots, T$ where $\mathbf{c}^{\tau}$ is the index of the Poisson process that is incremented at time $t_{\tau}$, i.e.,
\[
\mathbf{c}^{\tau} = (\mathbf{c}^{\tau}_{1}, \ldots, \mathbf{c}^{\tau}_{N} ) \in \bigotimes_{n = 1}^{N} [I_{n}], \quad \text{such that} \quad \mathbf{s}^{\tau}(\mathbf{c}^{\tau}) = 1.
\]
In the general case, we can imagine the joint distribution of $\mathbf{c}^{\tau}$ as an $N$-way array (an order $N$ tensor) $\theta$ with $\prod_{n} I_{n}$ entries, and $\theta({i_{1:N}})$ specifies the probability that a token is placed into the cell with index $i_{1:N}$,
\begin{equation} \label{eq: probability of increment given theta}
\Pr(\mathbf{c}^{\tau} = i_{1:N} | \theta) = \theta(i_{1:N}), \quad \tau \geq 1; \quad i_{1:N} \in  \bigotimes_{n = 1}^{N} [I_{n}]. 
\end{equation}
For modeling purposes, we will further assume that $\theta$ respects a factorization implied by a Bayesian network ${\mathcal G} = (V_{\mathcal G}, E_{\mathcal G})$ (see Section \ref{sec: Probabilistic Graphical Models}). As such, we have a factorization of $\theta$ that has the form 
\begin{equation} \label{eq:bn}
\theta({i_{1:N}}) = \prod_{n=1}^{N} \theta_{n|\pa{n}} (i_n, i_\pa{n}).
\end{equation}

\begin{rem}
We will generally use $S$, $S^{\tau}$, $s^{\tau}$, and $c^{\tau}$ for realizations of the random variables $\mathbf{S}$, $\mathbf{S}^{\tau}$, $\mathbf{s}^{\tau}$, $\mathbf{c}^{\tau}$, and preserve the relation between the random variables for their realizations as well. (For example, for some $S^{\tau-1}$ and $S^{\tau}$ we will use $s^{\tau}$ directly, with reference to the relation between the random variables they correspond to). Finally, we will resort to the bold-faced notation when the distinction between random variables and realizations is essential in the notation; otherwise we will stick to the lighter notation $S$, $S^{\tau}$, $s^{\tau}$, and $c^{\tau}$, etc.\ in the flow of our discussion.
\end{rem}

\subsection{Bayesian Allocation Model} \label{sec: Bayesian Allocation Model (BAM)}
To complete the probabilistic description,  we let $\lambda \sim \mathcal{GA}(a, b)$. For $\theta$, we start with an order-$N$ tensor $\alpha$ with nonnegative entries $\alpha(i_{1:N}) \geq 0$ for all $i_{1:N}$. In practice, we do not need to explicitly store or construct $\alpha$; we will use it to consistently define the contractions for all $n\in[N]$
\begin{equation} \label{eq:margal}
\alpha_{\fa{n}}(i_n, i_{\pa{n}})  \equiv \sum_{i_{\overline{\fa{n}}}} \alpha(i_{1:N})
\end{equation}
We model each factor $\theta_{n|\pa{n}}$ in \eqref{eq:bn} as an independent\footnote{Though we introduced the probability tables as independent, it is straightforward to generalize the generative framework to the cases where tables are dependent as well; see subsection \ref{subsec:letter} for an example.} random conditional probability table, and $\alpha_{\fa{n}}$ is used to assign the prior distribution on those tables.


Specifically, given the hyperparameters $(\alpha, a, b)$, the hierarchical generative model for variables $\lambda$, $\Theta \equiv \{\theta_{n|\pa{n}} : n \in [N]\}$ and $\mathbf{S}$ is constructed as:
\begin{align}
\begin{aligned}
& \lambda \sim  \mathcal{GA}(a, b), & \\
& \theta_{n|\pa{n}}(:, i_{\pa{n}}) \sim  \mathcal{D}(\alpha_{\fa{n}}(:, i_{\pa{n}})), & \forall n \in [N], \forall i_{\pa{n}}, \\ 
& \mathbf{S}(i_{1:N}) | (\Theta, \lambda)  \sim \mathcal{PO}\left(\lambda \prod_{n=1}^N \theta_{n|\pa{n}}(i_n, i_{\pa{n}})\right), & \forall i_{1:N}.
\end{aligned}
\label{eq:allocationmodel}
\end{align}
Here, $\alpha_{\fa{n}}(:,i_{\pa{n}})$ is an $I_n \times 1$ vector obtained by fixing $i_{\pa{n}}$ and varying $i_n$ only, and $\mathcal{D}(v)$ is the Dirichlet distirbution with parameter vector $v$. Given the hierarchical model, the joint distribution for the \emph{Bayesian allocation model} (BAM) is expressed as
\begin{align} 
\pi(S, \Theta, \lambda) = p_{a, b}(\lambda) p_{\alpha}(\Theta) p(S | \Theta, \lambda) \label{eq: joint of S theta lambda},
\end{align}
where the distributions (densities) $p_{a, b}(\lambda)$, $p_{\alpha}(\Theta)$ and $p(S | \Theta, \lambda)$ are induced by \eqref{eq:allocationmodel}. To make our reference to the assumed probability measure clear, we will use the notation $\pi$ denote distributions regarding the model in \eqref{eq:allocationmodel}.

\begin{rem}
Our construction of defining the prior parameters from a common $\alpha$ seems restrictive, however this is required for consistency: the factorization in \eqref{eq:bn} is not unique; there are alternative factorizations of $\theta$ that are all Markov equivalent \citep{Murphy2012}. If we require that the distribution of $\theta(i_{1:N})$ should be identical for all equivalent models then defining each measure as in \eqref{eq:margal} is necessary and sufficient to ensure consistency by the equivalent sample size principle \citep{Geiger_and_Heckerman_1995}. We discuss this further in Section \ref{sec: Hyperparameters}, where we provide a numerical demonstration in Example \ref{ex: Example (Consistent Priors)}.
\end{rem}

\subsubsection{Observed data as contractions} \label{sec: Observed data as contractions}
\begin{figure}[ht]
    \centering
\begin{center}
\begin{psmatrix}[rowsep=0.2cm,nodesep=.05cm,colsep=0.3cm]
\begin{tikzpicture}
\DrawCubeLabels{1}{1}{0.8}{0}{0}{0}{$i$}{$j$}{$k$}
\end{tikzpicture} 
&
\begin{tikzpicture}
\foreach \z in {0,...,3}
\foreach \y in {0,...,7}
\foreach \x in {0,...,3}
\DrawCube{0.15}{0.15}{0.15}{-\y*0.25}{\x*0.25}{\z*0.25};
\end{tikzpicture}
& 
\begin{tikzpicture}
\foreach \y in {0,...,7}
\foreach \x in {0,...,3}
\DrawCube{0.15}{0.15}{1}{-\y*0.25}{\x*0.25}{0};
\end{tikzpicture}\\
& $(i\;k\;j)$ & $(i\;\mycolon j)$ \\
& $S_{i_{1:N}}$ & $S_V(i_V) = X(i_V)$
\end{psmatrix}
\end{center}
    \caption{Visualization of the allocation tensor: each cube is a cell that is indexed by a tuple $i_{1:N}$ (here $ikj$) and $S(i_{1:N})$ denotes the numbers of tokens placed  (here $S_{ikj}$). In a topic model, precise allocations are unknown, but counts over fibers are observed $S_V(i_V)$ (here, $S_{i+j} = X_{ij}$). }
    \label{fig:XasContraction}
\end{figure}
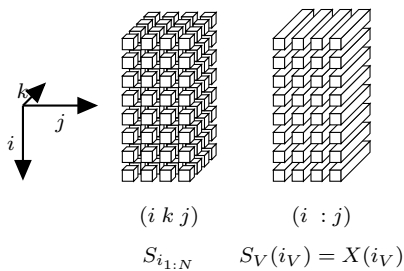

In most applications, the elements of the allocation tensor $\mathbf{S}$ are not directly observed. 
In this paper, we will focus on specific types of constraints where we assume that we observe particular contractions of $\mathbf{S}$, of form 
\begin{equation} \label{eq: X as partially observed S}
\mathbf{X}(i_\vi) = \sum_{i_{\bar{V}}}  \mathbf{S}({i_{1:N}}), \quad \forall i_{V},
\end{equation}
or, shortly,
\[
\mathbf{X} = \mathbf{S}_{V}.
\]
Here $V \subset [N]$ is the set of `visible' indices, and $\vibar = [N] \setminus \vi$ are the latent indices; see Figure \ref{fig:XasContraction} for an illustration. Many hidden variable models such as topic models and tensor factorization models have observations of this form. 

Given the observed tensor $\mathbf{X} = X$, one main inferential goal in this paper is to estimate the marginal likelihood of $X$, defined by
\begin{equation}
\mathcal{L}_{X} \equiv \Pr(\mathbf{X} = X) = \sum_{S} \pi(S) \mathbb{I}(S_{V} = X). \label{eq: data marginal likelihood}
\end{equation}
When we view the marginal likelihood as a function of a certain parameter, we will use the notation $\mathcal{L}_{X}(\cdot)$ where the value of the parameter appears in the brackets. Furthermore, we are interested in Bayesian decomposition of $X$ by targeting the posterior distribution
\begin{equation}
\pi(S | X) = \Pr(\mathbf{S} = S | \mathbf{X} = X) \propto \pi(S) \mathbb{I}(S_{V} = X). \label{eq: posterior of S given X}
\end{equation}
Many other inferential goals, such as model order estimation and tensor factorization, rely on the quantities stated above. For example, for model order estimation, one needs to calculate the marginal likelihood given the model orders to be compared. For tensor factorization, $\mathbf{S}$ provides the information for the underlying graphical model and therefore its posterior distribution given $X$ needs to be found.

\begin{ex}[\textbf{KL-NMF as a BAM}]  \label{sec: Specific Example: KL-NMF}

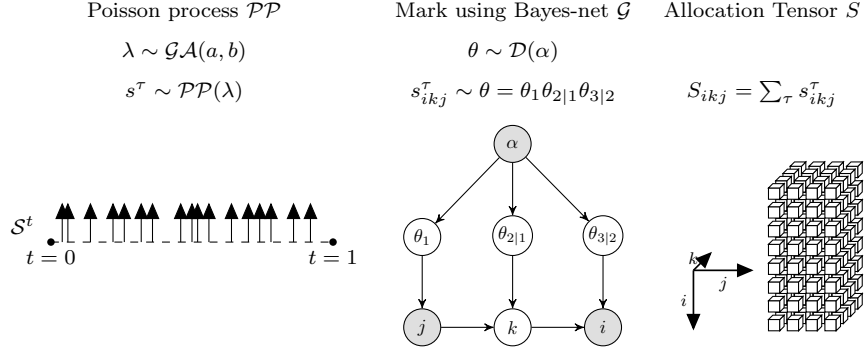
\begin{figure}
\begin{psmatrix}[arrows=->,rowsep=0.2cm,colsep=0.4cm,arrowsize=3.5pt 2]
Poisson process $\mathcal{PP}$ & Mark using Bayes-net $\mathcal{G}$ & 
\raisebox{0cm}{Allocation Tensor $S$} \\
$\lambda \sim \mathcal{GA}(a, b)$ & 
$\theta \sim  \mathcal{D}(\alpha)$  
& 
\\
$s^\tau \sim \mathcal{PP}(\lambda)$ & 
$s^\tau_{ikj} \sim \theta = \theta_{1}\theta_{2|1}\theta_{3|2} $ 
& $S_{ikj} = \sum_\tau s^\tau_{ikj}$ \\
\begin{tikzpicture}[scale=0.75]
\draw [dashed] (0,0) -- (5,0);

\node at (-0.5,0.3) {$\mathcal{S}^t$};
\node at (0,0) [circle,fill,inner sep=1pt]{};
\node [below] at (0,0) {$t=0$};
\node at (5,0) [circle,fill,inner sep=1pt]{};
\node [below] at (5,0) {$t=1$};

\draw[black,->] (0.2,0) -- ++(0,0.7);
\draw[black,->] (0.3,0) -- ++(0,0.7);
\draw[black,->] (0.7,0) -- ++(0,0.7);
\draw[black,->] (1.1,0) -- ++(0,0.7);
\draw[black,->] (1.3,0) -- ++(0,0.7);
\draw[black,->] (1.6,0) -- ++(0,0.7);
\draw[black,->] (1.8,0) -- ++(0,0.7);
\draw[black,->] (2.3,0) -- ++(0,0.7);
\draw[black,->] (2.5,0) -- ++(0,0.7);
\draw[black,->] (2.6,0) -- ++(0,0.7);
\draw[black,->] (2.8,0) -- ++(0,0.7);
\draw[black,->] (3.2,0) -- ++(0,0.7);
\draw[black,->] (3.5,0) -- ++(0,0.7);
\draw[black,->] (3.7,0) -- ++(0,0.7);
\draw[black,->] (3.9,0) -- ++(0,0.7);
\draw[black,->] (4.3,0) -- ++(0,0.7);
\draw[black,->] (4.6,0) -- ++(0,0.7);
\end{tikzpicture}
&
\raisebox{-1cm}{
\begin{tikzpicture}[>=stealth',auto,scale=0.7, every node/.style={transform shape}]        

        \node[obs]                       (j)     {$j$};
        \node[latent, right=of j]        (k)     {$k$};
        \node[obs, right=of k]           (i)     {$i$};

        \node[latent,above=of j]         (t1)     {$\theta_1$};
        \node[latent, above=of k]        (t2)     {$\theta_{2|1}$};
        \node[latent, above=of i]           (t3)     {$\theta_{3|2}$};

        \node[obs, above=of t2]           (a)     {$\alpha$};
           
        \edge {j} {k};
        \edge {k} {i};
        \edge {t1} {j};
        \edge {t2} {k};
        \edge {t3} {i};
        \edge {a}{t1} ;
        \edge {a}{t2} ;
        \edge {a}{t3} ;
\end{tikzpicture}
}
 & 
\raisebox{-0.8cm}{
\scalebox{0.8}{
\begin{tikzpicture}
\DrawCubeLabels{1}{1}{0.8}{0}{0}{0}{$i$}{$j$}{$k$}
\end{tikzpicture} 
}
\begin{tikzpicture}
\foreach \z in {0,...,3}
\foreach \y in {0,...,7}
\foreach \x in {0,...,3}
\DrawCube{0.15}{0.15}{0.15}{-\y*0.25}{\x*0.25}{\z*0.25};
\end{tikzpicture}}
\end{psmatrix}
\caption{A schematic description of BAM for KL-NMF and LDA. Increments $s^\tau$ of the homogeneous Poisson process are marked using a Bayesian network $\mathcal G$ $j\rightarrow k \rightarrow i$ to obtain the 'marked' token $s^\tau_{ikj}$ and placed into the allocation tensor. The number of tokens $S_{ikj}$ at the box corresponding to the color $(i,k,j)$ are Poisson distributed with  $S_{ikj}  \sim  \mathcal{PO}\left(\lambda \theta_{1}\theta_{2|1}\theta_{3|2}\right)$. For KL-NMF, the counts at each box are not directly observed, instead we observe a contraction of the allocation tensor $S$ over the latent index $k$ as $X_{ij} = \sum_k S_{ikj}$. For LDA, we have the same model, however, the total number of tokens is assumed to be known.
}
\label{fig:bam_klnmf}
\end{figure}

We first consider the following specific undirected graphical model $i \;\text{--}\; k \;\text{--}\; j$ that corresponds to the KL-NMF model. This simple graph has three equivalent Bayesian network parametrizations as $j \rightarrow k \rightarrow i$, $j \leftarrow k \rightarrow i$ and $j \leftarrow k \leftarrow i$, e.g., see \citep{Murphy2012}. To have a concrete example, we will focus on an allocation process that assigns tokens generated from the base process to a document $j$, then conditioned on the document on a topic $k$ and finally conditioned on the topic to a word $i$, as in $\text{doc} \rightarrow \text{topic} \rightarrow \text{word}$, the other graph structures would be equivalent.  To complete the model specification, we let $(i_1, i_2, i_3) \equiv (j,k,i)$. This implies that we have $\pa{1} = \emptyset, \fa{1} = \{1\}$, $\pa{2} = \{ 1\}, \fa{2} = \{1,2\}$ and $\pa{3} = \{2\}, \fa{3} = \{2, 3\}$. In a topic model, the topic assignments are hidden, and we observe only the word-document pairs $i,j$ so the visible set is $V = \{1,3\}$. The specialized generative process for this model is
\begin{align*}
& \lambda \sim  \mathcal{GA}(a, b) \\
& \theta_{1}(:) \sim  \mathcal{D}(\alpha_{1}(:)), \quad\quad \theta_{2|1}(:,j)  \sim  \mathcal{D}(\alpha_{2, 1}(:,j)), \quad \quad \theta_{3|2}(:,k)  \sim  \mathcal{D}(\alpha_{3, 2}(:,k) )  \\ 
&  S_{ikj}  \sim  \mathcal{PO}(\lambda \theta_{3|2}(i,k) \theta_{2|1}(k,j) \theta_{1}(j) ),\\
& X_{ij}  =  \sum_{k = 1}^{K} S_{ikj}.
\end{align*}
This model is illustrated in Figure~\ref{fig:bam_klnmf}. Now, if we define the matrices $W_{ik} \equiv \theta_{3|2}(i,k)$ and $H_{kj} \equiv \lambda \theta_{2|1}(k,j) \theta_{1}(j)$ and sum over the allocation tensor $S$, we can arrive at the following model 
\begin{align}
H_{kj} & \sim  \mathcal{GA}(\alpha_{2, 1}(k, j) , b), & 
W_{:k} & \sim  \mathcal{D}( \alpha_{3, 2}(:, k) ),  &  
X_{ij} & \sim  \mathcal{PO} \left( \sum_{k = 1}^{K} W_{ik} H_{kj} \right). \label{eq:klnmf}
\end{align}
Apart from the specific restricted choice of prior parameters for factors, the generative model is closely related to the Bayesian KL-NMF or Poisson factorization \citep{Cemgil2009, Paisley2014}. The formulation of KL-NMF in \citet{Cemgil2009} proposes Gamma priors on both $W$ and $H$, thereby introducing an extra scaling redundancy which prohibits integrating out $W$ and $H$ exactly, hence leads to a subtle problem in Bayesian model selection. 
\end{ex}

\subsection{The Marginal Allocation Probability of BAM} \label{sec: Marginal Allocation Model}
In the sequel, our aim will be deriving the marginal distribution of the allocation tensor $\mathbf{S} = \mathbf{S}^{T}$, the number of tokens accumulated at step $\tau=T$, or equivalently $t=1$. We will refer to $\pi(S)$ as the \emph{marginal allocation probability}. 

We can obtain an explicit analytical expression for $\pi(S)$ with the help of conjugacy. We can integrate out analytically the thinning probabilities $\Theta$ and intensity $\lambda$. The closed form conditional distribution is available as
\begin{align} 
& \pi(\lambda, \Theta| S) = \mathcal{GA}(\lambda; a + S_+, b+1) \nonumber \\
& \quad \times \prod_{n=1}^{N} \prod_{i_{\pa{n}}} \mathcal{D}(\theta_{n | \pa{n}}(:, i_{\pa{n}}); \alpha_{\fa{n}}(:, i_{\pa{n}}) + S_{\fa{n}}(:, i_{\pa{n}})). \label{eq: post of lamda and theta given S}
\end{align}
In \eqref{eq: post of lamda and theta given S}, we have used the definition
\[
S_{\fa{n}}(i_n, i_{\pa{n}}) = \sum_{i_{\overline{\fa{n}}}} S({i_{1:N}})
\]
in analogy with \eqref{eq:margal}, and $S_{+}$ is defined in \eqref{eq: Splus}. (Note that, surprisingly the posterior remains  factorized as $\pi(\lambda, \Theta| S) = \pi(\lambda| S) \pi(\Theta| S)$.) The expression for the marginal allocation probability is obtained using \eqref{eq: joint of S theta lambda} and \eqref{eq: post of lamda and theta given S} as
\begin{equation}
\pi(S)  =  \frac{b^{a}}{(b+1)^{a+ S_+}}\frac{\Gamma(a + S_+)}{\Gamma(a)} \left( \prod_{n = 1}^{N} \frac{B_n( \alpha_{\fa{n}} + S_{\fa{n}} )}{B_n( \alpha_{\fa{n}}  )}  \right) \frac{1}{\prod_{i_{1:N}} S({i_{1:N}})!}, \label{eq:pS1}
\end{equation}
where, to simplify our notation, we define a multivariate beta function of a tensor argument as follows
\begin{equation} \label{eq: multivariate beta function}
B_n(Z_{\fa{n}}) = \prod_{i_\pa{n}}\left( \frac{\prod_{i_n} \Gamma(Z_{\fa{n}}({i_n, i_{\pa{n}}}))}{\Gamma(\sum_{i_n} Z_{\fa{n}}({i_n, i_{\pa{n}}}))}\right).     
\end{equation}
Here $Z_{\fa{n}}$ is an object of the same shape signature as $S_{\fa{n}}$, hence
$B_n$ is a function that computes first a nonlinear contraction of a tensor with indices $i_{\fa{n}}$ over index $i_n$ to result in a reduced tensor over indices $i_\pa{n}$ and multiplies all the entries of this tensor. We will denote the usual multivariate beta function as $\text{Beta}(z) = \prod_i \Gamma(z_i)/\Gamma(\sum_i z_i)$.

\subsubsection{The marginal allocation probability conditioned on its sum} \label{sec: Marginal allocation probability conditioned to its sum}
In many instances, the sum $S_{+}$ is observable, such as via a partially observed $X$ as in \eqref{eq: X as partially observed S}, while $S$ itself remains unknown. For such cases, it is informative to look at the probability of the marginal allocation conditioned on the sum $S_{+}$. For that, consider the allocation model described at the beginning of Section \ref{sec: Allocation Model} and let $\pi_{\tau}$ denote the marginal distribution of $\mathbf{S}^{\tau}$,
\begin{equation} \label{eq: marginal of S conditioned on the sum}
\pi_{\tau}(S^{\tau})  = \Pr(\mathbf{S}^{\tau} = S^{\tau}), \quad t \geq 1.
\end{equation}
Recalling $\mathbf{S} = \mathbf{S}^{T}$ and $T = \mathbf{S}_{+}$, one can decompose \eqref{eq:pS1} as
\begin{align}
\pi(S) = \Pr(\mathbf{S}^{T} = S) = \Pr(T = S_+) \pi_{S_{+}}(S). \label{eq: decomposition of pS}
\end{align}
The first factor in \eqref{eq: decomposition of pS} is the marginal probability of $T$, obtained after integrating over the latent intensity parameter $\lambda$ as
\begin{align} 
\Pr(T = S_+)  & =  \frac{b^{a}}{(b+1)^{a+ S_+}}\frac{\Gamma(a + S_+)}{\Gamma(a)\;S_+!}  \nonumber \\
&  = \frac{\Gamma(a + S_+)}{\Gamma(a)\Gamma(S_+ + 1)} \left(\frac{b}{b+1}\right)^a \left(\frac{1}{b+1}\right)^{S_+}. \label{eq:pS+}
\end{align}
This is a negative Binomial distribution and can be interpreted as observing $S_+$ `successes', each with probability $1/(b+1)$ before we have observed $a$ `failures', each with probability $b/(b+1)$. The second factor in \eqref{eq: decomposition of pS} follows from \eqref{eq:pS1} and \eqref{eq:pS+} as
\begin{align} 
\pi_{S_{+}}(S) = \left( \prod_{n = 1}^{N} \frac{B_{n}( \alpha_{\fa{n}} + S_{\fa{n}} )}{B_{n}( \alpha_{\fa{n}}  )} \right) \binom{S_+}{S}. \label{eq:pS2}
\end{align}
In \eqref{eq:pS2}, the first factor is the score of a Bayesian network when complete allocations $S$ are observed \citep{Cooper1992,Murphy2012} and the second factor is the multinomial coefficient 
\[
\binom{S_+}{S} = \frac{S_+!}{\prod_{i_{1:N}} S({i_{1:N}})!}
\]
that counts the multiplicity of how many different ways $S_+$ tokens could have been distributed to $|I_1|\times|I_2|\dots |I_N|$ cells such that the cell $i_{1:N}$ has $S({i_{1:N}})$ tokens allocated to it. 

 Note that for large $z_+ = \sum_i z_i$, we have $ -H(z/z_+) \approx (\log {\text{Beta}}(z))/z_+ $ where $H(p)$ is the entropy of a discrete distribution $p$, see, e.g., \citet{Csiszar2004}. We have similarly $H_n(Z_{\fa{n}}/Z_+) \approx (\log B_n(Z_{\fa{n}}))/Z_+ $. Hence, 
\begin{align*}
\log \pi_{S_{+}}(S) & \approx (S_+) H(S/S_+) + \alpha_+ \sum_{n = 1}^{N}  H_n( \alpha_{\fa{n}}/\alpha_+  ) \\
& \quad - (\alpha_+ + S_+)\sum_{n = 1}^{N}  H_n((\alpha_{\fa{n}} + S_{\fa{n}})/(a_+ + S_+)),
\end{align*}
which suggests that the marginal allocation probability is high when the tokens are distributed as uniformly as possible over $S$ corresponding to high entropy $H(S/S_+)$ while the entropy of the marking distribution encoded by the Bayesian network $\mathcal{G}$ has a low entropy. 

\subsection{The Marginal Allocation Probability for the KL-NMF}
The specific form of \eqref{eq:pS1} for the KL-NMF provides further insight about the marginal allocation probability. Assuming that all entries of $X$ are observed, some of the terms in \eqref{eq:pS1} become merely constants. We group the terms that are constant when $X$ is fixed as $\mathcal{C}_{f}(S)$. These terms only depend on fixed hyperparameters $\alpha$, $a$, $b$ or can be directly calculated from observed data $X$. The remaining $S$ dependent terms $\mathcal{C}_{d}(S)$ are given as
\begin{eqnarray}
\log \pi(S) & = & \mathcal{C}_{f}(S) + \mathcal{C}_{d}(S), \nonumber \\
\mathcal{C}_{d}(S) & = & + \sum_{i, k} \log \Gamma( \alpha_{ik+} + S_{ik+}) +\sum_{k, j} \log \Gamma( \alpha_{+kj} + S_{+kj}) \label{eq:concentrate}\\
& & - \sum_k \log{\Gamma(\alpha_{+k+} + S_{+k+})}  - \sum_{i, k, j} \log \Gamma( S_{ikj}+1).  \label{eq:disperse}
\end{eqnarray}
To understand $\mathcal{C}_{d}(S)$, we consider Stirling's formula $\log \Gamma (s+1) \sim s\log s - s + O(\log s)$ so terms such as $-\sum_{i=1}^N \log \Gamma(s_i) \approx -\sum_{i=1}^N s_i\log s_i + x$ where $\sum_i s_i = x$ can be interpreted as an entropy of a mass function defined on $N$  cells where $s_i$ is the mass allocated to cell $i$. The entropy increases if the total mass is distributed evenly across cells and decreases if it concentrates only to a few cells. We thus imagine a physical system where a total of $S_+$ balls are placed into $I \times K \times J$ bins, where there must be exactly $X_{ij}$ balls in the cells $(i1j), (i2j), \dots, (iKj)$, that we will refer as a fiber $(i\mycolon j)$ (as $S_{i+j} = X_{ij}$). We can think of each tensor $S$ as a mass function, where $S_{ikj}$ counts the number of balls being placed into bin $(i\;k\;j)$ and $\pi(S)$ is a distribution over these mass functions. The individual bins $(i\;k\;j)$ and the fibers $(i\;:k)$ with fixed marginal sum are depicted as in the  `cubic' plots in Figure.\ref{fig:XasContraction}.

We can see that the expression has four terms that are competing with each other: terms on line \eqref{eq:concentrate} that try to make the marginal sums $S_{ik+}$ and $S_{+kj}$ as concentrated as possible to a few cells while the terms in \eqref{eq:disperse} force $S_{+k+}$ and $S_{ikj}$ to be as even as possible. Following the physical systems analogy, $\pi(S)$ assigns an higher probability to configurations, where balls are aligned as much as possible across $(ik\mycolon)$ and $(\mycolon k\;j)$, but are still evenly distributed among slices of form $(\mycolon\;k\mycolon)$ and to individual bins $(i\;k\;j)$; see Figure \ref{fig:visterms}. This observation also explains the clustering behavior of KL-NMF from a different perspective. It is often reported in the literature that NMF has, empirically and theoretically, a representation by parts property, that is, the matrices $W^*$ or $H^*$ obtained as the solution to the minimization problem in \eqref{eq:nmf-original} tend to be usually sparse with occasional large entries. Remembering that expectations under $p(W,H|S^*)$ are $\E{W_{:k}} \propto S_{:+k}$ and $\E{H_{:j}} \propto S_{+j:}$, we see that this is what the terms in \eqref{eq:concentrate} enforce. Yet, this property is not directly visible from the original generative model \eqref{eq:KLNMFx} but is more transparent from the alternative factorization. In Section \ref{sec: Inference by Sequential Monte Carlo}, we will also give a dynamic interpretation of this clustering behaviour.

\subsection{Decomposing the Marginal Allocation Probability}
\begin{figure}
    \centering
\begin{center}
\begin{psmatrix}[rowsep=0.2cm,nodesep=.05cm,colsep=0.3cm]
\begin{tikzpicture}
\DrawCubeLabels{1}{1}{0.8}{0}{0}{0}{$i$}{$j$}{$k$}
\end{tikzpicture} 
&
\begin{tikzpicture}
\foreach \z in {0,...,3}
\foreach \y in {0,...,7}
\DrawCube{0.15}{1}{0.15}{2-\y*0.25}{0}{\z*0.25};
\end{tikzpicture}
&
\begin{tikzpicture}
\foreach \x in {0,...,3}
\foreach \z in {0,...,3}
\DrawCube{2}{0.15}{0.15}{0}{\x*0.25}{\z*0.25};
\end{tikzpicture}
&
\begin{tikzpicture}
\foreach \z in {0,...,3}
\DrawCube{2}{1}{0.15}{0}{0}{\z*0.25};
\end{tikzpicture}
&
\begin{tikzpicture}
\foreach \z in {0,...,3}
\foreach \y in {0,...,7}
\foreach \x in {0,...,3}
\DrawCube{0.15}{0.15}{0.15}{-\y*0.25}{\x*0.25}{\z*0.25};
\end{tikzpicture}\\
& $(ik\mycolon)$ & $(\mycolon k\;j)$ & $(\mycolon k\mycolon)$ & $(i\;k\;j)$
\\
& $S_{ik+}$ & $S_{+kj}$ & $S_{+k+}$ & $S_{ikj}$
\end{psmatrix}
\end{center}
    \caption{Visualization of the clique and separator domains of the marginal allocation probability $\pi(S)$ for KL-NMF/LDA.}
    \label{fig:visterms}
\end{figure}
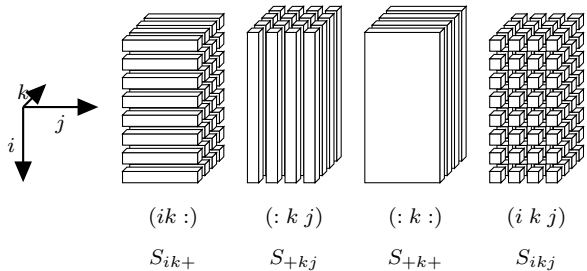

An important observation in the previous section is that the marginal allocation probability $\pi(S)$ inherits the same factorization properties of the underlying graphical model $\mathcal{G}$, as it can also be written as a product of a function of the clique marginals ($S_{ik+}$ and $S_{+kj}$) divided by a function of the separator marginal $S_{+k+}$
\[
\pi(S) \propto \frac{\prod_{ik} \Gamma(\alpha_{ik+} + S_{ik+}) \prod_{kj} \Gamma(\alpha_{+kj} + S_{+kj})}{\prod_{k}\Gamma(\alpha_{+k+} + S_{+k+})} \frac{1}{\prod_{ikj} \Gamma(S_{ikj} + 1)}.
\]
This property, that the parameter posterior has the analogous factorization is named as a hyper Markov law \citep{Dawid1993}. A direct consequence of this factorization is that $\log \pi(S)$ can be written as a sum of convex and concave terms, corresponding to clique and separator potentials respectively. The last term corresponding to the base measure also appears as a concave term but it would be cancelled out when comparing different models. The presence of convex terms renders the maximization problem difficult in general. The terms are pictorially illustrated in Figure \ref{fig:visterms}.

For a general graph $\mathcal{G}$, the form of $\pi(S)$ may not be easy to interpret. However, the junction tree factorization for decomposable graphs provides a nice interpretation  
\begin{equation}
\pi(S) \propto \frac{\prod_{q=1}^{|\Cliques|} \prod_{i_{C_q}} \Gamma(\alpha_{C_q}(i_{C_q}) + S_{C_q}(i_{C_q}))}{\prod_{q=1}^{|\Separators|} \prod_{i_{D_q}} \Gamma(\alpha_{D_q}(i_{D_q}) + S_{D_q}(i_{D_q}))} \frac{1}{\prod_{i_{1:N}} \Gamma(S(i_{1:N}) + 1)}. \label{eq:pS}
\end{equation}
where, similarly, the proportionality constant can be solely determined from $X = S_{V}$. Consequently, as in the special KL-NMF case, the general $\log \pi(S)$ can also be expressed as a sum of convex terms corresponding to the clique potentials and concave terms corresponding to the separator potentials. Not surprisingly, the presence of convex terms renders this problem also a hard optimization problem in general. Here is an interesting trade-off: the factorization structure renders maximization difficult but the model can be expressed with far less parameters. It is possible getting rid of separators by clumping more cliques together but this implies that that the computational requirement increases, in the extreme case of a complete graph, all convex terms disappear but the model becomes intractable to compute \citep{Wainwright2007}. 

In the following section, we will give a specific example of our rather general and abstract construction to illustrate that KL-NMF and LDA, upto the choice of priors, are equivalent to a particular allocation model.

\subsection{KL-NMF and LDA Equivalence} \label{sec: KL-NMF and LDA equivalence}
In this section, we will establish the equivalence of KL-NMF and LDA, we will first state the LDA model and construct an equivalent allocation model. In a sense, this equivalence is analogous to the close connection of the joint distribution of Bernoulli random variables and their sum, a Binomial random variable. LDA is a generative model for the token allocations, conditioned on the total number $S_+$ tokens while KL-NMF is a joint distribution of both. 

To follow the derivation of LDA more familiar in the literature, we first define the following indicators: $d_{j\tau}$, $z_{k\tau }$ and $w_{i\tau }$ that encode the events that token $\tau \in [S_+]$ selects $j$'th document, $k$'th topic and $i$'th word respectively. Then we define a hierarchical generative model
\begin{align*}
\chi_{:} & \sim \mathcal{D}(\eta_\chi), &\vartheta_{:j} & \sim  \mathcal{D}(\eta_\vartheta), &\beta_{:k} & \sim \mathcal{D}(\eta_\beta), \\
d_{: \tau} & \sim \mathcal{M}(\chi_{:}, 1), & z_{: \tau}|d_{:\tau} & \sim \prod_{j = 1}^{J}\mathcal{M}(\vartheta_{:j}, 1)^{d_{j\tau}}, & w_{: \tau}|z_{: \tau} & \sim \prod_{k = 1}^{K} \mathcal{M}(\beta_{:k}, 1)^{z_{k\tau}}. 
\end{align*}
The inference goal is, given $d$ and $w$ estimating $z, \beta$ and $\vartheta$. The variables $\chi$ and $d$ are omitted in the original formulation of LDA as these can be directly estimated from data, when all the tokens are observed, i.e., when there are no missing values. We consider the joint indicator $s^\tau_{ikj} = d_{j \tau}z_{k \tau}w_{i \tau}$ and define $S_{ikj} = \sum_\tau s^\tau_{ikj}$. As $s^\tau$ is multinomial with $\mathcal{M}(\theta, 1)$, each cell having probability $\theta_{ikj} \equiv \beta_{ik}\vartheta_{kj}\chi_{j}$, the allocation tensor $S$, being the sum of multinomial random variables with the same probability is also multinomial with $\mathcal{M}(\theta, S_+)$. To see the connection with the allocation model, remember that the multinomial distribution can be characterized as the posterior distribution of independent Poisson random variables, conditioned on their sum \citep{Kingman1993}. This property is succinctly summarized below using a Kronecker $\delta$ as   
\begin{eqnarray*}
\delta(S_+ - \sum_{i,k,j} S_{ikj}) \prod_{i,k,j} \mathcal{PO}(S_{ikj}; \theta_{ikj} \lambda) & = & \mathcal{M}(S; \theta, S_+) \mathcal{PO}(S_+; \lambda).
\end{eqnarray*}
The original LDA formulation also omits $S_+$, the total number of tokens; as there are no missing values. This suggests that if we would condition the following allocation model also on $S_+$, we will exactly get the LDA 
\begin{align*}
\chi_{:} & \sim \mathcal{D}(\eta_\chi), & \vartheta_{:j} & \sim  \mathcal{D}(\eta_\vartheta), & \beta_{:k} & \sim \mathcal{D}(\eta_\beta), \\
\lambda & \sim  \mathcal{GA}(a, b), & S_{ikj} & \sim  \mathcal{PO}( \lambda \beta_{ik}\vartheta_{kj}\chi_{j} ) &
X_{ij} & = \sum_{k = 1}^{K} S_{ikj}.
\end{align*}
We have thus shown that both LDA and Bayesian KL-NMF are equivalent to Bayesian learning of a graphical model $i \;\text{--}\; k \;\text{--}\; j$ where $k$ is latent. This observation suggests that there are one-to-one correspondences between other structured topic models, NTF and graphical models. Note that the link between KL-NMF and LDA has long been acknowledged and the literature is summarized in \citet{Buntine2006}. We conclude this section with an illustrative example of parameter consistency problem when estimating the rank of a decomposition, i.e., the cardinality of a latent index.

\subsection{On Hyperparameters of BAM} \label{sec: Hyperparameters}
The Poisson process interpretation guides us here in defining natural choices for model parameters: the (prior) expectation of $\lambda$, denoted as $\E{\lambda} = a/b$, is also the expected number of tokens observed until time $t=1$. Let 
\begin{equation}
\mathbf{S}_{+} \equiv \sum_{i_{1:N}} \mathbf{S}({i_{1:N}}) \label{eq: Splus}
\end{equation}
(with realizations $S_{+}$). Given an observed tensor $X$, $S_{+}$ is fully observable and we could choose the scale parameter as $b \approx a/S_+$, following an empirical Bayesian approach. 

When there are missing values in the observed tensor $X$, the exact $S_+$ is also unknown. In this case, $b$ should be integrated out, but we do not investigate this possibility here. In practice, it is also possible to get a rough estimate from data as an average.  

The key parameter that defines the behaviour of the model is $\alpha$, that describes our \emph{a-priori} belief, how the tokens will be distributed on $\mathbf{S}$.  In practice, we expect $\mathbf{S}$ to be sparse, i.e., tokens will accumulate in relatively few cells, however without \emph{a-priori} preference to a particular group. Here, a natural choice is a flat prior 
\[
\alpha({i_{1:N}}) = \frac{a }{\prod_{n=1}^N I_n}
\]
with $a>0$ being a sparsity parameter having typical values in the range of $0.05$ or $0.5$, also known as a BDeu prior. Here, BDeu is an abbreviation for Bayesian Dirichlet (likelihood) Equivalent Uniform prior choice \citep{Buntine91,Heckerman1995}. Fixing $\alpha$, we can also see that $a$ is also a shape parameter with $a=\sum_{i_{1:N}} \alpha(i_{1:N})$. The choice of $a$ is critical \citep{Steck02onthe}.

\begin{ex}[\textbf{BDeu Priors}]  \label{ex: Example (Consistent Priors)}
In this example, our goal is to illustrate with a simple example how choosing the prior parameter $\alpha$ effects model scoring and how an inconsistent choice may lead to misleading conclusions even in very simple cases, a point that seems to have been neglected in the topic modeling literature. 

Suppose we observe the following contingency table where rows and columns are indexed by $i$ and $j$ respectively 
\begin{align*}
S & = \left(\begin{array}{ccc} 2 & 1 \\ 0 &  1 \end{array} \right),
\end{align*}
and our goal is to decide if $S$ is a draw from an independent or a dependent model, that is to decide if the tokens are allocated according to the allocation schema $i \;\; j$  versus $i \leftarrow j$ (or $i \rightarrow j$).

As only $S_+ = 4$ tokens are observed, intuition suggests that the independent model maybe preferable over the one with dependence, but this behaviour is closely related to the choice of the hyperparameters. In a Bayesian model selection framework, one may choose taking a flat prior for all $i,j$ as $\alpha(i,j)  = a/4$. For the case $a=1, b=1$, we obtain for the independent model 
\[
\pi_{i \perp j}(S|\alpha_1(i) = a/2, \alpha_2(j) = a/2) = \exp(-7.977);
\]
and for the dependent model 
\begin{eqnarray*}
\pi_{i \leftarrow j}(S|\alpha_2(j) = a/2, \alpha_{1, 2}(i,j) = a/4) & = & \exp(-8.094), \\
\pi_{i \rightarrow j}(S|\alpha_1(i) = a/2, \alpha_{2, 1}(j,i) = a/4) & = & \exp(-8.094), \\
\pi_{i - j}(S|\alpha(i, j) = a/4) & = & \exp(-8.094).
\end{eqnarray*}
Here, the independent model is slightly preferred. Also, the numeric results are identical, as they should be, for all Markov equivalent dependent models. 
However, when consistency is not respected, and we choose all Dirichlet priors to be flat, say all cells as $a/4$, we would obtain for the independent model 
\[
\pi_{i \perp j}(S|\alpha_1(i) = a/4, \alpha_2(j) = a/4) = \exp(-8.808);
\]
and for the dependent model 
\begin{eqnarray*}
\pi_{i \rightarrow j}(S|\alpha_{1}(i) = a/4, \alpha_{2, 1}({j,i}) = a/4) & = & \exp(-8.472), \\
\pi_{i \leftarrow j}(S|\alpha_{2}(j) = a/4, \alpha_{1, 2}(i,j) = a/4) & = & \exp(-8.549).
\end{eqnarray*}
In the inconsistent case, not only the two alternative factorizations give different results; the independent model achieves a lower marginal allocation probability than the dependent model. 
\end{ex}

\begin{ex}[{\bf Dependence of $\pi(S)$ on the prior equivalent sample size $a$}]
In this example, we will illustrate the behaviour of the marginal allocation probability $\pi(S)$ for different equivalent sample size parameters $a$. We calculate $\pi(S)$ exactly for the NMF model in the range of $a$ from $10^1$ to $10^{-10}$ for a toy matrix $X^{(1)}$, given below. This range has been chosen to demonstrate the gradual change in the probability values of $\pi(S)$ due to the transition from large $a$ to small $a$.
\begin{align*}
X^{(1)} & = \left(\begin{array}{cccc} 
    2 & 1 & 1 & 0 \\
    0 & 0 & 1 & 2 \\
    0 & 0 & 1 & 1 
    \end{array} \right)
\end{align*}
\begin{figure}[htbp]
	\centering
	\includegraphics[width=1.\textwidth]{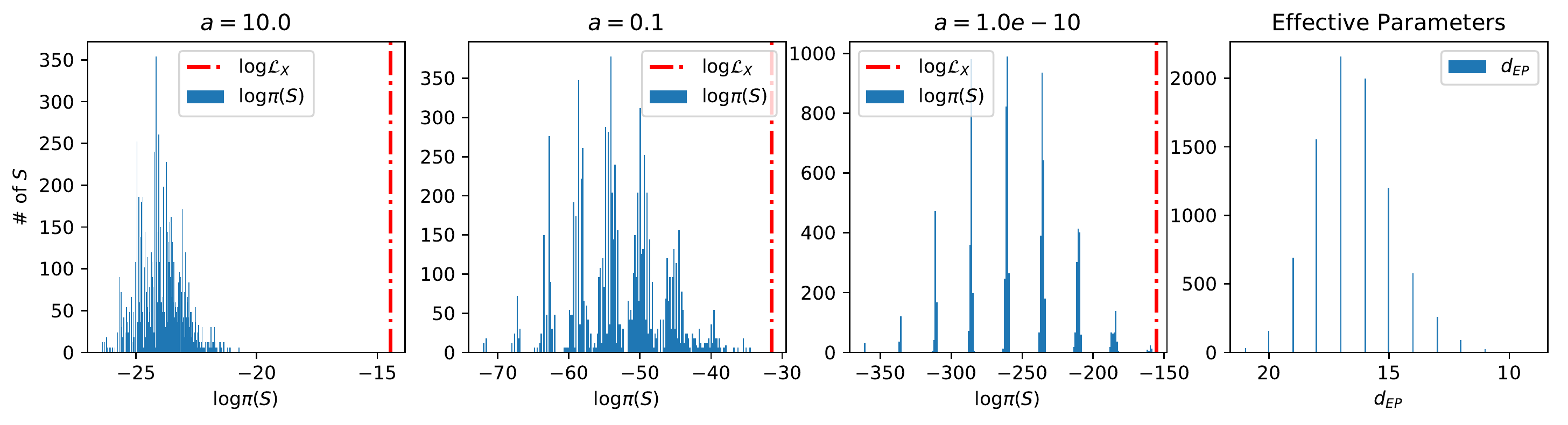}
	\caption{Histogram of the marginal allocation probability for different tensors with marginal count matrix $X^{(1)}$. The marginal likelihood of $X^{(1)}$ is shown with a red vertical line. Histogram of $d_{EP}$ is supplied for comparison.}
	\label{fig:likelihood-histogram}
\end{figure}

In this example, we enumerate all tensors with non-negative integer entries, and with the hidden index dimension $K=3$, that have the marginal count matrix $X^{(1)}$, i.e., we consider all $S$ such that $S_{i+j} = X^{(1)}_{ij}$. For each such tensor $S$, we calculate $\pi(S)$ and show the histogram of those values in Figure \ref{fig:likelihood-histogram}. The effect of $a$ is quite dramatic, especially when $a \rightarrow 0$ the values that the marginal allocation probability can take is confined on a grid (see Figure \ref{fig:likelihood-histogram}). This gives a clue about the nature of the distribution of the marginal allocation $S$, where we have discrete levels of probability values and each admissible tensor $S$ is assigned to one of these values.
\end{ex}

The behaviour of the marginal in the limit $a \rightarrow 0$ is perhaps surprising. In fact, \citet{Steck02onthe} have shown that the log-probability of the tensor $S$ becomes independent of the counts in particular entries, and dependent only on the number of different configurations in the data, defined as the effective number of parameters:
\begin{equation*}
    d_{EP}(S) = \sum_{n=1}^N \biggl(\sum_{i_\fa{n}} \ind{S_{\fa{n}} (i_\fa{n})>0} - \sum_{i_{\pa{n}}} \ind{S_{\pa{n}} (i_\pa{n})>0} \biggr).
\end{equation*} 
In the regime where $a$ is sufficiently small, the $d_{EP}(S)$ becomes the sole determiner of the probability of a data tensor. As seen in the equation above, this value is only dependent on the structure of the graphical model, and as \citet{Steck02onthe} shows, it determines the probability as specified below:
\begin{eqnarray*}
    \log \Pr(\mathbf{S}^{1:T} = S^{1:T}) & \approx & d_{EP}(S^T) \log a, \\
    \log \pi_{S_{+}}(S) & \approx & d_{EP}(S) \log a + \log\binom{S_+}{S}.
\end{eqnarray*}
This is corroborated by the our findings in Figure \ref{fig:likelihood-histogram} where, for example, the clear modes observed in the parameter setting $a = 10^{-10}$ are at distance from their neighbors by $\log a$.

\begin{figure}
\centering
\begin{subfigure}[t]{0.31\textwidth}
    \centering
    \begin{tikzpicture}[>=stealth',auto,scale=0.55, every node/.style={transform shape}]
        
        \node[obs]                       (j)     {\LARGE $i$};
        \node[latent, right=of j]        (k)     {\LARGE $k$};
        \node[obs, right=of k] 
        (i)     {\LARGE $j$};
           
        \edge {j} {k};
        \edge {k} {i};
    \end{tikzpicture}
    \caption{KL-NMF, LDA}
    \label{fig:KL-NMF}
\end{subfigure}
~
\begin{subfigure}[t]{0.31\textwidth}
    \centering
    \begin{tikzpicture}[>=stealth',auto,scale=0.55, every node/.style={transform shape}]
        
        \node[obs]                       (j)     {\LARGE $j$};
        \node[latent, below=of j]        (k1)     {\LARGE $k_1$};
        \node[latent, right=of k1]        (k2)     {\LARGE $k_2$};
        \node[latent, right=of k2]        (k3)     {\LARGE $k_3$};
        \node[obs, above=of k3]        (i)     {\LARGE $i$};
           
        \edge {j} {k1};
        \edge {k1} {k2};
        \edge {k2} {k3};
        \edge {k3} {i};
    \end{tikzpicture}
    \caption{Pachinko allocation}
\end{subfigure}
~
\begin{subfigure}[t]{0.31\textwidth}
    \centering
    \begin{tikzpicture}[>=stealth',auto,scale=0.55, every node/.style={transform shape}]
        
        \node[obs]                       (i1)     {\LARGE $i_1$};
        \node[latent, below=of i1]        (k1)     {\LARGE $k_1$};
        \node[obs, right=of k1]        (s)     {\LARGE $s$};
        \node[latent, right=of s]        (k2)     {\LARGE $k_2$};
        \node[obs, above=of k2]          (i2)     {\LARGE $i_2$};
           
        \edge {i1} {k1};
        \edge {k1} {s};
        \edge {i2} {k2};
        \edge {k2} {s};
    \end{tikzpicture}
    \caption{Mixed membership stochastic blockmodel}
\end{subfigure}
\\
\begin{subfigure}[t]{0.31\textwidth}
    \centering
    \begin{tikzpicture}[>=stealth',auto,scale=0.55, every node/.style={transform shape}]
        \node[obs]                       (j)     {\LARGE $j$};
        \node[latent, right=of j]        (r)     {\LARGE $r$};
        \node[obs, right=of r]        (i)     {\LARGE $i$};
        \node[latent, below=of r]        (k)     {\LARGE $k$};
           
        \edge {j} {k};
        \edge {k} {r};
        \edge {j} {r};
        \edge {k} {i};
        \edge {r} {i};
    \end{tikzpicture}
    \caption{Sum conditioned Poisson factorization}
\end{subfigure}
~
\begin{subfigure}[t]{0.31\textwidth}
    \centering
    \begin{tikzpicture}[>=stealth',auto,scale=0.55, every node/.style={transform shape}]
        \node[latent]        (r)     {\LARGE $r$};
        
        \node[obs, below=of r]        (i2)     {\LARGE $i_2$};
        \node[obs, left=of i2]         (i1)     {\LARGE $i_1$};
        \node[obs, right=of i2]        (i3)     {\LARGE $i_3$};
        \edge {r} {i1};
        \edge {r} {i2};
        \edge {r} {i3};        
    \end{tikzpicture}
    \caption{Canonical Polyadic \\ decomposition}
    \label{fig:naive-bayes}
\end{subfigure}
~
\begin{subfigure}[t]{0.31\textwidth}
    \centering
    \begin{tikzpicture}[>=stealth',auto,scale=0.55, every node/.style={transform shape}]
        
        \node[latent]        (r1)     {\LARGE $r_1$};
        \node[latent, right=of r1]        (r2)     {\LARGE $r_2$};
        \node[latent, right=of r2]        (r3)     {\LARGE $r_3$};
        
        \node[obs, below=of r1]        (i1)     {\LARGE $i_1$};
        \node[obs, below=of r2]        (i2)     {\LARGE $i_2$};
        \node[obs, below=of r3]        (i3)     {\LARGE $i_3$};
           
        \edge {r1} {r2,i1};
        \edge {r2} {r3,i2};
        \edge {r3} {i3};
        
        \path[->,every node/.style={font=\sffamily\small}] (r1) 
        edge[bend left] node [left] {} (r3);

    \end{tikzpicture}
    \caption{Tucker decomposition}
\end{subfigure}
\caption{Some Nonnegative matrix/tensor Factorization Models and Topic Models expressed as instances of BAM.}
    \label{fig:equivalent models}
\end{figure}
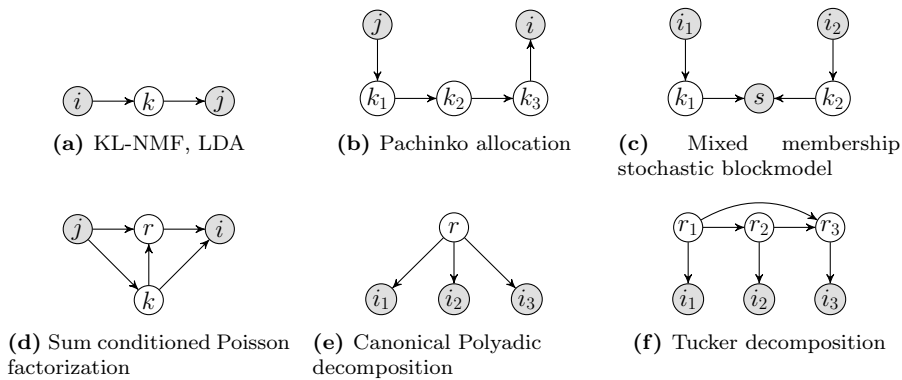

\begin{ex}[{\bf Posterior Distribution of the number of tokens  $S_+$}] \label{sec: Distribution of S+}

\begin{figure}[thbp]
	\centering
	\begin{subfigure}[t]{0.47\textwidth}
		\includegraphics[width=1.\textwidth]{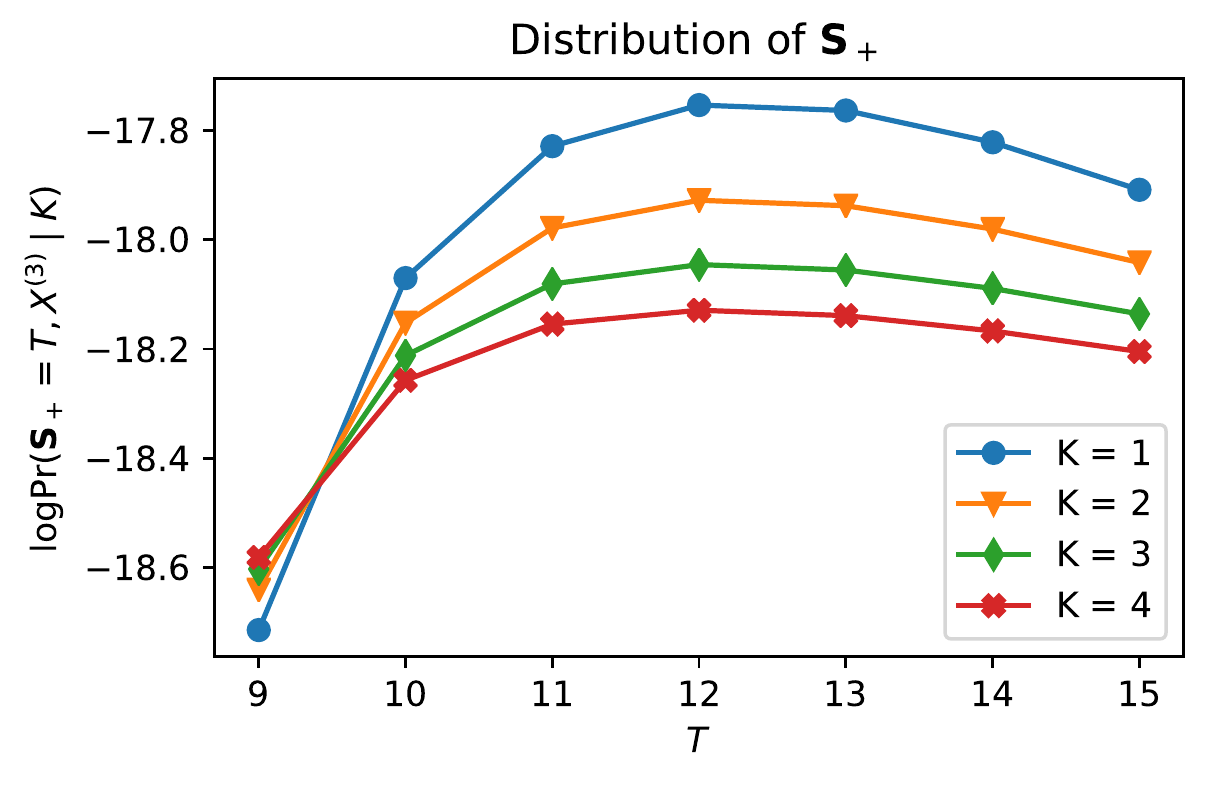}
		\caption{Unnormalized posterior distribution of $S_+$ given the matrix $X^{(3)}$.} 
	\end{subfigure}	~
	\begin{subfigure}[t]{0.47\textwidth}
		\includegraphics[width=1.\textwidth]{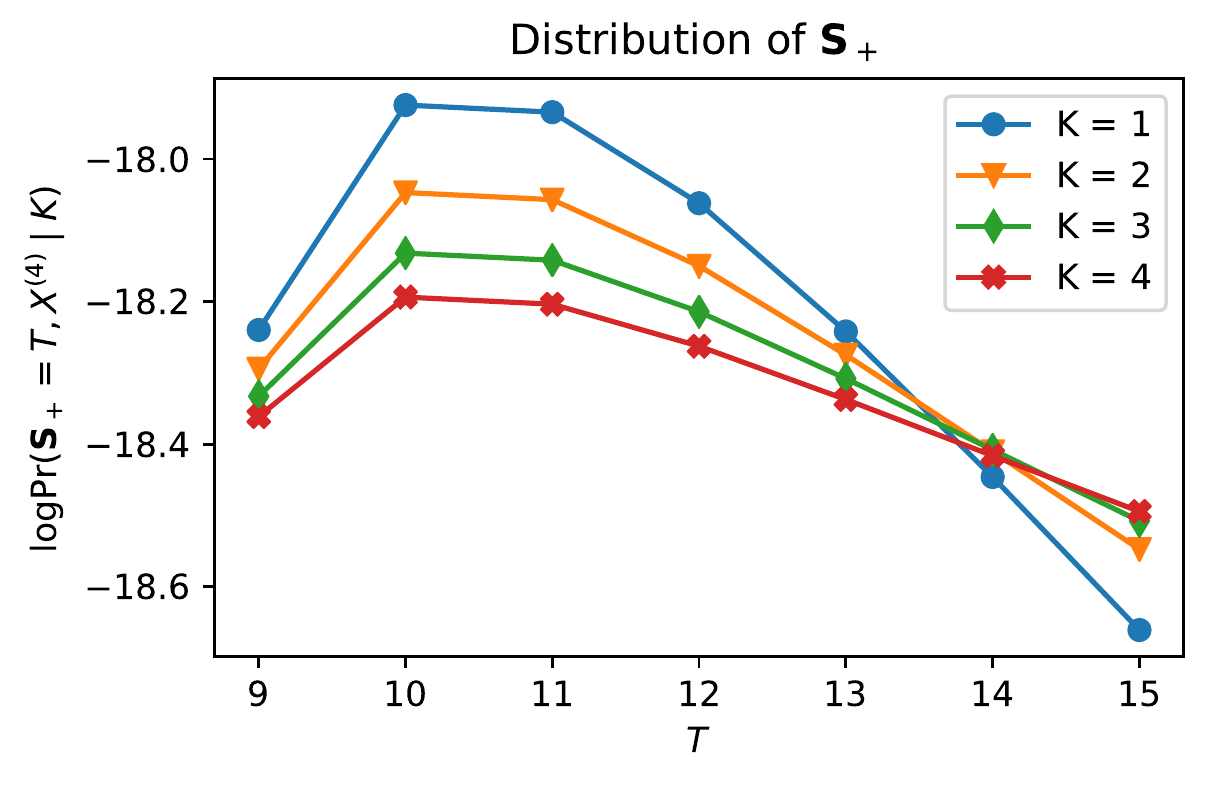}
		\caption{Unnormalized posterior distribution of $S_+$ given the matrix $X^{(4)}$.}
	\end{subfigure}		
	\caption{MAP estimations of the missing entries $X^{(3)}_{12}$ and $X^{(4)}_{12}$ are $3$ and $1$, since the posterior modes of $S_+$ are at $12$ and $10$ respectively.}
\end{figure}

In this example, we give an illustration of matrix completion problem with BAM. For simplicity, assume we have the following $2\times 2$ matrices $X^{(3)}$ and $X^{(4)}$ with a missing entry:
\begin{align*}
X^{(3)} & = \left(\begin{array}{cc} 
    3 & ? \\
    3 & 3 \\ 
    \end{array} \right) &
X^{(4)} & = \left(\begin{array}{ccc} 
    4 & ? \\
    4 & 1
    \end{array} \right) &
\end{align*}
As one entry is missing, the total number of tokens is not known here.

For each of the matrices, we have calculated the unnormalized posterior distribution $\Pr(\mathbf{S}_{+} = T, X \mid K)$ of $\mathbf{S}_{+}$ under KL-NMF model (Figure \ref{fig:KL-NMF}) which is also the posterior distribution of the missing entry in this case. Our calculations with exact enumeration for model shows that the distribution of the missing entry $X^{(3)}_{12}$ reaches its peak at $3$ whereas $X^{(4)}_{12}$ reaches its peak at $1$ for the all the cardinalities $K=1,\dots,4$ of the latent node. These results clearly indicate that the predictive distribution of the missing data is determined by the structure of the observed tensor $X$ rather than the observable moments of $X$ such as the mean.

\end{ex}

\subsection{NTF and Topic Models as Instances of BAM}\label{sec: Models as instances of a Bayesian Allocation Model}

In this subsection we illustrate examples where structured topic models or NTF models can be expressed as instances of BAM by defining an appropriate directed $\mathcal{G}$. 

\subsubsection{Pachinko allocation} \label{sec: Pachinko allocation}
Pachinko allocation \citep{Li2006} is proposed as an extension to LDA to model the fact that topics often have a hierarchical structure rather than the flat structure assumed by LDA. When expressed as a BAM, the generative model is essentially a chain as $j \rightarrow k_1 \rightarrow k_2 \dots \rightarrow k_L \rightarrow i$ where $L$ is the depth of the hierarchy. The observed indices are $V = \{i,j\}$, so we observe the counts for word-document pairs. As a concrete example with depth $L=3$, a token chooses first a document $j$, then conditioned on the document, chooses a super-topic $k_1$ (such as sports, politics, economics) , than a sub category $k_2$ (such as football, baseball, volleyball) and than conditioned on the sub category a topic $k_3$ (world-cup, premier league, fifa) and finally the word $i$. Like LDA, the observations are a matrix $S_{i+++j}$ and the original formulation of Pachinko allocation also conditions on the total number of tokens $S_+$.

\subsubsection{Mixed membership stochastic blockmodel} \label{sec: Mixed Membership Stochastic Blockmodel (MMB)}
A mixed membership stochastic blockmodel (MMB) \citep{Airoldi2008} is a generative model for relational data, essentially for the adjacency matrix of a directed graph. In this model, we have $I$ entities, indexed with $i_1$ and $i_2$, and for each of the possible $I^2$ pairs, where we observe a binary value that indicates if there is a directed edge or not. MMB can be viewed as an instance of BAM with the set of observed indices $V = \{i_1, i_2, s\}$, if we assume a one hot encoding where the observations are encoded as a tensor $S_{i_1i_2++s} $ where $i_1, i_2 \in [N]$ and $s \in [2]$.

The generative model corresponding to MMB is, $i_1 \rightarrow k_1 \rightarrow s \leftarrow k_2 \leftarrow i_2$. The token chooses a source $i_1$ and independently a destination $i_2$, followed by  the `blocks' for the source and destination. Finally, the category $s$ for the edge, here only a binary one, is chosen. As in LDA, we condition on the number of tokens $S_+ = I^2$.

\subsubsection{Sum conditioned Poisson factorization} \label{sec: Sum conditioned Poisson factorization}
Sum conditioned Poisson factorization is a recently proposed model for binary, categorical or ordinal data \citep{Capan2018} of form $X(i,j) \in [R]$ with possibly missing entries, where the aim is introducing a factorization structure on the moment parameters rather than the canonical parameters as in logistic or ordinal matrix and tensor factorizations \citep{Paquet2012, Nickel2016}. BAM is of form 
$p(j)p(k|j)p(r|k,j)p(i|k,r)$, where a token chooses first a source $j$, then jointly a topic $k$ and a category $r$ and finally a destination $i$. For observed $X(i,j)$, the observations are of form $S_{i+rj}$  and for missing $X(i,j)$, they are of form $S_{i++j} = M$ (hence the name sum conditioned). The model makes only sense if there are both missing and observed entries in $X$. If all entries are observed, the model degenerates to Poisson factorization.

\subsubsection{NTF with KL cost, Poisson tensor factorization} \label{sec: Nonnegative Tensor Factorization with KL cost (KL-NTF), Poisson Tensor Factorization}

The canonical Polyadic decomposition \citep{harshman1970fpp}, defined in \eqref{eq:parafac} is one of the widely used tensor models \citep{Kolda2009}. A closely related model is used for modelling multiway relations  \citep{ScheinPBW15,Schein16}. One parametrization of this model is 
\[
p(r)p(i_1|r)p(i_2|r)p(i_3|r)
\]
with visible indices $i_1$, $i_2$ and $i_3$. Another Markov equivalent model is 
\[
p(i_1)p(r|i_1)p(i_2|r)p(i_3|r),
\]
which is also a canonical Polyadic model. A nonnegative 3-way Tucker model \eqref{eq:tucker} is 
\[
p(i_6)p(i_5|i_6)p(i_4|i_5,i_6)p(i_3|i_6)p(i_2|i_5)p(i_1|i_4)
\]
with visible indices $V=\{1,2,3\}$. For both models, the corresponding graphical models are shown in Figure \ref{fig:equivalent models}. Many classical graphical model structures can also be viewed as tensor models. For example, a hidden Markov model can be viewed as a $T$-way Tucker model with $2T$ indices where the core tensor is factorized as a Markov chain. In the tensor literature, related models are also known as tensor trains or tensor networks \citep{Oseledets_2011, Cichocki2016}. 

\subsection{Tensor Factorizations as Directed Models} \label{sec: Tensor Factorizations as Directed Models}
In its full generality, a NTF model with KL cost \citep{yilmazGTF} (see in \eqref{eq:ntf-original}) can be  expressed as an undirected graphical model, as a product of individual component tensors, followed by a contraction over hidden indices. However, unlike the examples in the previous subsection, it is not always possible to find an equivalent directed graphical model for a tensor model, as not all undirected models have exact directed representations. A well known example is 
\[
X(i_{1:4}) \approx  W(i_1, i_2) W(i_2, i_4) W(i_1, i_3) W(i_3, i_4)
\]
that corresponds to a four-cycle. This undirected graphical model can not be  expressed exactly, in the sense that it implies certain Markov properties that can not exactly be expressed by a directed graphical model. Hence, this 
 tensor model can not be expressed exactly as a BAM. The technical condition for exact representation is the \emph{decomposability} \citep{Lauritzen1996} of the underlying undirected graph:  every decomposable tensor model
can always exactly formulated as a BAM.

Given a tensor model, it is always possible to construct a decomposable model by adding redundant edges, equivalently extending the factors. For the above example one such construction is 
\[
X(i_{1:4}) \approx W(i_1, i_2, i_3) W(i_2, i_3, i_4) 
\]
but this introduces redundant parameters hence ignores some structural constraints. Nevertheless, for many popular tensor models such as a nonnegative canonical Polyadic (PARAFAC) or nonnegative Tucker models, the models are decomposable hence it is easy to express a corresponding directed graphical model. Given a general tensor model, we can systematically construct a directed representation, hence a BAM that can represent data equivalently well as follows: extend each factor appropriately such that the corresponding undirected graph is decomposable, (corresponding to the called moralization and triangulation steps), and form an order $\sigma$ of nodes respecting a perfect numbering. We construct a directed graph $\mathcal{G}$ sequentially by adding each node $v$ in the order given by $\sigma$ and add directed edges from a node $w$ previously added to $\mathcal{G}$ if $v$ and $w$ are in the same clique, i.e., for all nodes $w$ such that $\sigma(w) < \sigma(v)$ and $v,w \in C$ for a clique $C \in \mathcal{C}_\mathcal{G}$.    

\section{Inference by Sequential Monte Carlo}\label{sec: Inference by Sequential Monte Carlo}
In this section, we will first provide an analysis of the allocation model as a dynamic \Polya urn process where tokens are placed according to an underlying Bayesian network, in a sense that we will later describe. It will turn out that the conditional independence relations implied by the Bayesian network also carry forward to the urn process that significantly simplifies inference and makes it possible to run sequential algorithms with minor space requirement. Based on this interpretation, we will describe a general sequential Monte Carlo (SMC) algorithm for BAM, with the primary motivation of estimating the marginal likelihood $\mathcal{L}_{X}$ in \eqref{eq: data marginal likelihood}. The BAM perspective is not only useful for devising novel inference algorithms that are favorable when data consists of small counts, it also provides further insight about the clustering nature of nonnegative decompositions in general.

\subsection{\Polya Urn Interpretation of the Allocation Model} \label{sec:Polya urn interpretation of the marginal allocation model}

A \Polya urn \citep{Mahmoud2008, Pemantle2007} is a self reinforcing counting process, that is typically described using the metaphor of colored balls drawn with replacement from an urn. In the basic urn model, each ball has one of the $I$ distinct colors; one repeatedly draws a ball uniformly at random and places it back with an additional ball of the same color so the process exhibits self reinforcement. 

To make the analogy to the allocation model, we will associate each color with a cell and we let $\mathbf{S}_i^\tau$  denote the number of tokens allocated to cell $i$ such that $i \in [I]$ at time $\tau$. Initially there are no tokens, with $\mathbf{S}_i^0 = 0$ where $i\in [I]$. The first token is placed into cell $i$ with probability $\alpha_i/\alpha_+$ and in each subsequent step $\tau$, a token is placed into cell $i$ with probability:
\begin{align*}
\Pr\{\text{token $\tau$ is placed in cell $i$}| \mathbf{S}^{\tau-1} = S^{\tau-1} \} &= \Pr\{\mathbf{s}^\tau_i = 1|S^{\tau-1} \} \\
&= \frac{\alpha_i + S_i^{\tau-1}}{ \alpha_+ + S_+^{\tau-1}} \equiv \zeta^\tau_i
\end{align*}
At step $\tau$, set $S_i^{\tau} = S_i^{\tau-1} + s_i^\tau$, where $s_i^\tau$ is the indicator of the event that the token $\tau$ is placed into cell $i$. Clearly, the placement of tokens has the same law as a \Polya urn. This is a Markov process with the special property that the future of the process heavily depends on the first few outcomes and its limit probability vector $\lim_{\tau\rightarrow \infty} \zeta_:^\tau$ has a Dirichlet distribution $\mathcal{D}(\alpha)$. 

\subsubsection{\PolyaBayes processes} \label{sec: Polya-Bayes networks} 
In the following, we show the close connection between \Polya urns and BAM developed in Section \ref{sec: Allocation Model}. In particular, we show that BAM in Section \ref{sec: Allocation Model} admits a \Polya urn interpretation if $\{ \theta_{n|\pa{n}} : n \in [N] \}$ have the prior distribution in \eqref{eq:allocationmodel}.

Consider the allocation model in Section \ref{sec: Allocation Model} and assume $\Theta = \{ \theta_{n|\pa{n}} : n \in [N] \}$ have the prior distribution in \eqref{eq:allocationmodel}, so that we have BAM in Section \ref{sec: Bayesian Allocation Model (BAM)}. For the sequence $\{ \mathbf{S}^{\tau} \}_{\tau \geq 1}$ of tensors generated by the allocation model, define the transition probabilities
\[
f_{\tau_{2} |  \tau_{1}}(S' | S) \equiv \Pr(\mathbf{S}^{\tau_{2}} = S'  |  \mathbf{S}^{\tau_{1}} = S)
\] 
where $S$ is a viable tensor at time $\tau_{1}$. We first look at the forward transition probability $f_{\tau | \tau-1}(S^{\tau}| S^{\tau-1})$ for a viable pair $(S^{\tau-1}, S^{\tau})$ with $s^{\tau}(i_{1:N}) = 1$ for some index $i_{1:N}$. Clearly, we can write 
\begin{align*}
f_{\tau | \tau-1}(S^{\tau}| S^{\tau-1}) &= \Pr(\mathbf{s}^{\tau}(i_{1:N}) = 1 | \mathbf{S}^{\tau-1} = S^{\tau-1}) \\
& = \Pr(\mathbf{c}^{\tau} = i_{1:N} | \mathbf{S}^{\tau-1} = S^{\tau-1})
\end{align*}

Since $\mathbf{c}^{\tau}$ is independent from $\mathbf{S}^{\tau-1}$ given $\Theta$, we can decompose the probability of the increment as
\begin{align}
\Pr(\mathbf{c}^{\tau} = i_{1:N}  | \mathbf{S}^{\tau-1} = S^{\tau-1}) &=  \int \Pr(\mathbf{c}^{\tau} = i_{1:N} | \Theta) \pi(\Theta | S^{\tau-1}) \, \mathrm{d}\Theta  \nonumber \\ 
& = \int \theta(i_{1:N}) \pi(\Theta | S^{\tau-1}) \, \mathrm{d}\Theta  \nonumber \\ 
&= \E{\left. \prod_{n = 1}^{N} \theta_{n|\pa{n}}(i_n, i_\pa{n}) \right\vert \mathbf{S}^{\tau-1} = S^{\tau-1} } \nonumber \\
&= \prod_{n = 1}^{N} \frac{ \alpha_{\fa{n}}(i_n, i_\pa{n}) + S^{\tau-1}_{\fa{n}}(i_n, i_\pa{n}) }{\sum_{i'_{n}} \alpha_{\fa{n}}(i'_{n}, i_\pa{n}) + S^{\tau-1}_{\fa{n}}(i'_{n}, i_\pa{n})} \label{eq: PU last line in incremental prob} 
\end{align}
where we have used \eqref{eq: probability of increment given theta}, \eqref{eq:bn}, and \eqref{eq: post of lamda and theta given S} in the second, third, and the last lines, respectively. 

The expression for $f_{\tau | \tau-1}(S^{\tau}| S^{\tau-1})$ in \eqref{eq: PU last line in incremental prob} exploits the structure in graph $\mathcal{G}$ and suggests a way to generate or evaluate a one-step transition for the sequence $\{ \mathbf{S}^{\tau} \}_{\tau \geq 0}$. Specifically, \eqref{eq: PU last line in incremental prob} implies a collection of dependent \Polya urn processes with a sequential dependence structure. Specifically, the process $\{ \mathbf{S}^{\tau} \}_{\tau \geq 1}$ can be viewed as a generalized form of a ``\Polya tree'' of depth $N$ as defined in \citet[Section 3]{Mauldin_et_al_1992}. A \Polya tree and our process is equivalent if the graph $\mathcal{G}$ is complete. Our process is more general due to capturing the conditional independence structures in the Bayesian network with respect to the graph $\mathcal{G}$. To highlight this, and avoid confusion with the term ``tree'' of a graph, we will call the process $\{ \mathbf{S}^{\tau} \}_{\tau \geq 1}$ a \Polya urn process on a Bayesian network, or shortly a \emph{\PolyaBayes process}.

In compact form, we can write the transition probability in \eqref{eq: PU last line in incremental prob} shortly as
\begin{equation} \label{eq: transition density compact}
f_{\tau | \tau -1}(S^{\tau} | S^{\tau-1}) = \prod_{n = 1}^{N}  \frac{B_n( \alpha_{\fa{n}} + S_{\fa{n}}^{\tau} )}{B_n( \alpha_{\fa{n}} + S_{\fa{n}}^{\tau-1} )} 
\end{equation}
which reveals a several interesting facts about the process $\{ \mathbf{S}^{\tau} \}_{\tau \geq 0}$.
For example, notice from \eqref{eq: transition density compact} that
\[
\Pr(\mathbf{S}^{1:\tau} = S^{1:\tau}) = \prod_{k = 1}^{\tau} f_{k | k-1}(S^{k} | S^{k-1}) = \prod_{n = 1}^{N} \frac{B_{n}(\alpha_{\fa{n}} + S^{\tau}_{\fa{n}})}{B(\alpha_{\fa{n}})}
\]
which is the non-combinatorial factor in \eqref{eq:pS2} and depends only on $S^{\tau}$. We deduce that the incremental process $\{ \mathbf{s}^{\tau} \}_{\tau \geq 1}$ is exchangeable. Moreover, noting that there are $\left(\begin{matrix} \tau \\ S^{\tau} \end{matrix} \right)$ distinct viable $S^{1:\tau}$ sequences ending with $S^{\tau}$, which is the second factor in \eqref{eq:pS2}, we verify that
\[
\Pr(\mathbf{S}^{\tau} = S^{\tau}) = \prod_{n = 1}^{N} \frac{B_{n}(\alpha_{\fa{n}} + S^{\tau}_{\fa{n}})}{B(\alpha_{\fa{n}})} \left(\begin{matrix} \tau \\ S^{\tau} \end{matrix} \right) = \pi_{\tau}(S^{\tau})
\]

Note that, as the factorization of $\Theta$ is not unique, all the Markov equivalent graphs of $\mathcal{G}$ can be used to generate the same \PolyaBayes process, albeit in a different topological order. In fact, for a general thinning mechanism according to a directed graph $\mathcal{G}$, if the corresponding moral graph is \emph{decomposable} \citep{Cowell2003}, a \Polya urn representation can be constructed from the Junction tree factorization (clique potentials divided by separator potentials and a normalization term) of \eqref{eq:pS3} as:
\begin{equation}
\Pr(\mathbf{s}^{\tau}(i_{1:N}) = 1 \mid \mathbf{S}^{\tau-1} = S^{\tau-1}) = \frac{1}{(\alpha_+ + S^{\tau-1}_+)}\frac{\prod_{q=1}^{|\Cliques|}  (\alpha_{C_q}(i_{C_q}) + S^{\tau-1}_{C_q}(i_{C_q}))}{\prod_{q=1}^{|\Separators|}  (\alpha_{D_q}(i_{D_q}) + S^{\tau-1}_{D_q}(i_{D_q}))} \ \label{eq:pSq}
\end{equation}
The individual terms here contain the sufficient statistics that need to be stored for exactly drawing samples from the urn process. This factorization is useful for sampling any desired conditional with a reduced space complexity: first propagate messages to the root clique and than sample starting from the root. In hidden Markov models, the special case of these algorithms are known as forward filtering  backward sampling \citep{Cappe2005}.

\begin{ex}
As a concrete example, the \PolyaBayes process corresponding to the NMF model has the transition probability
\begin{align*}
\Pr(\mathbf{s}^\tau_{ikj}=1 \mid \mathbf{S}^{\tau-1} = S^{\tau-1}) = \frac{( \alpha_{++j} + S^{\tau-1}_{++j} )}{(\alpha_{+} + S^{\tau-1}_{+})} \frac{( \alpha_{+kj} + S^{\tau-1}_{+kj} )}{(\alpha_{++j} + S^{\tau-1}_{++j})} \frac{( \alpha_{ik+} + S^{\tau-1}_{ik+})}{(\alpha_{+k+} + S^{\tau-1}_{+k+} )}
\end{align*}
This \PolyaBayes process has $I\times K\times J$ colors where each color is specified by the tuple $(i\;k\;j)$ and balls are drawn according to the underlying graph $j \rightarrow k \rightarrow i$, with probabilities given by the three ratios. In a sense, each node of the Bayesian network encodes partial information about the color, reminiscent to individual R, G, and B components in the RGB color representation. In the corresponding urn, at step $\tau$, the token is drawn proportional to $\alpha + S^{\tau-1}$ and after the draw the counts are updated as $S^\tau_{ikj} \leftarrow S^{\tau-1}_{ikj} + s^\tau_{ikj}$ and then the next token at $\tau+1$ is drawn proportional to $\alpha + S^{\tau}$. In the typical parameter regime where the pseudo-counts $\alpha$ are chosen small, the urn parameter enforces the tokens to get clustered and leads to the observed clustering behaviour of NMF. 

The other Markov equivalent graphs $j \leftarrow k \rightarrow i$ and $j \leftarrow k \leftarrow i$ correspond to following factorizations:
\begin{align*}
\Pr(\mathbf{s}^\tau_{ikj}=1 \mid \mathbf{S}^{\tau-1} = S^{\tau-1})
& =  \frac{( \alpha_{+kj} + S^{\tau-1}_{+kj} )}{(\alpha_{+k+} + S^{\tau-1}_{+k+})} \frac{( \alpha_{+k+} + S^{\tau-1}_{+k+} )}{(\alpha_{+} + S^{\tau-1}_{+})} \frac{( \alpha_{ik+} + S^{\tau-1}_{ik+})}{(\alpha_{+k+} + S^{\tau-1}_{+k+} )} \\
& =  \frac{( \alpha_{+kj} + S^{\tau-1}_{+kj} )}{(\alpha_{+k+} + S^{\tau-1}_{+k+})} \frac{( \alpha_{ik+} + S^{\tau-1}_{ik+})}{(\alpha_{i++} + S^{\tau-1}_{i++} )} \frac{( \alpha_{i++} + S^{\tau-1}_{i++} )}{(\alpha_{+} + S^{\tau-1}_{+})} \\
& = \frac{1}{(\alpha_{+} + S^{\tau-1}_{+})} \frac{( \alpha_{+kj} + S^{\tau-1}_{+kj} )( \alpha_{ik+} + S^{\tau-1}_{ik+})}{(\alpha_{+k+} + S^{\tau-1}_{+k+})} 
\end{align*}
The last line corresponds to the basic urn schema that corresponds to the junction tree factorization specifically for the NMF model.
\end{ex}

\paragraph{Reverse process:} We can also calculate the transition probabilities for the reverse \PolyaBayes process, that is how to remove tokens from their allocations. The procedure is simple to describe, select any token uniformly at random and remove it. To see this, consider a viable pair $(S^{\tau-1}, S^{\tau})$ with $s^{\tau}(i_{1:N}) = 1$ for some index $i_{1:N}$. The Bayes rule for the reverse transition probability can be written as
\[
f_{\tau-1 | \tau}(S^{\tau-1} | S^{\tau}) =  f_{\tau | \tau-1}(S^{\tau} | S^{\tau-1}) \frac{\pi_{\tau-1}(S^{\tau-1}) }{\pi_{\tau}(S^{\tau})}
\]
By \eqref{eq:pS2} (replacing $S_{+}$ with  $\tau-1$ and $\tau$), \eqref{eq: transition density compact} and cancelling $B_{n}(\cdot)$'s, we obtain
\begin{eqnarray}
f_{\tau-1 | \tau}(S^{\tau-1} | S^{\tau}) & = & 
\frac{S^{\tau-1}_+!}{\prod_{i'_{1:N}} S^{\tau-1}({i'_{1:N}})!} \frac{\prod_{i'_{1:N}} S^{\tau}({i'_{1:N})}!} {(S^{\tau-1}_+ + 1)!} \nonumber \\
& = & \frac{S^{\tau}({i_{1:N}})}{S^\tau_+} \label{eq:back_probability}
\end{eqnarray}
since the only difference between $S^{\tau-1}$ and $S^{\tau}$ is at $i_{1:N}$. Note that \eqref{eq:back_probability} is  $\Pr(\mathbf{s}^{\tau+1}({i_{1:N}}) = 1| \mathbf{S}^{\tau}  = S^{\tau})$. This leads to a multinomial sampling without replacement procedure, where each cell $i'_{1:N}$ is sampled proportional to its occupancy and the selected token is removed.

\subsection{Sequential Monte Carlo for Estimating the Marginal Likelihood} \label{sec: Sequential Importance Sampling for Estimating Likelihood}
In this section, we propose a Monte Carlo based estimator of the marginal likelihood $\mathcal{L}_{X}$ in \eqref{eq: data marginal likelihood} for a given $\mathbf{X} = X$ generated as in \eqref{eq: X as partially observed S}.

We will assume that all entries of $X$ are observed and therefore the total number of tokens $T$ is available. As a result, we will treat $T$ as a fixed number in this section. Moreover, note that $\mathcal{L}_{X}$ is factorized as
\[
\mathcal{L}_{X} = \Pr(\mathbf{S}_{+} = T) \Pr(\mathbf{S}_{V}^{T} = X) 
\]
where $\Pr(\mathbf{S}_{+} = T)$ can easily be calculated from \eqref{eq:pS1} and does not depend on any structural properties of the graph $\mathcal{G}$. Therefore, we focus on estimation of the probability $\Pr(\mathbf{S}_{V}^{T} = X)$. This probability can be written as
\begin{equation} \label{eq: likelihood as a sum}
\Pr(\mathbf{S}_{V}^{T} = X) 
= \sum_{S} \pi_{T}(S) \ind{X = S_V} 
\end{equation}  
This formulation suggests that calculating the marginal likelihood is equivalent to computing the probability that the \PolyaBayes process $\{ \mathbf{S}^{\tau} \}_{\tau \geq 1}$ hits the target set
\[
\Omega = \{S : S_V = X\}
\]
at step $T$, so $\Pr(\mathbf{S}_{V}^{T} = X) = \Pr\{\mathbf{S}^{T} \in \Omega\}$. 

\begin{figure}
    \centering
    \includegraphics[width=0.4\textwidth]{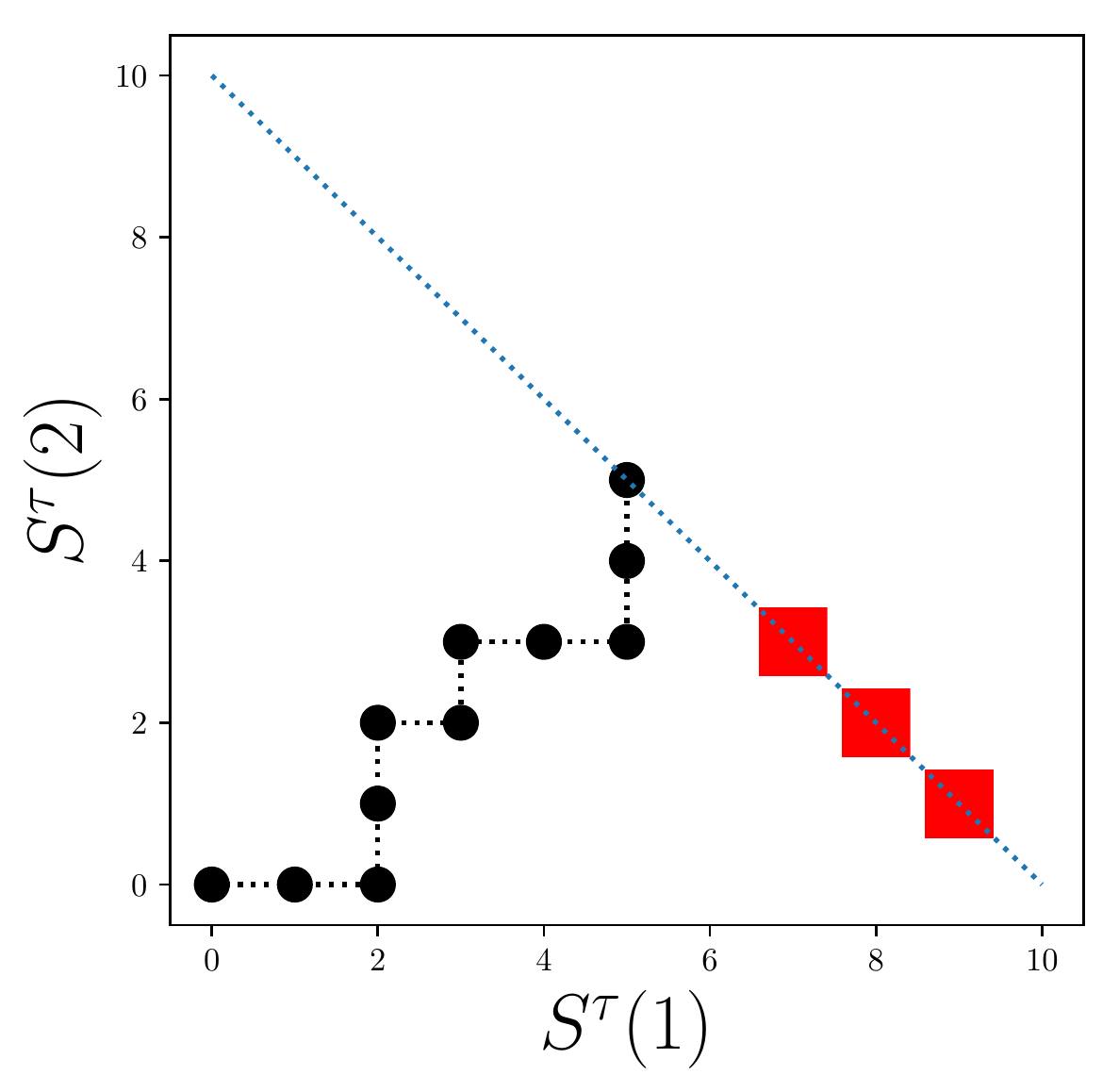}
    \caption{Computing the probability of a \PolyaBayes urn process (a realization is shown by $\bullet$'s) hitting the target set $\red \blacksquare$  constitutes a rare event problem, prohibiting the use of a standard importance sampling. In a matrix factorization setting, the target set would consist of tensors whose marginal are equal to a given marginal tensor $X$.}
    \label{fig:polya_hit}
\end{figure}


\subsubsection{Sequential importance sampling} \label{sec: Sequential importance sampling}
A naive way of performing importance sampling to estimate the sum in \eqref{eq: likelihood as a sum} is simply to generate $M$ independent configurations, $S^{T(1)}, \ldots, S^{T(M)}$, by simulating the \PolyaBayes process until time $T$ and calculate the estimate
\begin{equation} \label{eq: naive IS for likelihood}
\Pr(\mathbf{S}_{V}^{T} = X) \approx \frac{1}{M} \sum_{m=1}^M \ind{X = S_V^{T (m)}}.  
\end{equation}
However, this estimator would be extremely poor in the sense of having a high variance unless $S_+ = T$ is small. Instead, we can use the more efficient sequential importance sampling (SIS) technique to design a sequentially built proposal that takes $X$ into account to avoid zero weights.

As the urn process is a non decreasing counting process with increments of one at each step, the indicator of the set $\Omega$ can be written for all viable $S^{1:T}$ equivalently as:
\begin{equation} \label{eq: equivalence of indicators}
\ind{S_V^T = X} = \prod_{\tau=1}^T g(S^\tau),
\end{equation}
where the factorization is enabled with the indicator function 
\[
g(S^\tau) \equiv \ind{{X \geq S_V^\tau}} 
\]
with the relation ``$\geq$'' meant to apply element-wise. The indicator encodes the condition that the total number of balls allocated to the fiber $S(i_V,i_{\bar{V}})$ for all values of $i_{\bar{V}}$ should never exceed the observations $X(i_{V})$ as otherwise the urn process will never hit the target set $\Omega$. Conversely, the process will inevitably hit the target set if this condition is satisfied through the entire trajectory. 

The factorization of the indicator over $\tau$ in \eqref{eq: equivalence of indicators} enables us to formulate the calculation of the marginal likelihood as a sequential rare event problem. We will formulate our proposed importance sampling within this context. 

We remind the reader the one-to-one relation among $\{ S^{\tau} \}_{\tau \geq 0}$, $\{ s^{\tau} \}_{\tau \geq 1}$ and $\{ c^{\tau} \}_{\tau \geq 1}$, and that we will be using those variables together with reference to their relationship. That in mind, let
\[
p_{\tau}(c^{\tau} | c^{1:\tau-1}) = p_{\tau}(c^{\tau} | S^{\tau-1}) = f_{\tau | \tau - 1}(S^{\tau} | S^{\tau-1})
\]
be the conditional probability of $\mathbf{c}^{\tau}$ given its past, 
and note that the probability of the entire trajectory can be written as
\[
\pi(S^{1:T}) =  \prod_{\tau = 1}^{T} p_{\tau}(c^{\tau} | c^{1:\tau-1}).
\]  
(We have chosen to use the increment indices $c^{\tau}$ for notational convenience in the derivations to follow.) Then, one can write 
\begin{eqnarray}
\Pr(\mathbf{S}_{V}^{T} = X) 
& = & \sum_{S^{1:T}} \pi(S^{1:T}) \left[ \prod_{\tau = 1}^{T} g(S^{\tau})  \right]  \nonumber\\
& = & \sum_{c^{1:T}} \prod_{\tau = 1}^{T} p_{\tau}(c^{\tau} | c^{1:\tau-1})  g(S^{\tau}) \label{eq: likelihood as a sum in terms of increment indices} \\
& \equiv & \sum_{c^{1:T}} \phi(c^{1:T}) \nonumber
\end{eqnarray}
Equation \eqref{eq: likelihood as a sum in terms of increment indices} suggests for designing a sequential proposal mechanism that can respect the observed data $X$ at each time step, thanks to the factor $g(S^{\tau})$. 

Our proposal mechanism for SIS is based on the observation that we can obtain exact samples from the conditional distribution of $\mathbf{S}_V^{1:T}$, conditioned on $X = S_V^T$. More formally, we partition the index of increment as 
\[
c^\tau =(c_{V}^{\tau}, c_{\bar{V}}^{\tau}).
\]
and construct our proposal in two steps 
\[
q(c^{1:T}) = q_{V}(c_{V}^{1:T}) q_{\bar{V}}(c_{\bar{V}}^{1:T} | c_{V}^{1:T})
\]
where the densities on the right hand side are further decomposed into one-step conditional densities
\[
q_{V}(c_{V}^{1:T}) = \prod_{\tau = 1}^{T} q_{\tau, V}(c_{V}^{\tau} | S_{V}^{\tau-1}), \quad q_{\bar{V}}(c_{\bar{V}}^{1:T} | c_{V}^{1:T}) = \prod_{\tau = 1}^{T} q_{\tau, \bar{V}}(c_{\bar{V}}^{\tau} | c^{1:\tau-1}, c_{V}^{\tau})
\]
which suggests the sequential nature of the proposal mechanism. We describe those two steps below:
\begin{enumerate}
\item Our proposal for $c_{V}^{1:T}$ corresponds to simulating \emph{backward} the sequence of marginal tensors $S_V^{1:T-1}$ conditioned on $S_{V}^{T} = X$. More formally, the one-step conditional proposal distribution is 
\[
q_{\tau, V}(c_{V}^{\tau} | S_{V}^{\tau-1}) = \frac{(X - S_{V}^{\tau-1})(c_{V}^{\tau})}{\sum_{i_{V}} (X - S_{V}^{\tau-1})(i_{V})} = \frac{(X - S_{V}^{\tau-1})(c_{V}^{\tau})}{T - \tau + 1}
\]
This is multinomial sampling of cells with probabilities proportional to the (remaining) number of tokens in those cells. In practice, this mechanism can be implemented by uniform sampling of the tokens in $X$ without replacement.
\item The remaining part $c_{\bar{V}}^{1:T}$ is proposed sequentially by sampling $c_{\bar{V}}^{\tau}$ from the conditional distribution of $\mathbf{c}_{\bar{V}}^{\tau}$ given $c^{1:\tau-1}$ and $c_{V}^{\tau}$,
\[
q_{\tau, \bar{V}}(c_{\bar{V}}^{\tau} | c^{1:\tau-1}, c_{V}^{\tau}) = p_{\tau}(c_{\bar{V}}^{\tau} | c_{V}^{\tau}, S^{\tau-1}) = \frac{p_{\tau}(c^{\tau} | S^{\tau-1})}{p_{\tau, V}(c_{V}^{\tau} | S^{\tau-1})}
\]
where
\[
p_{\tau, V}(c_{V}^{\tau} | S^{\tau-1}) = \sum_{i_{\bar{V}}}p_{\tau}((c_{V}^{\tau}, i_{\bar{V}}) | S^{\tau-1})
\]
\end{enumerate}
The resulting importance weight function is 
\begin{align}
W(c^{1:T}) = \frac{\phi(c^{1:T})}{q(c^{1:T})} 
& = \prod_{\tau = 1}^{T} \frac{T - \tau + 1}{(X - S_{V}^{\tau-1})(c_{V}^{\tau})}  p_{\tau, V}(c_{V}^{\tau} | S^{\tau-1}) \label{eq: weight function suggesting recursion} \\
& = \frac{T !}{\prod_{i_{V}} X(i_{V})!} \prod_{\tau = 1}^{T}  p_{\tau, V}(c_{V}^{\tau} | S^{\tau-1}) \label{eq: weight function compact}
\end{align}
Notice that the first part of \eqref{eq: weight function compact} corresponds to Step 1 of the proposal mechanism
and it does not depend on the choice of $c^{\tau}$'s.  

The SIS proposal mechanism allows us to calculate the weight function sequentially. Observing \eqref{eq: weight function suggesting recursion}, we can write the weight function as
\[
W(c^{1:T}) =  \prod_{\tau=1}^T u^\tau(c_{V}^\tau, S^{\tau-1})
\]
where the incremental weight is 
\begin{eqnarray}
u^\tau(c_{V}^\tau, S^{\tau-1}) = p_{\tau, V}(c_{V}^{\tau} | S^{\tau-1}) \frac{T - \tau + 1}{(X - S_{V}^{\tau-1})(c_{V}^{\tau})} \label{eq: incremental weight function}
\end{eqnarray}
Note that since $S^{\tau-1}$ can recursively be computed from $c^{1:\tau-1}$, computational load of \eqref{eq: incremental weight function} does not increase with $\tau$.

An important observation is that the incremental weight in \eqref{eq: incremental weight function} does not depend on the sampled value $c^{\tau}_{\bar{V}}$. In fact, this is a good sign when the variability of the importance weights is concerned: If confined to sequential proposal mechanisms, our choice for $q$ overall is optimal in the sense that it minimizes the variance of incremental weights \citep{liu_and_chen_1998, doucet2009tutorial}.

\begin{algorithm}
\caption{BAM-SIS: SIS for BAM} \label{alg:sis}
\begin{algorithmic}[1]
\Procedure{BAM-SIS}{$X$} 
\State $T = X_{+}$
\State $Z_{0} = \frac{b^{a}}{(b+1)^{a+ T}}\frac{\Gamma(a + T)}{\Gamma(a)} \frac{1}{T!}$
\State Set $W = 1$, 
\State Set $S = 0$  
\For{$\tau=1, \dots, T$}
\State Sample $c_V \sim q_{\tau, V}(c_V | S_{V})$
\State Sample $c_{\bar{V}}  \sim  p_{\tau}(c_{\bar{V}}  | c_V, S)$ 
\State Set $c = (c_{V}, c_{\bar{V}})$
\State Set $u =  p_{\tau, V}(c_{V} | S) \frac{T - \tau + 1}{(X - S_{V})(c_{V})}$
\State Update $W \leftarrow W \times u$
\State Update $S(c) \leftarrow  S(c) + 1$
\EndFor\label{sisfor-1}
\State \textbf{return} $(S, W)$ \Comment{weighted sample for $\pi(S | X)$}
\State \textbf{return} $Z = Z_{0} \times W$ \Comment{estimate of $\mathcal{L}_{X}$}
\EndProcedure
\end{algorithmic}
\end{algorithm}

Algorithm \ref{alg:sis} 
presents the BAM-SIS algorithm, which can be used to generate a weighted sample from the posterior of BAM $\pi(S | X)$ defined in \eqref{eq: posterior of S given X}. In order to demonstrate the recursive nature of BAM-SIS and emphasize on the fixed memory requirement, we have dropped the time index $\tau$.  The implicit requirement is that it should be feasible to do exact sampling from the conditional distribution $p_{\tau}(s_{\bar{V}}^\tau  | s_V^\tau, S^{\tau-1} )$ of the directed graphical model $\mathcal{G}$. The complexity of inference is closely related to the junction tree factorization and for many models of interest in practice, we are not required to explicitly store $S$ but only need to have access to counts of form  $S_{\fa{n}}$, hence it is sufficient to maintain marginal counts corresponding to the clique marginals of form \eqref{eq:pSq}. We illustrate this for the KL-NMF in the following example.

\begin{ex}
We provide in Algorithm \ref{alg:sis-klnmf} a particular instance of BAM-SIS for the KL-NMF model that corresponds to the graph $i\leftarrow k \rightarrow j$. For this model, it is sufficient to maintain two matrices $S^\tau_{ik+}$ and $S^\tau_{+kj}$ and the reconstruction $S^\tau_{i+j}$.  
\begin{algorithm}
\caption{BAM-SIS for KL-NMF ($i\leftarrow k \rightarrow j$)}\label{alg:sis-klnmf}
\begin{algorithmic}[1]
\Procedure{KL-NMF-SIS}{$X$} 
\State $T = X_{+}$.
\State $S_{+k+} = 0, S_{ik+} = 0, S_{+kj} = 0, S_{i+j} = 0$, for all $i, j, k$.  
\State $W = 1$, 
\State $Z_{0} = \frac{b^{a}}{(b+1)^{a+ T}}\frac{\Gamma(a + T)}{\Gamma(a)} \frac{1}{T!}$
\For{$\tau = 1,\dots,T$}
\State Sample $c_{V}^\tau = (i, j)$ with probability $\frac{X_{i, j} - S_{i + j}}{T-\tau + 1}$
\State Sample $c_{\bar{V}} = k$ with probability $\theta_{k}/\theta_{+}$ where
\[
\theta_{k} = \frac{( \alpha_{+kj} + S_{+kj} )( \alpha_{ik+} + S_{ik+})}{(\alpha_{+} + S_{+})(\alpha_{+k+} + S_{+k+})}, \quad k = 1, \ldots, K, \quad \theta_{+} = \sum_{k = 1}^{K} \theta_{k}
\]
\State Set $u = \theta_{+} \frac{ T - \tau + 1}{S_{i + j}}$
\State Update $W \leftarrow W \times u  $
\State Update the marginal tensors
\begin{align*}
& S_{ik+} \leftarrow S_{ik+} + 1, \quad S_{+kj} = S_{+kj} + 1, \\
& S_{i+j} \leftarrow S_{i+j} + 1, \quad S_{+k+} = S_{+k+} + 1
\end{align*}
\EndFor\label{sisfor-2}
\State \textbf{return} $(S, W)$ \Comment{weighted sample for $\pi(S | X)$}
\State \textbf{return} $Z = Z_{0} \times W$ \Comment{estimate of $\mathcal{L}_{X}$}
\EndProcedure
\end{algorithmic}
\end{algorithm}
\end{ex}

\paragraph{Sequential importance sampling - resampling} \label{sec: Sequential importance sampling - resampling} 
Although better than \eqref{eq: naive IS for likelihood}, the SIS estimator in Algorithm \ref{alg:sis} is still impoverished by increasing $T$, as the variance of the importance weights increases exponentially in $T$, resulting in what is called weight degeneracy. To overcome weight degeneracy, SIS is accompanied with a resampling procedure, where (in its default implementation) the particles are resampled according to their weights, which then become $1/M$ after resampling. The resampling idea is first introduced in \citet{gordon_et_al_1993}, leading to the famous SIS - resampling (SIS-R) method, aka the particle filer. Standard resampling schemes include multinomial resampling \citep{gordon_et_al_1993}, residual resampling \citep{whitley_1994, liu_and_chen_1998}, stratified resampling \citep{kitagawa_1996}, and systematic resampling \citep{whitley_1994, Carpenter_et_al_1999}.  

Algorithm \ref{alg:sisR} presents the SIS-R algorithm for BAM, or shortly BAM-SIS-R (with steps \ref{alg-step: update Z} and \ref{alg-step: resample} always implemented). BAM-SIS-R produces weighted samples for the posterior $\pi(S^{T} | X)$ and, more importantly for our work, an unbiased estimator of the marginal likelihood $\mathcal{L}_{X}$. The unbiasedness is established in \citet{del_moral_2004}.

One way to improve the performance of SIS-R is adaptive resampling, i.e., resampling only at iterations where the effective sample size drops below a certain proportion of $M$. For a practical implementation, the effective sample size should be estimated from particle weights, such as in \citet[pp. 35-36]{liu_2001}. Algorithm \ref{alg:sisR} covers adaptive resampling as well, where the decision to resample or not is taken at step \ref{alg-step: adaptive resampling}. Although adaptive resampling does reduce the variance of the marginal likelihood estimator, unbiasedness is no more guaranteed.

\begin{algorithm}
\caption{BAM-SIS-R: SIS-R for BAM}\label{alg:sisR} 
\begin{algorithmic}[1]
\Procedure{BAM-SIS-R}{$X$} 
\State $T = X_{+}$
\State $Z_{0} = \frac{b^{a}}{(b+1)^{a+ T}}\frac{\Gamma(a + T)}{\Gamma(a)} \frac{1}{T!}$
\For{$i = 1 \ldots, M$}
\State Set $W^{(i)} = 1$, $S^{(i)} = 0$
\EndFor
\For{$\tau=1, \dots, T$}
\For{$i = 1, \ldots, M$}
\State Sample $c_V \sim q_{\tau, V}(c_V | S^{(i)}_{V})$
\State Sample $c_{\bar{V}}  \sim  p_{\tau}(c_{\bar{V}}  | c_V, S^{(i)})$ 
\State Set $c = (c_{V}, c_{\bar{V}})$
\State Set $u =  p_{\tau, V}(c_{V} | S^{(i)}) \frac{T - \tau + 1}{(X - S^{(i)}_{V})(c_{V})}$
\State Update $W^{(i)} \leftarrow W^{(i)} \times u$
\State Update $S^{(i)}(c) \leftarrow S^{(i)}(c) + 1$
\EndFor\label{particlefor}
\If{Resampling is on} \label{alg-step: adaptive resampling} \Comment{relevant in the adaptive version}
\State \label{alg-step: update Z} Update $Z \leftarrow Z \times \frac{1}{M} \sum_{i = 1}^{M} W^{(i)}$.
\State Resample particles: \label{alg-step: resample}
\[
\{ S^{(i)}, W^{(i)} = 1 \}_{i = 1, \ldots, M}  \leftarrow \textup{Resample}( \{ S^{(i)}, W^{(i)} \}_{i = 1, \ldots, M} )
\]
\EndIf
\EndFor\label{sisfor-3}
\State \textbf{return} the weighted samples $\{ S^{(i)}, W^{(i)} \}_{i = 1, \ldots, M}$.
\State \textbf{return} the marginal likelihood estimate $Z_{0} \times Z$.
\Comment{weighted samples for $\pi(S | X)$}
\Comment{estimate of $\mathcal{L}_{X}$}
\EndProcedure
\end{algorithmic}
\end{algorithm}

\subsection{Variational Algorithms} \label{sec: Variational Algorithms}
In this subsection, for the sake of completeness, we will develop variational inference methods for the allocation model. The variational techniques are well known in the context of NMF and related topic models \citep{Cemgil2009,  Paisley2014, Blei2003} and the derivations are technical but straightforward, so we omit most of the technical details and mainly state the results.

In the sequel, we will focus on the special case when the observations have the form of a contraction as in \eqref{eq: X as partially observed S}, i.e.\, $X = S_{\vi}$. This case corresponds to a hidden variable model, where entities corresponding to the hidden indices $\bar{V}$ are never observed. We will denote the target distribution of BAM, i.e., the full posterior distribution $p(S, \lambda, \Theta \mid X)$, as $\mathcal{P}$.

Variational Bayes (VB) \citep{beal2006variational} is a technique where a target posterior distribution $\mathcal{P}$ is approximated by a variational distribution $\mathcal{Q}$ via minimizing Kullback-Leibler divergence $\textup{KL}(\mathcal{Q}||\mathcal{P})$. In the context of Bayesian model selection, minimization of the $\textup{KL}(\mathcal{Q}||\mathcal{P})$ corresponds to establishing a tight lower bound for the marginal log-likelihood $\log \mathcal{L}_{X}$, which we refer as \textit{evidence lower bound} (ELBO) and denote by $\mathcal{B}_{\mathcal{P}}[\mathcal{Q}]$. This correspondence is due to the following decomposition of marginal log-likelihood
\begin{equation*}
\log \mathcal{L}_{X} = \mathcal{B}_{\mathcal{P}}[\mathcal{Q}] + \textup{KL} ( \mathcal{Q} \| \mathcal{P}) \geq \mathcal{B}_{\mathcal{P}}[\mathcal{Q}]
\end{equation*}
where the ELBO is explicitly defined as
\begin{equation}
\mathcal{B}_{\mathcal{P}}[\mathcal{Q}] \equiv \E[\mathcal{Q}]{\log \frac{\pi(\mathbf{S}, \lambda, \Theta) ~ \mathbb{I}(\mathbf{S}_{V} = X)}{\mathcal{Q}(\mathbf{S}, \lambda, \Theta)}}
\label{eq:elbo_definition}
\end{equation}

In a typical scenario of VB, variational distribution $\mathcal{Q}$ is assumed to be a member of a restricted family of distributions. In its most common form, also known as \textit{mean-field} approximation, $\mathcal{Q}$ is assumed to factorize over some partition of the latent variables, in a way that is reminiscent to a rank-one approximation in the space of distributions. For BAM, a natural choice of mean-field approximation is the family of factorized distributions of the form
\begin{equation*}
\mathcal{Q}(S, \lambda, \Theta) =  q(S) ~ q(\lambda,\Theta)
\end{equation*}

To minimize the $\textup{KL}(\mathcal{Q}||\mathcal{P})$, i.e., to maximize ELBO, one can utilize a fixed point iteration algorithm (with local optimum guarantees) on $q(S)$ and $q(\lambda,\Theta)$ where the updates are:
\begin{eqnarray}
q(S) & \propto & \exp\bigl(\E[q(\lambda, \Theta)]{\log \pi(\mathbf{S}, \lambda, \Theta) + \log \mathbb{I}(\mathbf{S}_{V} = X)} \bigr) \label{eq:vb_step_1}
\\
q(\lambda, \Theta) & \propto & \exp\bigl(\E[q(S)]{\log \pi(\mathbf{S}, \lambda, \Theta) + \log \mathbb{I}(\mathbf{S}_{V} = X)} \bigr)
\label{eq:vb_step_2}
\end{eqnarray}
and explicit evaluation of the equations above implies the following set of marginal variational distributions
\begin{align*}
q(S) & = \prod_{i_\vi} \mathcal{M}\bigl(S(:, i_\vi); X(i_\vi), \Phi_\vi(:, i_\vi) \bigr) \\
q(\lambda, \Theta) & = q( \lambda) \prod_{n = 1}^{N} q(\theta_{n \mid \pa{n}}) 
\end{align*}
where 
\begin{align*}
q(\lambda) & = \mathcal{GA}(\lambda; \hat{a}, b+1) \\
q(\theta_{n \mid \pa{n}}) & = \prod_{i_\pa{n}} \mathcal{D}(\theta_{n \mid \pa{n}}(:,i_\pa{n}); \hat{\alpha}_{\fa{n}}(:,i_\pa{n})) & \forall n \in [N].
\end{align*}
Hence, $\hat{a}$, $\hat{\alpha}$ and $\Phi_\vi$ are the variational parameters. Thereby the updates in \eqref{eq:vb_step_1} and \eqref{eq:vb_step_2} on the variational parameters are found as
\begin{align*}
\hat{a} & \gets a + \E[\mathcal{Q}]{\mathbf{S}_{+}}  \\
\hat{\alpha}_{\fa{n}}(:, i_{\pa{n}})) &\gets \alpha_{\fa{n}}(:, i_{\pa{n}}) + \E[\mathcal{Q}]{\mathbf{S}_{\fa{n}}(:, i_{\pa{n}})}, \quad\quad \forall n \in [N], \forall i_{\pa{n}}  \\
\Phi_\vi(i_{1:N}) &\leftarrow \frac{\exp\left(\sum_{n = 1}^{N} \E[\mathcal{Q}]{\log \theta_{n|\pa{n}}(i_{n}, i_{\pa{n}})} \right) }{\sum_{i_\vibar} \exp\left(\sum_{n = 1}^{N} \E[\mathcal{Q}]{\log \theta_{n|\pa{n}}(i_{n}, i_{\pa{n}})} \right) }, \quad\quad\quad \forall i_{1:N}
\end{align*}
Here the expectations are to be taken with respect to $Q$ with the most recently updated parameters. When there are no missing values in $X$, $S_+$ is known as $S_+ = \sum_{i_\vi} X(i_\vi)$. For $n\in [N]$, the expected sufficient statistics of the model
\begin{eqnarray*}
\E[{\mathcal{Q}}]{\mathbf{S}_{\fa{n}}(i_n, i_{\pa{n}})} = \sum_{i_{\overline{\fa{n}}}} \E[\mathcal{Q}] { \mathbf{S}(i_{1:N}) }
\end{eqnarray*}
need to be estimated. These sufficient statistics are various marginal statistics of the possibly intractable object
\begin{eqnarray*}
\E[\mathcal{Q}]{ \mathbf{S}(i_{1:N})} = X(i_{\vi}) \Phi_{\vi}(i_{1:N})
\end{eqnarray*}
Yet, as $\Phi_{\vi}$ respects a factorization implied by the DAG $\mathcal{G}$, and depending on the structure of the graph $\mathcal{G}$ and the set of visible indices $V$, it is possible to calculate the required sufficient statistics $\E[\mathcal{Q}]{ \mathbf{S}_{\fa{n}} }$ exactly by the junction tree algorithm. This is very attractive because we do not need to explicitly store or construct the tensor $\E[\mathcal{Q}]{\mathbf{S}}$ but only typically much lower dimensional clique potentials.

Once the expected sufficient statistics are estimated, the evidence lower bound (ELBO) in equation \eqref{eq:elbo_definition} yields 
\begin{eqnarray*}
e^{\mathcal{B}_\mathcal{P}[\mathcal{Q}]} & =& \frac{b^a}{(b+1)^{a + \mathsf{E}_{\mathcal{Q}} \{\mathbf{S}_+ \}}}\frac{\Gamma(a + \mathsf{E}_{\mathcal{Q}} \{\mathbf{S}_+ \})}{\Gamma(a)} \\
& & \left(\prod_{n = 1}^{N} 
\frac{B_n\left(\alpha_{\fa{n}} + \mathsf{E}_{\mathcal{Q}} \{\mathbf{S}_{\fa{n}}\} \right)}{B_n(\alpha_{\fa{n}})} \right) 
 \frac{\prod_{i_{1:N}} \Phi_{\vi}(i_{1:N})^{-\mathsf{E}_{\mathcal{Q}} \{\mathbf{S}(i_{1:N})\}}}{\prod_{i_\vi} \Gamma(X(i_\vi)+1)} 
\end{eqnarray*}
When we contrast the ELBO expression to the marginal allocation probability in \eqref{eq:pS1}, we see that the expressions are almost identical, with the $S$ replaced by expectations of form $\mathsf{E}_{\mathcal{Q}} \{ \mathbf{S} \}$, i.e., the `mean field' approximation. The last term can be written as 
\begin{eqnarray*}
\prod_{i_\vi}  \frac{  \left(\prod_{i_\vibar} \Phi_{\vi}(i_{1:N})
^{- \Phi_{\vi}(i_{1:N})} \right)^{X(i_\vi)}}
{\Gamma(X(i_\vi)+1)} = \prod_{i_\vi}  \frac{ \exp(H_{\Phi}(i_\vi))^{X(i_\vi)}}
{\Gamma(X(i_\vi)+1)}
\end{eqnarray*}
where $H_{\Phi}(i_{V}) = -\sum_{i_\vibar} \Phi_{\vi}(i_{1:N}) \log \Phi_{\vi}(i_{1:N})$ is the posterior entropy. The lower bound is large when the entropy of the posterior is large, discounted by the number of different ways the $X(i_\vi)$ tokens may have arrived.

\subsubsection{Alternative optimization methods for BAM}
As BAM is a conjugate hierarchical Bayesian model, some other standard optimization based methods are suitable for it. These method include EM, dual-EM, and iterative conditional modes (ICM), which is essentially a coordinate method. These algorithms solve different but closely related problems, where the key algorithmic difference becomes if posterior modes (such as $S^*$) or expectations (such as $\E{S}$) are computed during iterations. Table \ref{tbl: Other optimization methods for BAM} below succinctly lists the available algorithms for BAM, where we partition the parameters into two groups as $S$, and $(\lambda, \theta)$.
\begin{table}[ht]
\normalsize
\caption{Other optimization methods for BAM}
\begin{center}
\begin{tabular}{|c|l|l|}\hline
\textbf{Method} & \textbf{Iterative updates} & \textbf{Objective} \\
\hline \hline
ICM & $\lambda^*, \Theta^* \leftrightarrow S^* $ &  $\arg\max_{\lambda,\Theta, S} \mathcal{P}(S, \lambda, \Theta )$     \\\hline
EM  & $\lambda^*, \Theta^* \leftrightarrow \mathsf{E}_{\mathcal{P}}\{\mathbf{S}\} $ &  $\arg\max_{\lambda,\Theta} \sum_S \mathcal{P}(S, \lambda,\Theta )$  \\\hline
Dual-EM & $\mathsf{E}_{\mathcal{P}} \{\lambda, \Theta \} \leftrightarrow S^* $ & $\arg\max_{S} \int d\lambda d\Theta  \mathcal{P}(S, \lambda, \Theta )$  \\\hline
VB & $\E{\lambda, \Theta} \leftrightarrow \mathsf{E}_{\mathcal{Q}}\{\mathbf{S}\} $  & $\arg\max_{q(S), q(\lambda, \Theta) }  \mathcal{B}_\mathcal{P}[\mathcal{Q}]$   \\\hline
\end{tabular}
\end{center}
\label{tbl: Other optimization methods for BAM}
\end{table}

\section{Simulation Results} \label{sec: Simulation Results}
In this section, our goal is to evaluate the algorithms developed in Section \ref{sec: Inference by Sequential Monte Carlo} on two tasks: 
i) computing the marginal likelihood, and ii) computing approximate decompositions of count tensors. We will compare variational approximations with sequential Monte Carlo to show the relative merits and weaknesses of each approach. Besides key indicators such as memory requirement and computational efficiency in terms of run time, we will investigate the \SMC approach on model scoring and parameter estimation. Throughout the section, the \SMC method refers to BAM-SIS-R in Algorithm \ref{alg:sisR}, and VB refers to the mean-field approximation as described in Section \ref{sec: Variational Algorithms} and derived in detail in the Appendix.

Rank estimation or more generally model order selection is a prevalent problem in matrix and tensor factorization models. From a Bayesian perspective, Bayes factors \citep{Raftery1995}, that is the ratio of the marginal likelihoods of two alternative models, provide a systematic and principled approach for model comparison. For the comparison of multiple models, one can use the closely related quantity ``posterior log odds", which corresponds to the normalized values of the marginal log-likelihoods for each model, assuming a uniform prior on the models. A central question for this section is: Given several alternative models and parameter regimes, which algorithms should be preferred for calculating the marginal likelihood for Bayesian model comparison? Moreover, how do complexities of those algorithms change with respect to variable dimensions and number of tokens? In Section \ref{subsec:synth}, we use synthetic data experiments to provide answers to these questions. In Section \ref{subsec:causal}, we present an application of model selection using BAM for causality inference, as well as providing an examination of the nature of the latent variables inferred by the \SMC algorithm. The experiments in Section \ref{subsec:letter} demonstrate that BAM framework can be extended to conduct model selection between models with vs.\ without parameter tying. Lastly, in Section \ref{subsec:decomposition}, we investigate the nature of decompositions computed by \SMC and VB of BAM and contrast them to one obtained via an optimization approach.

\subsection{Synthetic Data Experiments} \label{subsec:synth}
In this section, we carry out a comparison for the calculation of the marginal likelihood and posterior log odds. It is known that SMC (without adaptive resampling) is unbiased, in contrast to variational methods, in estimating the marginal likelihood. Hence, our aim is to numerically explore the parameter regimes where an SMC-based marginal likelihood estimation may be preferable to a variational approximation to the marginal likelihood. We start our exploration with experiments on small to data to feasibly compute ground truth solutions for comparison.

\subsubsection{Comparison of SMC, VB and exact enumeration on toy data}

\begin{figure}[htbp]
	\centering
	\begin{subfigure}[t]{0.95\textwidth}
		\includegraphics[width=1.\textwidth]{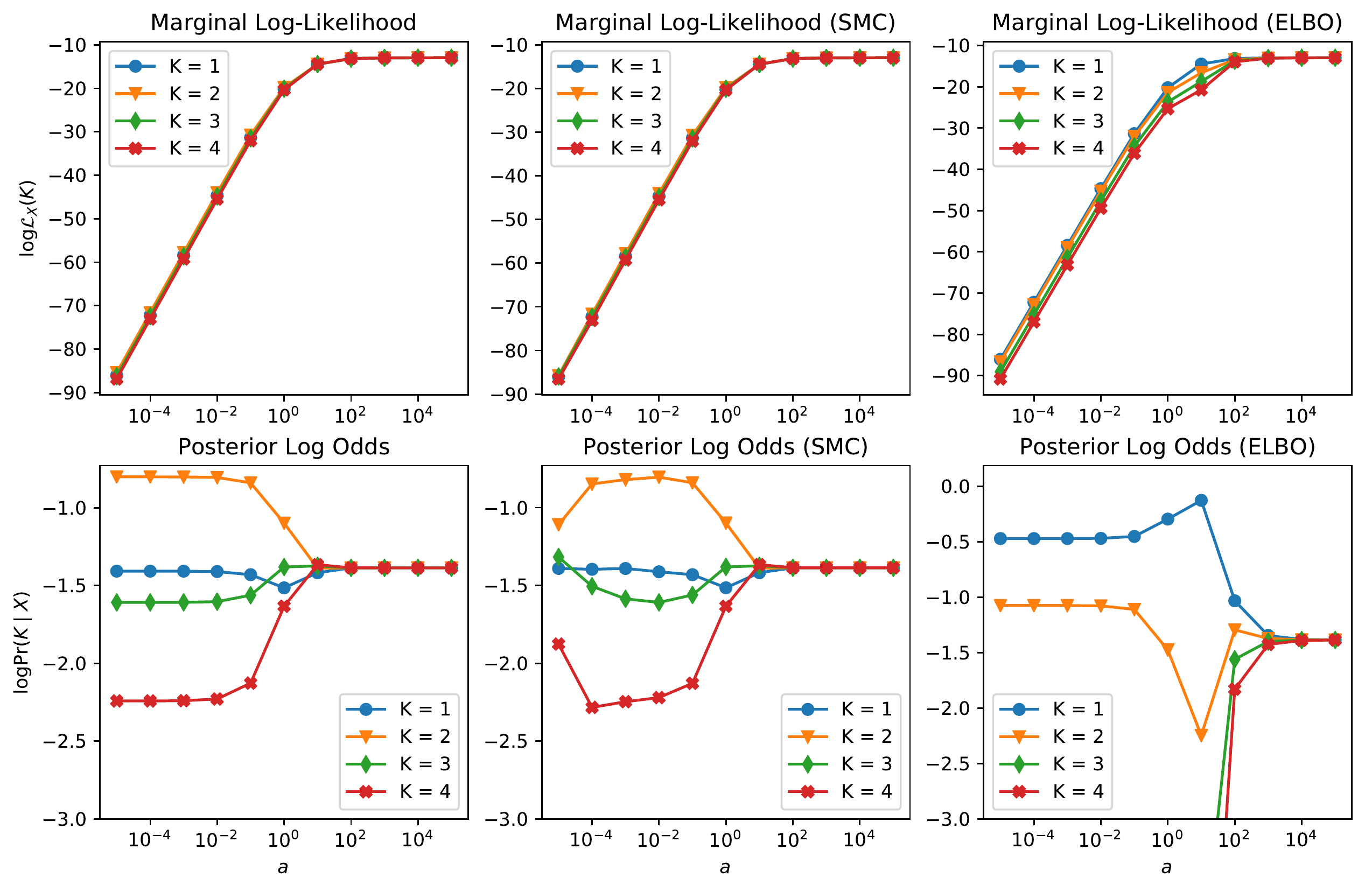}
		\caption{Marginal likelihood estimations (or lower bounds) of both \SMC and VB algorithms seem accurate for the matrix $X^{(1)}$. However, VB fails to identify the model order $K$ as opposed to \SMC.}
		\label{fig:mg-lkhd-1}
	\end{subfigure}	\\
	\begin{subfigure}[t]{0.95\textwidth}
		\includegraphics[width=1.\textwidth]{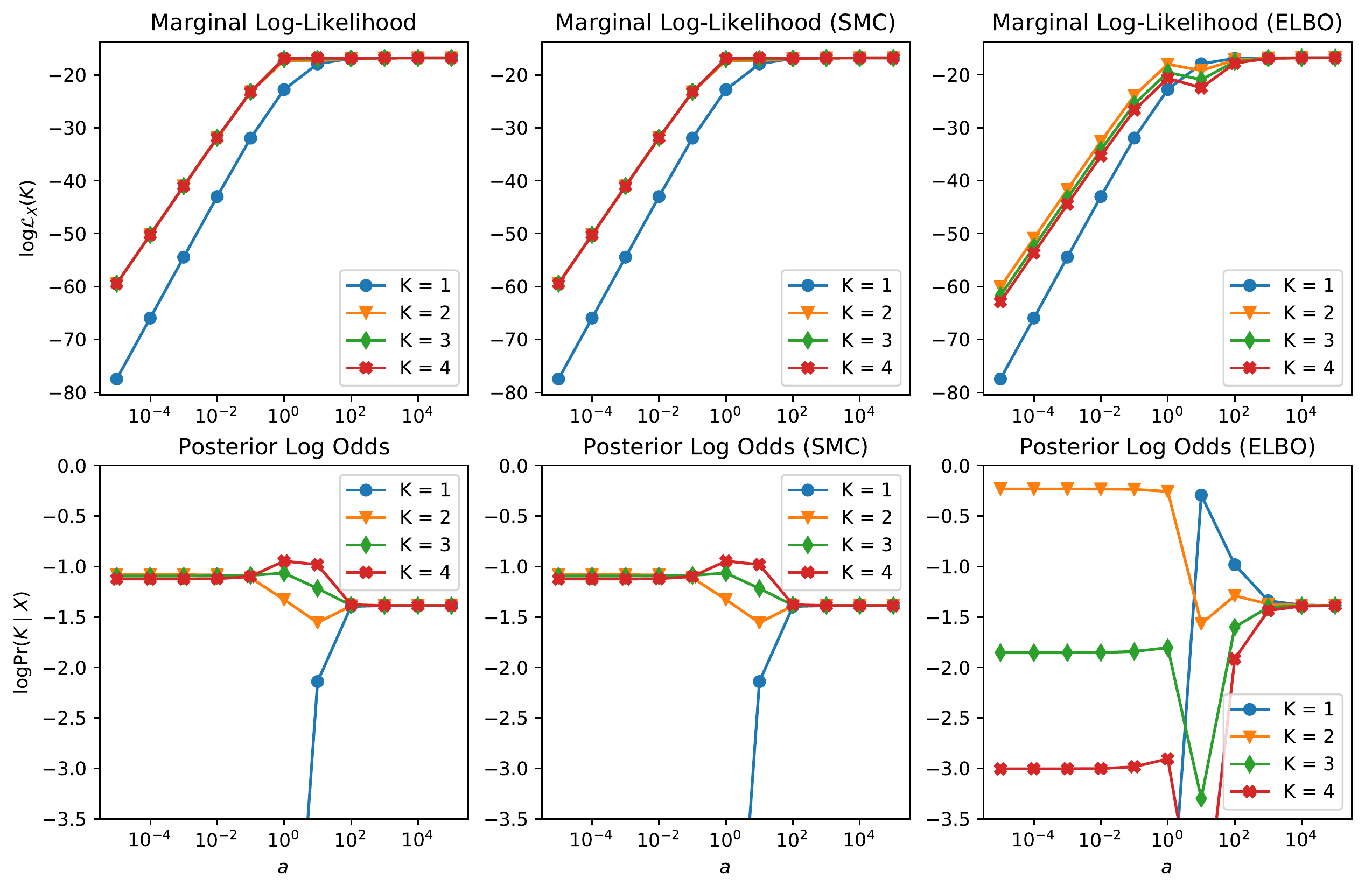}
		\caption{Marginal likelihood estimations of \SMC is highly accurate for the matrix $X^{(2)}$. Both \SMC and VB algorithms are able to identify the model order $K$ as in exact calculation.}
		\label{fig:mg-lkhd-2}
	\end{subfigure}		
	\caption{Marginal log-likelihood estimations of \SMC and VB compared to exact marginal log-likelihood.}
	\label{fig:mg-lkhd}
\end{figure}

To compare the accuracy of marginal likelihood estimations by VB and \SMC, we consider KL-NMF/LDA model (Figure \ref{fig:KL-NMF}) and select two small toy matrices $X^{(1)}$ and $X^{(2)}$ for which it is feasible to calculate the exact marginal likelihood by exhaustive enumeration. Moreover, we wish to see how the approximation affects the relative ordering of models of different ranks according to the posterior log odds. As the model selection is highly dependent on the prior parameters, we will investigate the behaviour of the marginal likelihood and log odds according to the equivalent sample size parameter $a$.
The toy matrices are
\begin{align*}
X^{(1)} & = \left(\begin{array}{cccc} 
    2 & 1 & 1 & 0 \\
    0 & 0 & 1 & 2 \\
    0 & 0 & 1 & 1 
    \end{array} \right) &
X^{(2)} & = \left(\begin{array}{ccc} 
    4 & 3 & 0 \\
    0 & 0 & 3 \\
    0 & 0 & 3 
    \end{array} \right) &
\end{align*}
are chosen to have certain intuitive properties: The first matrix $X^{(1)}$ 
could be a draw from a uniform $\Theta$, i.e. each $\theta_{n \mid \pa{n}}(i_n,i_\pa{n}) \approx 1/I_n$,  
where the model will be of rank $K=1$, corresponding to independence of the indices $i$ and $j$. Alternatively, we can have a model of order $K=2$ with two topics, one for the first two columns where only word $1$ is active and one for the last column where words $2, 3$ are active, with the third column being a superposition of the two. 
The second toy matrix $X^{(2)}$ does not appear to be a draw from an uniform $\Theta$ and is more clustered and could be conveniently described by a model of rank $K=2$.

In Figure \ref{fig:mg-lkhd}, we illustrate the results of the exact algorithm and \SMC and algorithms, for different model orders $K=1,\dots,4$ and over the range $a \in [10^{-5}, 10^{5}]$. This range for $a$ captures both the sparse and the dense prior regimes, corresponding to small and large $a$ values respectively. 

The \Polya urn interpretation of the allocation process indicates that the allocations drawn from relatively bigger initial counts are affected negligibly by the previous draws and mostly determined by the initial configuration. Hence the flat priors on $\Theta$ with large $a$ parameter result in allocation tensors where the tokens are allocated uniformly. This effect of the parameter $a$ is depicted on the Figures \ref{fig:mg-lkhd-1} and \ref{fig:mg-lkhd-2}.

Since the entries of $X^{(1)}$ could be interpreted as being drawn from a uniform categorical distribution, its marginal likelihood $\mathcal{L}_{X^{(1)}}(K)$ increases monotonically with the increasing $a$; see Figure \ref{fig:mg-lkhd-1}. However, in Figure \ref{fig:mg-lkhd-2} we observe a different behaviour for $X^{(2)}$: Increasing $a$ decreases the marginal likelihood after some point, hence indicates $X^{(2)}$ is less likely to be drawn from a balanced \PolyaBayes process. 

Another observation about the Figures \ref{fig:mg-lkhd-1} and \ref{fig:mg-lkhd-2} is that the model selection may highly be affected by the chosen value of $a$. For instance, our simulation results show that bigger values of $a$ favor a larger model order $K$. This behaviour of $a$ is also not surprising in the context of matrix factorization. Model parameters $\theta_{n \mid \pa{n}}$'s are normally defined on unit simplices, but as we increase $a$, we effectively assign zero prior probability to the vectors that are distant from the center. Therefore, increasing the value of parameter $a$ is practically equivalent to restricting the domain of the basis $\theta_{n \mid \pa{n}}$'s. If we view the model order $K$ as the rank of the matrix decomposition, then it is natural to expect bigger rank when the domain of the basis is restricted.

In the second part of the experiment, we compared the marginal likelihood estimations of \SMC and VB algorithms to exact marginal likelihood values. For each rank $K$ and prior parameter $a$, we used $1000$ particles for \SMC and ran both of the \SMC and VB algorithms $100$ times to obtain the final estimation. In the case of SMC, we report the average of the estimations whereas in VB we report the maximum of the evidence lower bounds as the final estimation of marginal likelihood. 

It is clear from the Figure \ref{fig:mg-lkhd}, the marginal likelihood estimations of \SMC are highly accurate for both of the matrices $X^{(1)}$ and $X^{(2)}$ (Figures \ref{fig:mg-lkhd-1} and \ref{fig:mg-lkhd-2}). Regardless of the magnitude of $a$, \SMC is able to correctly identify the model order $K$ as in exact calculation. However, estimations of VB in the sparse data regime are less accurate and it is not able to identify the correct model order for the matrix $X^{(1)}$ (Figure \ref{fig:mg-lkhd-1}). Likewise, posterior log odds estimations in the sparse regime highly differ from the ground truth for both matrices (Figures \ref{fig:mg-lkhd-1} and \ref{fig:mg-lkhd-2}).

\subsubsection{Model order scoring for PARAFAC}

As we discussed in previous sections, model selection may highly depend on the hyperparameter $a$. However, in this section we will analyze the model selection capabilities of our algorithms, and we will not struggle with tuning the parameter $a$, rather its value will be assumed fixed and known. Hence the only unknown parameter we will be dealing with is the generative model itself. 

\begin{figure}[htbp]
	\centering
	\begin{subfigure}[t]{0.48\textwidth}
		\includegraphics[width=1.\textwidth]{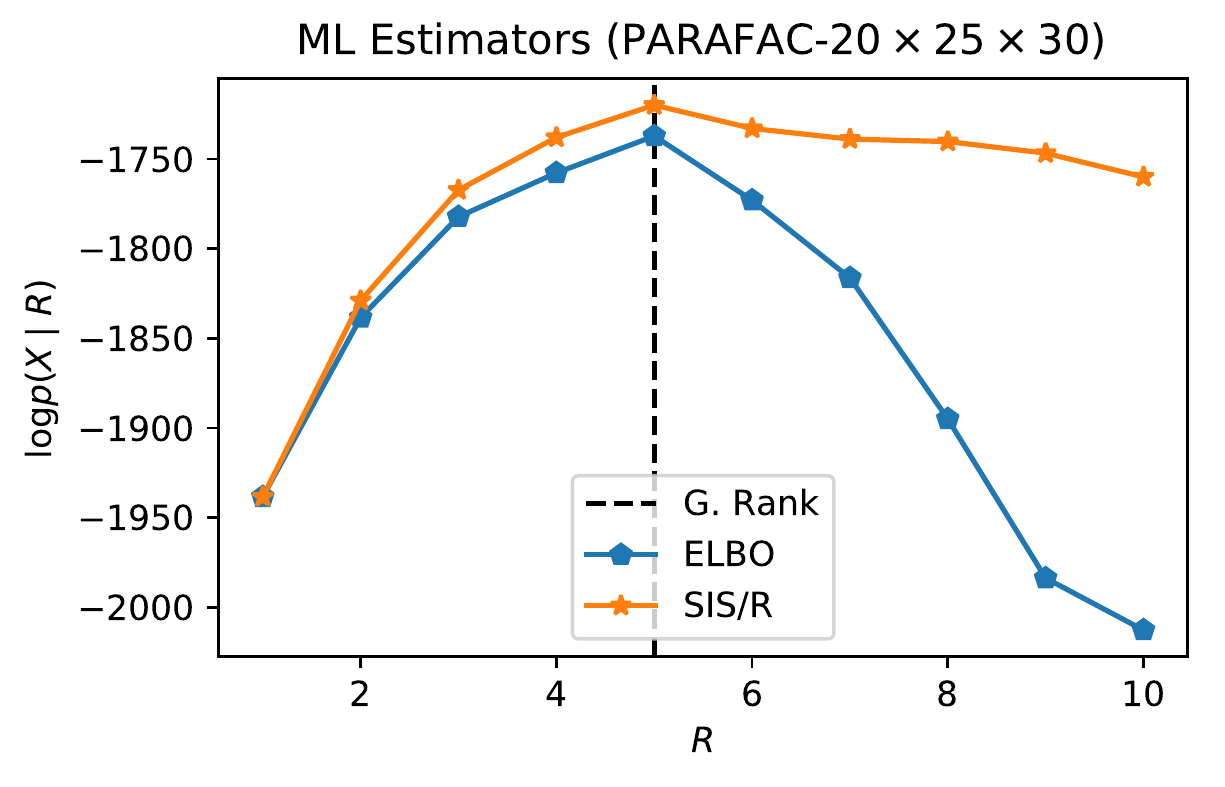}
		\caption{Model order selection for PARAFAC.}
		\label{fig:cp-model-selection}
	\end{subfigure}
	~
	\begin{subfigure}[t]{0.45\textwidth}
		\includegraphics[width=1.\textwidth]{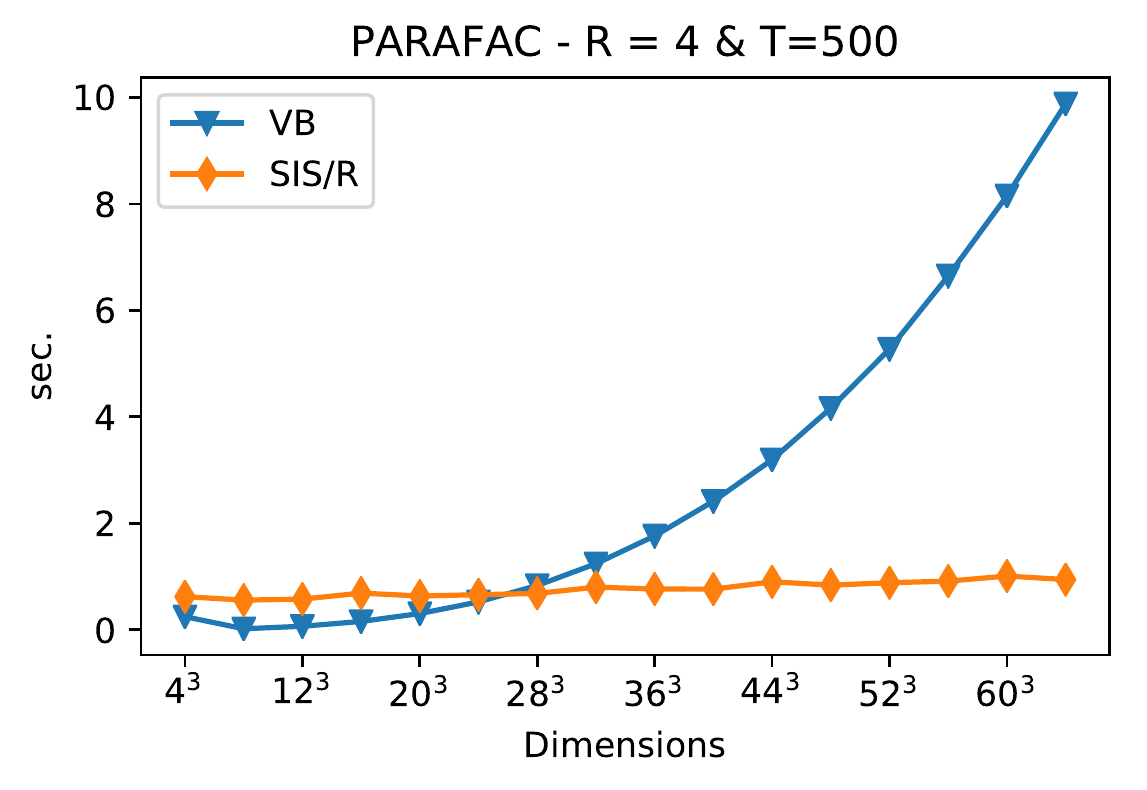}
		\caption{Comparison of runtimes under fixed $T$.}
		\label{fig:runtimes}
	\end{subfigure}
	\caption{Comparison of VB and \SMC in terms of average runtime and model selection on simulated data.}
\end{figure}

In the first part of the experiment, we generated a $20 \times 25 \times 30$ tensor $X$ from the PARAFAC model with rank $R=5$, number of tokens $T = 500$ and prior parameter $a=30$. Here rank refers to the cardinality of the latent index $r$ in Figure \ref{fig:naive-bayes}. Our goal is to identify the rank $R$ of the decomposition, i.e., our hypothesized models are the Naive Bayes Models as in Figure \ref{fig:naive-bayes} each having different cardinality for the latent variable $r$. Figure \ref{fig:cp-model-selection} shows the marginal likelihood and ELBO estimations of \SMC and VB for various ranks $R$. Although we are not able to calculate the true marginal likelihood for even small $R$ and compare it with our estimations due to intractability, both \SMC and VB estimate maximum $\mathcal{L}_{X}(R)$ at true $R$ as one might hope, however it should be noted that VB seems to deviate below at higher $R$ due to the gap between true marginal likelihood and ELBO, i.e., KL divergence between true posterior distribution and variational distributions.

To support our claim about the time complexity of the \SMC algorithm being independent of the size of $X$, we also compared the runtimes of the VB and \SMC algorithms for the different sizes of the observed tensor $X$ ranging from $4 \times 4 \times 4$ to $64 \times 64 \times 64$. For each of sizes, we first generated $X$ by sampling $T=1000$ index configurations from the Naive Bayes Model (PARAFAC) in Figure \ref{fig:naive-bayes} and then ran both \SMC and VB algorithms on the same tensors $X$ until the convergence. We repeated this process 10 times and reported average runtime results in Figure \ref{fig:runtimes}. Our findings support our algorithm analyses that the time complexity of VB scales with the size of $X$ whereas the time complexity of \SMC (without parallelization) is independent from it. 

Our conclusion in this section is it is possible to do reliable model selection using both popular variational algorithms and our proposed \SMC algorithm. However, each algorithm has favorable properties in different data regimes, which make them complementary alternatives. For instance, the time complexity of \SMC is independent of the size of $X$ which makes it a promising choice in the sparse data regime, while the time complexity of the variational algorithms are independent of the total number of tokens making them preferable in the dense data regime.
\subsection{Model Selection for Causal Inference}
\label{subsec:causal}

Deciding whether the relationship between two or more variables is truly causal or spurious, i.e. explainable through the existence of a common hidden cause variable, can be cast as a model selection problem \citep{Heckerman1995, cai2018, Kocaoglu2018}; in this section we will apply our methodology to make a decision of this nature on the popular Abalone data set. The Abalone data set is a collection of observations including physical measurements, sex, and age from the marine animal abalone \citep{dua2017}. All measurements except sex are real valued, thus we will be categorizing them while conducting our analysis with BAM. 

\begin{figure}[h!]
\begin{subfigure}[t]{0.47\textwidth}
\centering
\begin{tikzpicture}[>=stealth',auto,scale=0.6, every node/.style={transform shape}]        
        \node[obs,minimum size=1.6cm]                    (i1)     {Length};
        \node[obs, right=of i1,minimum size=1.6cm]       (i2)     {Diameter};
        \node[const, right=of i2,minimum size=1.6cm]     (in)     {\Huge $\dots$};
        \node[obs, right=of in,minimum size=1.6cm]       (iN)     {S. Weight};

        \edge {i1}{i2} ;
        \edge {i2}{in} ;
        \edge {in}{iN} ;
        
        \path[->,every node/.style={font=\sffamily\small}] (i1) 
        edge[bend left] node [left] {} (in);
        \path[->,every node/.style={font=\sffamily\small}] (i1) 
        edge[bend right] node [right] {} (iN);
        \path[->,every node/.style={font=\sffamily\small}] (i2) 
        edge[bend left] node [left] {} (iN);
\end{tikzpicture}
\caption{Complete Graph}
\end{subfigure}
\qquad
\begin{subfigure}[t]{0.47\textwidth}
\centering
\begin{tikzpicture}[>=stealth',auto,scale=0.6, every node/.style={transform shape}]        
        \node[obs,minimum size=1.6cm]                    (i1)     {Length};
        \node[obs, right=of i1,minimum size=1.6cm]       (i2)     {Diameter};
        \node[const, right=of i2,minimum size=0.8cm]     (in)     {\Huge $\dots$};
        \node[obs, right=of in,minimum size=1.6cm]       (iN)     {S. Weight};

        \node[latent, above right=of i2,minimum size=1.6cm]       (r)     {Cause};

        \edge {r}{i1,i2,in,iN} ;
\end{tikzpicture}
\caption{CP/PARAFAC Model}
\end{subfigure}
\caption{Graphical models that correspond to truly causal relationship, the model with the complete graph (left) vs. spurious relationship, CP/PARAFAC model (right) for abalone physical measurement variables.}
\label{fig:cause}
\end{figure}
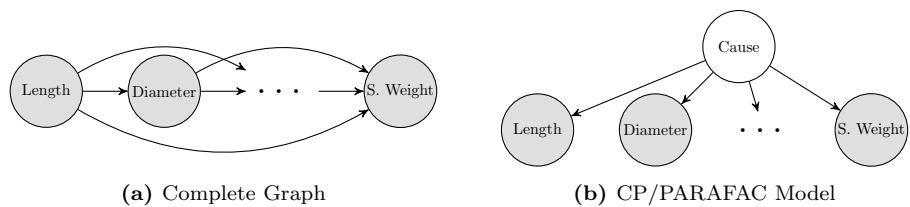

In the Abalone data set one can surmise that all seven variables that pertain to physical magnitude (i.e. size, weight) should be better explained by a model involving a common hidden cause variable, rather than a model defined by a complete graph where the different physical variables are assumed to cause each other. From a Bayesian model selection perspective, deciding between the two hypotheses corresponds to deciding between the two graphical models, that are shown in Figure \ref{fig:cause}. The first model corresponds to a complete graph (CG) model with no latent variables and the alternative is the CP/PARAFAC, i.e. the naive Bayes model. Without going into a detailed discussion of causal inference \citep{pearl2009causality, mooij2016distinguishing}, we will focus only on the Bayesian model selection aspects.

We use all seven physical magnitude variables from the Abalone data set to apply our methodology: length, diameter, height, whole weight, shucked weight, viscera weight, shell weight. Given the fact that our method works with categorical data and the original data includes non-categorical data, we categorized each non-categorical variable into 5 clusters, using $k$-means clustering separately on each variable. We conduct and present our simulations with two hyperparameter settings: $a = 1$ and $0.001$. In each experiment, to compare the two models we compare the marginal likelihood estimate produced by \SMC (as described in Algorithm \ref{alg:sisR}) and ELBO produced by VB (as described in in Section \ref{sec: Variational Algorithms}) for the CP model with the marginal likelihood for the CG model. The marginal likelihood estimates for the CP model is presented for all model orders between $1$ and $30$. Since there are no latent variables in the CG model, its marginal likelihood can be analytically calculated and does not vary according to the model order. Based on the graphical model presented at Figure \ref{fig:cause} (left), the marginal likelihood for the CG can be computed as below:
\begin{eqnarray*} 
\mathcal{L}_{X} = \frac{b^{a}}{(b+1)^{a+ X_+}}\frac{\Gamma(a + X_+)}{\Gamma(a)} \left( \prod_{n = 1}^{N} \frac{B_n( \alpha_{\fa{n}} + X_{\fa{n}} )}{B_n( \alpha_{\fa{n}}  )}  \right) 
\frac{1}{\prod_{i_{1:N}} X({i_{1:N}})!}
\end{eqnarray*}

\begin{figure}[ht!]
	\centering
	\begin{subfigure}[t]{0.45\textwidth}
		\includegraphics[width=1.\textwidth]{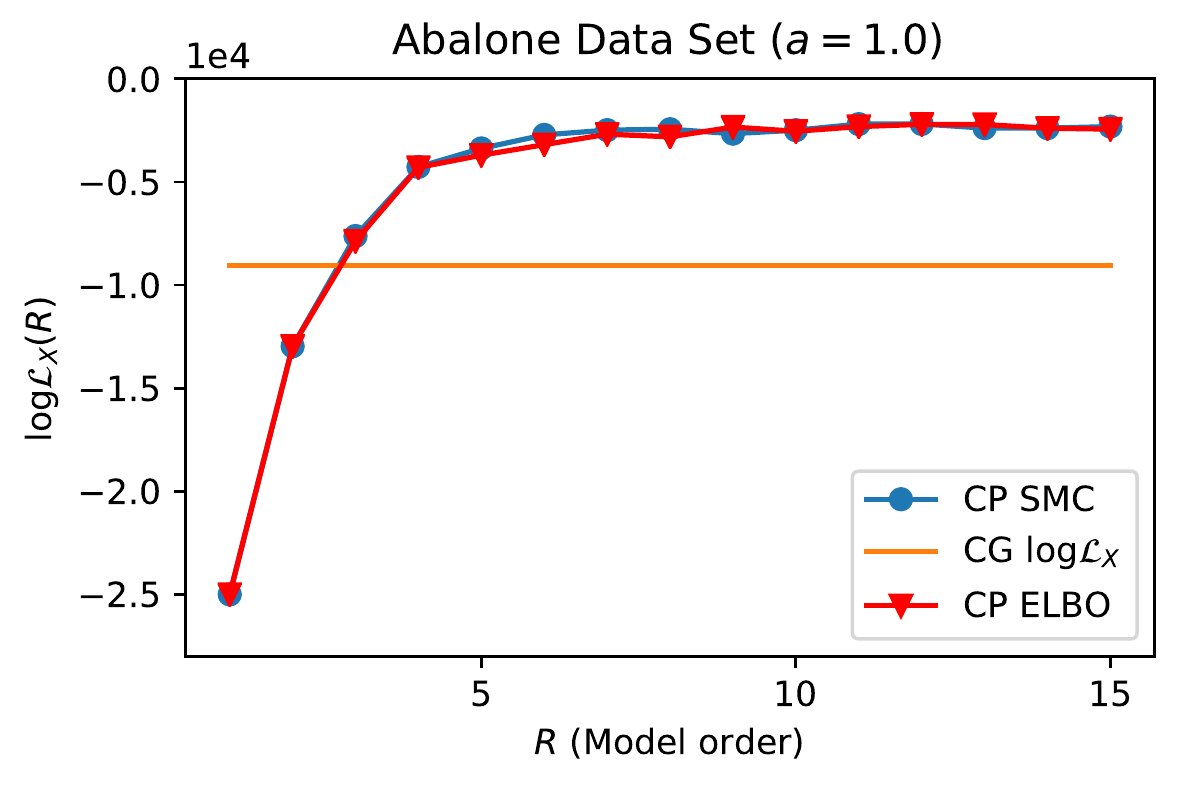}
	\end{subfigure}	
	\begin{subfigure}[t]{0.45\textwidth}
		\includegraphics[width=1.\textwidth]{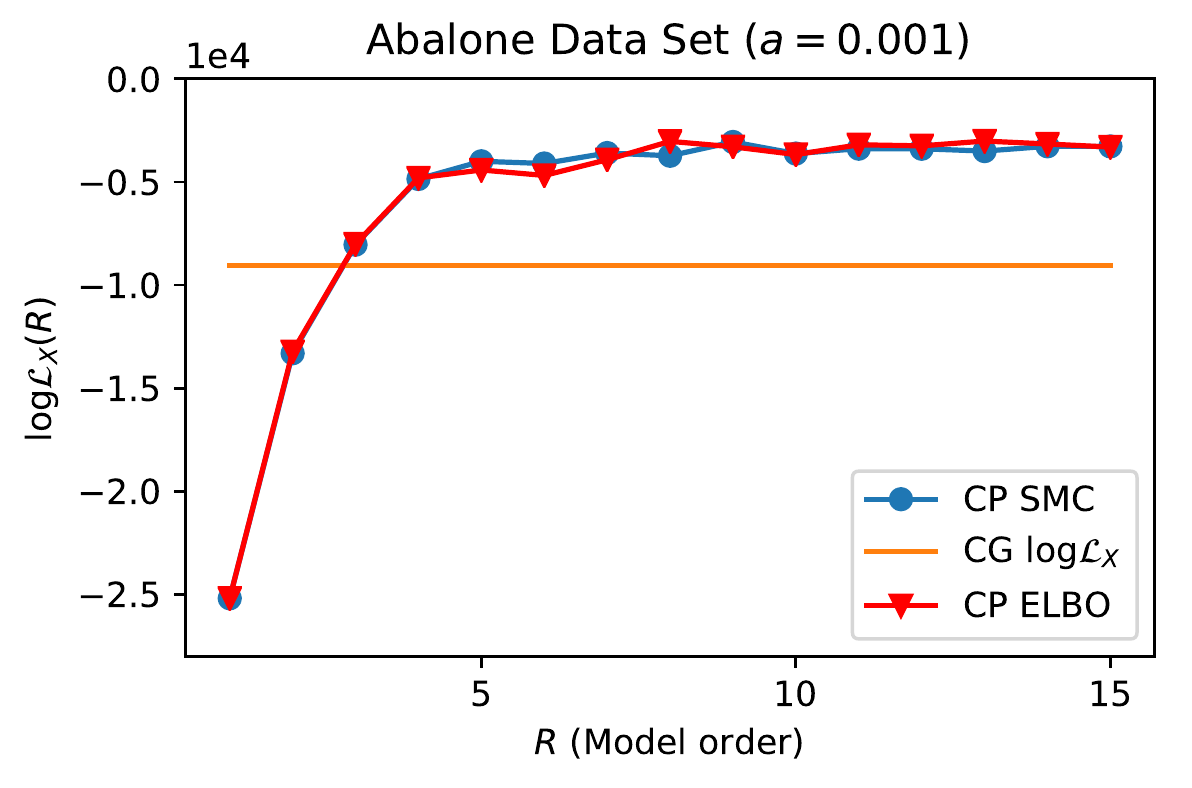}
	\end{subfigure}	
	\caption{Marginal likelihood for both hypotheses for abalone physical magnitude variables for $a = 1$ (left), and $a = 0.001$ (right). CP \SMC and CP ELBO correspond to the estimates by \SMC and VB for CP respectively. CG $\log \mathcal{L}_{X}$ corresponds to the marginal likelihood value calculated for the CG.}
	\label{fig:abalone}
\end{figure}

The results of the experiments can be seen in Figure \ref{fig:abalone}. The results presented show that for both hyperparameter settings, the CP model is a better explanation for the data. This is because for all model orders larger than $2$ for the case of $a = 1$ and $a = 0.001$, the marginal likelihood estimate by \SMC and VB for CP model is larger than the marginal likelihood for the CG model. Our results therefore demonstrate that all physical variables are more likely to be caused by a latent variable than to function as causes for each other.

\begin{figure}[h!]
	\centering
	\includegraphics[width=0.8\textwidth]{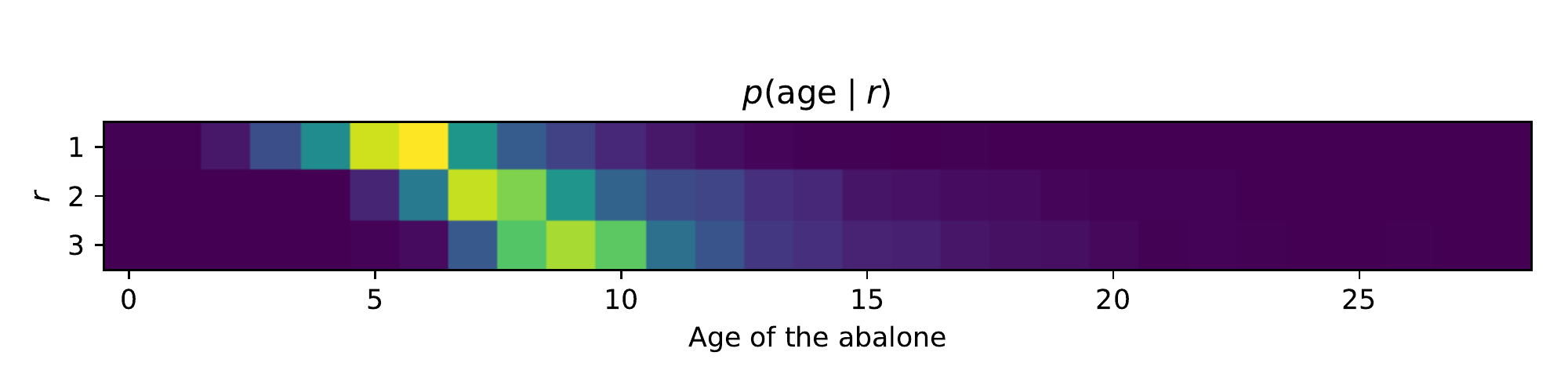}
	\caption{The distributions $p(\text{age}|r)$ of tokens belonging to different latent variable affiliations ($r$), with model order $R=3$ and $a = 1$. The figure shows that abalones belonging to different latent variable levels demonstrate different age distributions. The examination was conducted on a single particle obtained from \SMC. The particle with the most likely decomposition at the end of the procedure was used.}
	\label{fig:p_age_k}
\end{figure}

An immediate examination of interest is the nature of the latent variable produced by the \SMC algorithm. A reasonable candidate for this latent cause variable is the age of the abalone. If this is indeed the case, we expect the abalones allocated to different levels of the latent variable to differ from each other according to age. To see whether this is the case, we examine the age 
distributions of the abalones that are allocated to different levels of the latent variable - the dataset includes a variable, number of rings, that represents the age of the abalone. We present such an examination for $a = 1$ and model order $R = 3$ in Figure \ref{fig:p_age_k}. The rows of the figure illustrate the distributions $p(\text{age}|r)$. As can be seen, the age distributions for each level of the hidden variable is different, potentially corresponding to young, medium, and older aged abalones. Therefore we can conclude that the physical measurements are better explained by a common latent cause variable, which is likely to be the age of the abalone.

\subsection{Letter Transition Data} \label{subsec:letter}

\begin{figure}[h!]
\begin{subfigure}[t]{0.4\textwidth}
\centering
\begin{tikzpicture}[>=stealth',auto,scale=0.7, every node/.style={transform shape}]        

        \node[obs]                       (i)     {$i$};
        \node[latent, right=of j]        (k)     {$r$};
        \node[obs, right=of k]           (j)     {$j$};

        \node[latent,above=of j]         (t1)     {$\theta_{3|1}$};
        \node[latent, above=of k]        (t2)     {$\theta_1$};
        \node[latent, above=of i]           (t3)     {$\theta_{2|1}$};

        \node[draw=none, above=of t2]           (a)     {\large$\alpha$};
           
        \edge {k} {j};
        \edge {k} {i};
        \edge {t1} {j};
        \edge {t2} {k};
        \edge {t3} {i};
        \edge {a}{t1} ;
        \edge {a}{t2} ;
        \edge {a}{t3} ;
\end{tikzpicture}
\end{subfigure}
\quad
\begin{subfigure}[t]{0.4\textwidth}
\centering
\begin{tikzpicture}[>=stealth',auto,scale=0.7, every node/.style={transform shape}]        

        \node[obs]                       (i)     {$i$};
        \node[latent, right=of i]        (k)     {$r$};
        \node[obs, right=of k]           (j)     {$j$};

        \node[latent, above=of k]        (t2)     {$\theta_1$};
        \node[draw=none, left=of t2]           (t3)     {\large$\alpha$};
        
        \node[latent, above=of t2]           (a)     {\large$\theta_{2\mid1}$};
           
        \edge {a} {i};
        \edge {a} {j};
        \edge {k} {i};
        \edge {k} {j};
        \edge {t3} {t2};
        \edge {t3} {a};
        \edge {t2} {k};
        
\end{tikzpicture}
\end{subfigure}
\caption{Non-symmetric (left) and symmetric  (right) NMF models' graphical representation.}
\label{fig:symnonsym}
\end{figure}
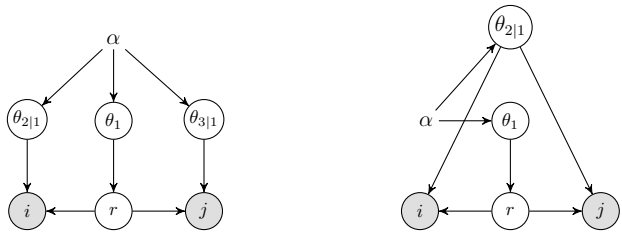

The BAM framework also allows inference in models where parameters of different probability tables are tied. An example to parameter tying would be modeling the data using NMF with the graphical model $i \leftarrow k \rightarrow j$, where both conditional probability tables are forced to be the same instead of being modeled as independent. We will provisionally call such a model \emph{symmetric NMF (sNMF)} for obvious reasons. The graphical models that depict the ordinary non-symmetric model, NMF, and the symmetric model, sNMF, can be seen in Figure \ref{fig:symnonsym}. From a matrix decomposition perspective, these models would correspond to the factorizations $WH \approx X$ vs.\ $W W^\text{T} \approx X$ respectively. The \SMC derivation for the sNMF is provided in the appendix. 

An example data for comparing the two models would be the letter transition data obtained from \citet{norvig2013}. Here, non-symmetric NMF would correspond to representing a letter differently according to whether it is a preceding letter of a pair vs. the following. The symmetric model, sNMF, is oblivious to the order of the letters, but models the co-occurrence of the letters instead. The actual data set includes an approximate total of $2.8 \times 10^{12}$ transitions. Examining this transition matrix, it is obvious that the data cannot be explained better by a symmetric model, given that the original matrix is not approximately symmetric. This is unsurprising given that sounds or letters in a language follow each other according to certain regularities which are not normally expected to be order independent. However, making this inference would be much more challenging if only a tiny sample from this data was available. We use our framework to compare the two models in this much noisier setting. We obtain a sample from the data by normalizing the counts of the original matrix, use it as the parameter of a multinomial distribution and draw $2000$ tokens from this distribution to conduct our experiments. 

Here we want to answer three questions: 1- How well does \SMC fare in choosing the correct model for the data; 2- How does VB compare to \SMC; 3- How the selection of hyperparameters affects these results. Instead of choosing a specific model order to make this comparison, we will be conducting the experiments for all cardinalities between $1$ and $50$. In this way, we will be able to observe whether model order affect the decision between the two models, and if it does, how it does so. We will be conducting the experiments at the hyperparameter settings of $a = 1$, and $0.001$.

\begin{figure}[h!]
	\centering
	\begin{subfigure}[t]{0.45\textwidth}
		\includegraphics[width=1.\textwidth]{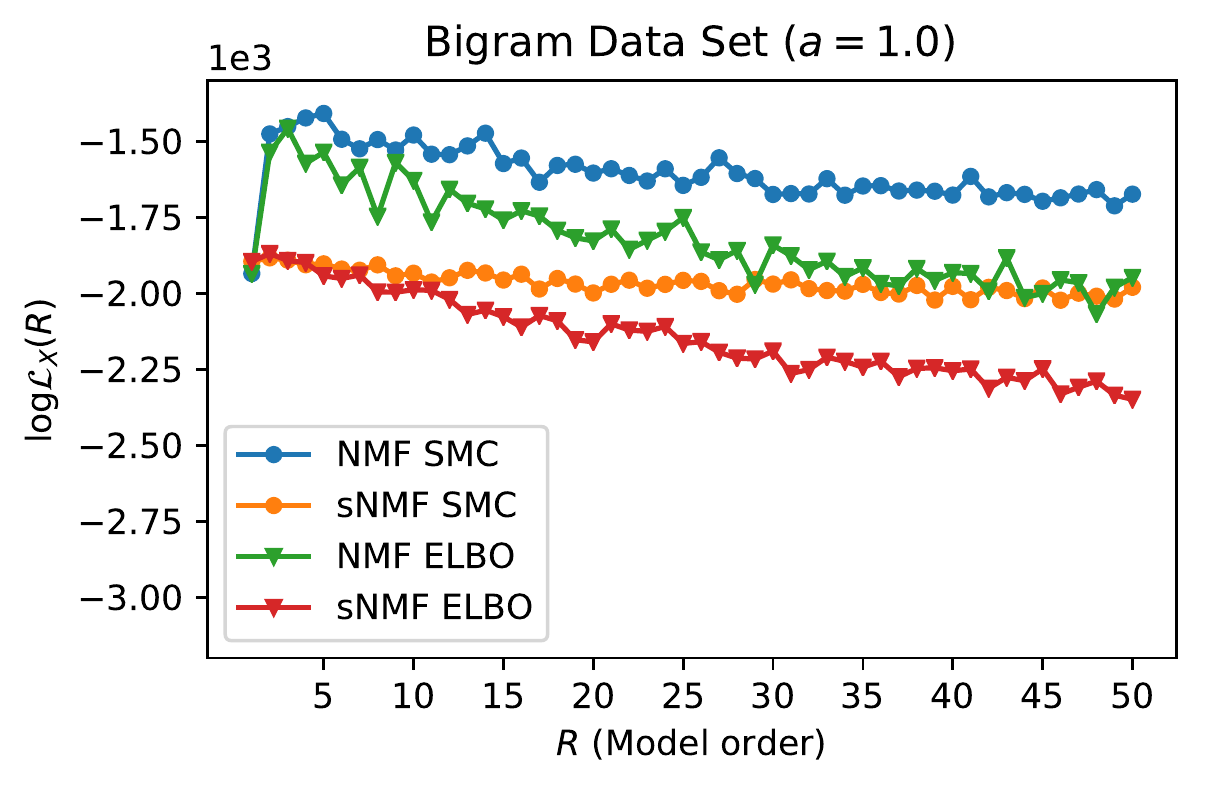}
		\label{fig:bigram_a_1}
	\end{subfigure}	
	\begin{subfigure}[t]{0.45\textwidth}
		\includegraphics[width=1.\textwidth]{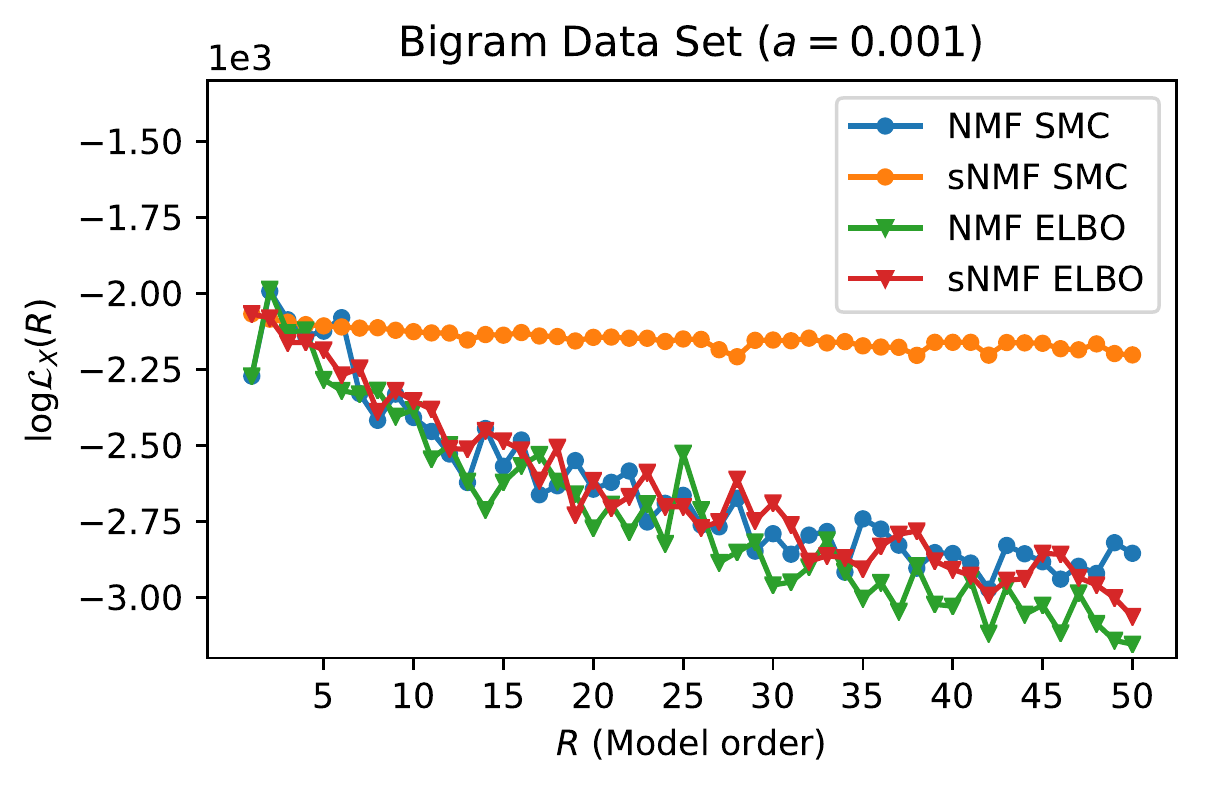}
		\label{fig:bigram_a_0_001}
	\end{subfigure}	
	\caption{Comparing symmetric vs. non-symmetric NMF marginal likelihood using \SMC and VB for $a = 1$ (left) and $a = 0.001$ (right).}
	\label{fig:sym_ml}
\end{figure}

The results for $a = 1$ are depicted in Figure \ref{fig:sym_ml}. The first observation is that at this level, the scores obtained from \SMC for NMF dominate those obtained for sNMF at every model order. Therefore, for this parameter setting, \SMC favors the non-symmetrical model unequivocally. 
This is paralleled by the ELBO scores obtained from VB which also favor NMF over sNMF at every model order. This shows that at least for the given problem, VB would also be a sufficient inferential tool for model selection. However, model orders with the highest marginal likelihood for NMF and sNMF differ when \SMC and VB are compared, and VB is seen to favor models of lower orders.

The results for $a = 0.001$ demonstrate a different pattern. Here, the symmetric model is preferred by \SMC above the non-symmetric model for almost all model orders. Analysis by \citet{Steck02onthe} show that this behavior is expected regardless of the specific inference procedure used: When the priors are very weak, model selection procedures favor models with fewer parameters. Thus, at very weak prior regime the symmetric model is favored over the non-symmetric model since it has fewer number of parameters (due to parameter tying). The distinction is not as clear with VB, however, where the ELBOs for the sNMF is not consistently above the ones for NMF. Similar behavior by VB at weak prior settings was also observed in the experiments with synthetic data at Subsection \ref{subsec:synth}.


\subsection{Examining Decompositions Obtained by \SMC} \label{subsec:decomposition}
In our final experiment, we compare the nature of decompositions obtained by \SMC for BAM with those obtained by an optimization algorithm, that are popular among practitioners. Such an algorithm is alternating least squares using projected gradient descent (ALS-PGD) \citep{Lin2007}. We use the bigram letter transition data presented above for this task, with the number of tokens being set to $2000$ as above. We evaluate the decompositions produced through this experiment according to two criteria: sparsity and accuracy. For sparsity we use \citeauthor{Hoyer2004}'s (\citeyear{Hoyer2004}) definition:
\[
\text{sparsity}(X) = \frac{\sqrt{n} - (\sum |X_i|)/\sqrt{\sum X_i^2}}{\sqrt{n}-1}
\]
where $n$ is the number of elements in the matrix. Given that NMF decomposition results in two different matrices, the mean of the two sparsity results were used as the ultimate score for sparsity for a given decomposition. For accuracy we use two divergence measures between the original matrix $X$ and its approximation $\hat{X}$: the squared loss, that is the Frobenius norm of the difference between the two matrices, and KL-divergence, defined as below:

$$
D_{KL}(X \mid \mid \hat{X}) = \sum_{ij} \big(X_{ij}\log\frac{X_{ij}}{\hat{X}_{ij}} - X_{ij} +\hat{X}_{ij} \big)
$$
We let $i$ and $j$ be the indices of the observed matrix, and $k$ be the index of the latent variable. To obtain the decomposition of the data using BAM, we set $W= \theta_{i|k}$ and $H= T \theta_{j|k} \theta_{k}$ with $T$ being the total number of tokens, and $WH \approx X$.

\begin{figure}[t!]
	\centering
	\begin{subfigure}[t]{0.32\textwidth}
		\includegraphics[width=1.\textwidth]{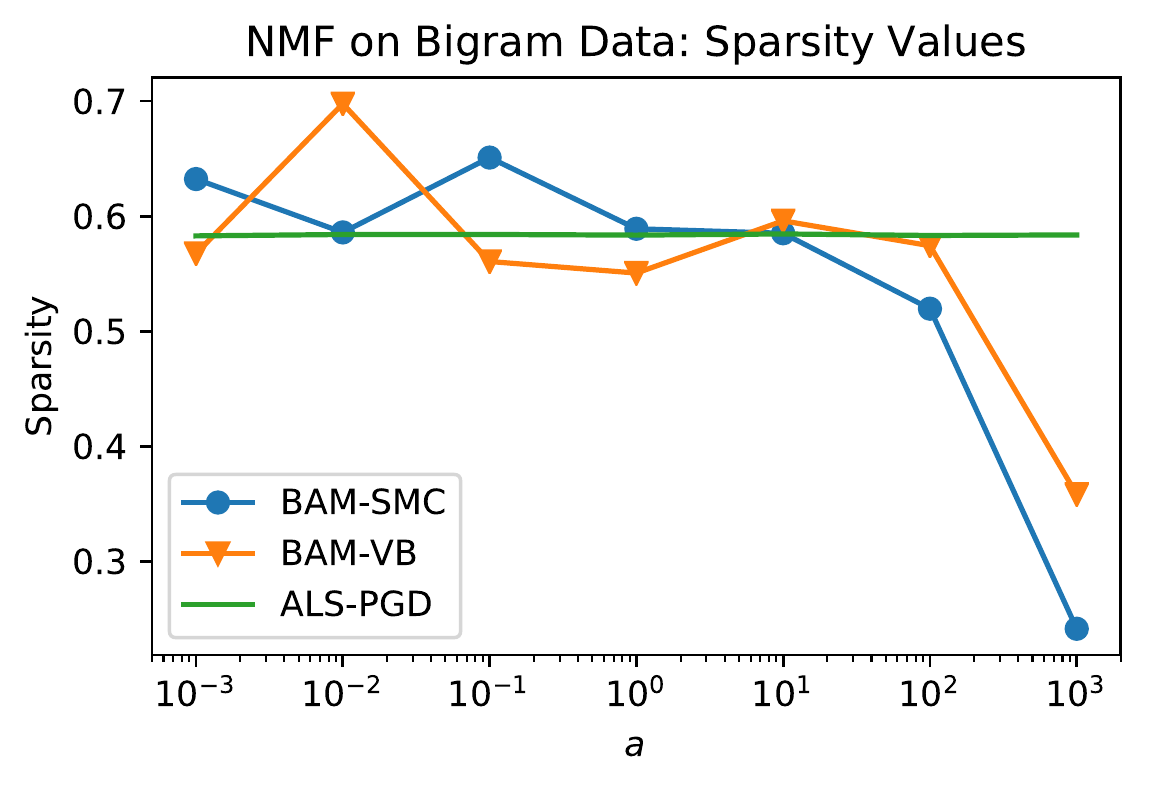}
	\end{subfigure}	
	\begin{subfigure}[t]{0.32\textwidth}
		\includegraphics[width=1.\textwidth]{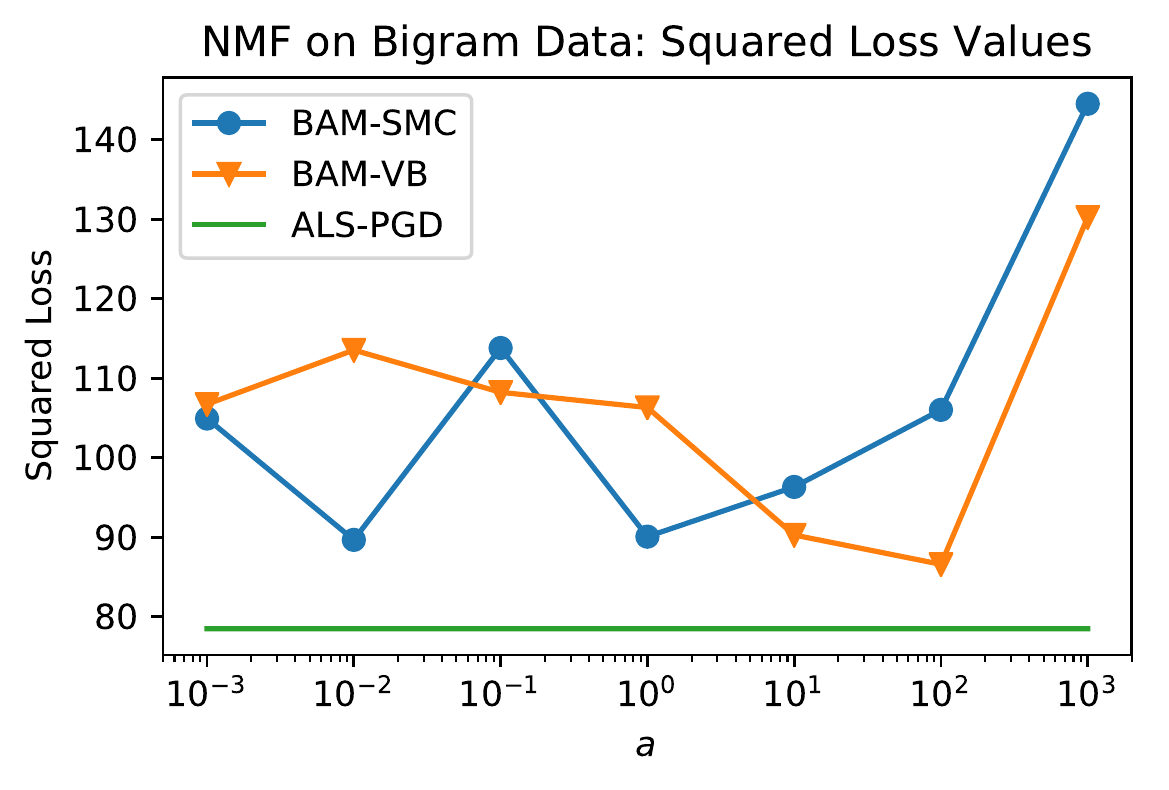}
	\end{subfigure}	
	\begin{subfigure}[t]{0.32\textwidth}
		\includegraphics[width=1.\textwidth]{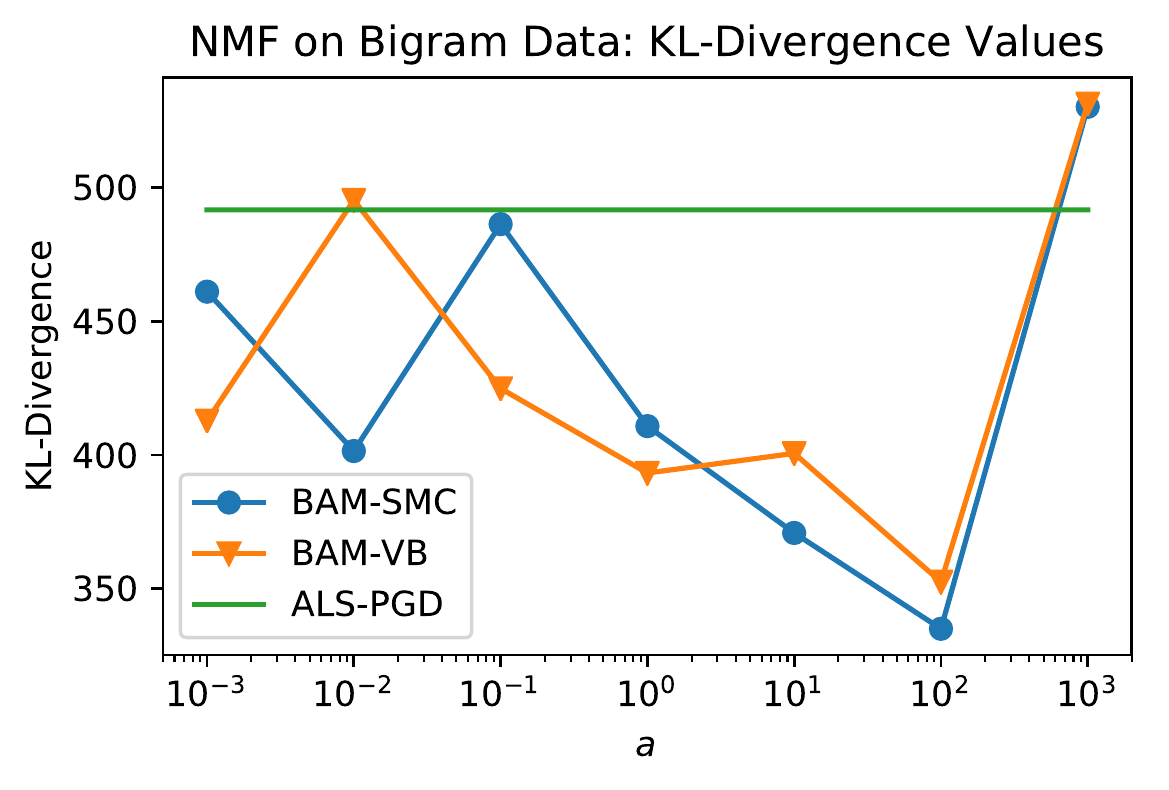}
	\end{subfigure}	
	\caption{Sparsity (left), squared loss (center), and KL-divergence (right) values for hyperparameters $a = 10^{-3}$ to $a = 10^{3}$ with $R = 3$.}
	\label{fig:a_range_decomp}
\end{figure}

We have compared the representations produced by the two algorithms fixing a model order ($R=3$) and varying $a$ to see how composition quality of the two methods compare at different hyperparameter strength levels. The results for this can be seen in Figure \ref{fig:a_range_decomp}. The results show that for a considerable range of hyperparameter values, decompositions produced by SMC for BAM have larger sparsity than those produced by ALS-PGD, while the results for VB are less consistent. Not surprisingly, while the ALS-PGD performs better for squared loss metric, methods of BAM mostly dominate for KL-divergence. This is unsurprising given the fact that these metrics are what the two different frameworks optimize for respectively. 

\section{Discussion and Conclusions} \label{sec: Discussion and Conclusions}
The central message of this paper is that the problem of calculating the marginal likelihood in nonnegative tensor models and topic models are closely related to scoring Bayesian networks and this problem can be treated as a sequential rare event estimation problem. We introduced BAM to explicitly exploit this connection and derive an SMC algorithm. 

The BAM also provides us to see the equivalence of general nonnegative tensor factorizations, topic models, and graphical models. Similar equivalences in specific cases such as the LDA and Bayesian KL-NMF has been pointed out several times \citep{ Buntine2002, Girolami2003, Gaussier2005, Ding2008, Faleiros2016}, but the emphasis is given on algorithmic equivalence rather than the equivalence of the generative models, with the exception of \citet{Buntine2006}. Our approach shows that the equivalence of generative models holds for a large class of decomposable tensor models and discrete graphical models \citep{Heckerman1995, Geiger2002, Geiger_and_Heckerman_1995}. 

Perhaps surprisingly, the literature on calculating the marginal likelihood for graphical models with hidden nodes is not extensive \citep{Friedman2003, beal2006variational, Riggelsen2006, Adel2017}. The graphical model literature focuses typically on structure learning on fully observed models \citep{Tarantola2004} rather than dealing with latent variables. In this paper, we have investigated the problem using a \Polya urn interpretation of the marginal allocation probability of BAM, a model that can be viewed as a \Polya tree \citep{Mauldin_et_al_1992} albeit with tied parameters. This observation provided us a novel yet intuitive sequential inference framework, particularly practical in the data sparse regime where the calculation of the marginal likelihood is simply the probability that the \Polya urn hits a small set, that is the set of all tensors with the observed marginal.

It is important to note that for parameter estimation in topic models and discrete graphical models with latent variables (such as Hidden Markov models or mixture models), alternative tensor methods are also popular, known as the method of moments or spectral methods. Here, the key idea is representing certain higher order moments of the observed data as a tensor and calculating a decomposition to identify directly the hidden parameters. Under certain conditions the exact moment tensors can be shown to be orthogonally decomposable, 
and iterative algorithms with recovery guarantees are known \citep{Anandkumar:2014:TDL:2627435.2697055, Robeva2016}. 
However, such spectral algorithms may not be very suitable for the sparse data regime where the estimates of higher order moments can have a high variance. In this regime, one can obtain invalid (such as negative) estimates as spectral methods minimize implicitly the Frobenius norm, which is not the natural divergence measure for a Poisson likelihood. To our knowledge, a spectral estimate that can be used as a proxy for the marginal likelihood is also not known. 

Our approach can be used for developing algorithms for model selection in other structured topic models and NTF, tensor trains or tensor networks as all these related problems can be viewed as instances of scoring Bayesian networks with latent variables. 

The proposed dynamical process view of tensor factorization turns out to be also useful for understanding the nature of the nonnegative decomposition problem and explain the surprising success of variational approximation methods or Monte Carlo sampling schemata on some data sets and failure on others. From the algorithmic perspective, many asymptotic consistency results for variational inference require that the data size goes to infinity but for tensors models this statement becomes ambiguous for a tensor with a fixed dimension. Our model gives an alternative perspective as the number of tokens: we can view any nonnegative tensor as a limit of a count process where each entry is normalized by the total count. This perspective gives additional justification for variational algorithms in the large data regime when combined with the results presented in \citet{Wang2018}. 
From the modeling perspective, it is long known as an empirical fact, that the factor tensors obtained by decomposition of nonnegative matrices and tensors is sparse, informally known as a clustering behaviour or by parts representation. Indeed, this was one of the initial motivations of the seminal paper \citet{Lee2001}. While some theoretical justifications have been provided \citep{Donoho_and_Stodden_2004} we believe that the observed clustering properties can also be understood as a rich-get-richer phenomenon as a consequence of the self reinforcement property of the underlying \Polya urn process.

In our development, we also point out a subtle issue that seems to have been neglected in the topic modeling and probabilistic NTF literature when choosing priors. We illustrate that the seemingly intuitive and convenient choice of independent Dirichlet priors on factor parameters turns out having a quite dramatic effect on posterior inference and may even lead to possibly misleading conclusions about the model structure, such as the cardinality of a hidden node. This concept of model structure is also closely related to a central problem in tensor computations such as when inferring the rank of a decomposition. The equivalence of NTF and graphical models suggests us that the Dirichlet choice is inevitable under certain assumptions \citep{Geiger_and_Heckerman_1995} while for structure learning the Dirichlet hyperparameters should be chosen consistently as marginal pseudo-counts from a fixed, common imaginary data set in contrast of being merely independent \emph{a priori} \citep{Heckerman1995}.

A possible future work is developing efficient algorithms where the observations topic modeling algorithms are designed for handling large document corpora and missing data is typically not a concern -- absence of certain words from a document collection is an indicator about a topic, and a zero count is an informative observation. Missing data here corresponds to censoring, where  certain words are deliberately erased from certain documents. This is not a typical scenario in document processing and some inference algorithms, such as collapsed Gibbs sampling \citep{Teh2007}, do not handle this type of missing data naturally. This is in contrast to matrix and tensor factorization models, where one can naturally handle such constraints by simply removing the corresponding terms from the marginal likelihood. 

In the topic modeling literature, there are several inference algorithms that have been developed for fast inference specifically for LDA \citep{Li2014,Yu2015,Magnusson2015,Terenin2017}. It is not always clear in the literature how to extend and apply these powerful inference methods to more general structured topic models and tensor factorizations. In a sense, our paper outlines a generic method for designing algorithms for tensors.

\appendix

\section*{Acknowledgements}
The work was partly supported as a bilateral Turkish-French research programme by the French National Research Agency under Grant ANR-16-CE23-0014 (FBIMATRIX) and by TUBITAK Turkish Science, Technology and Research Foundation under Grant 116E580 (FBIMATRIX).

\bibliographystyle{bib/abbrvnat.bst}
\bibliography{bib/bam_references}

\newpage
\begin{appendices}

\section{Source Code}

Code is available at  \url{https://github.com/atcemgil/bam}.

\section{Basic Distributions}

\subsection{Poisson Distribution}

Gamma function: $\Gamma(z) = (z-1)!$ for nonnegative integer $z$.
$$
\mathcal{PO}(s; \lambda) = \exp(s\log \lambda -\lambda -\log \Gamma(s+1))
$$
$\lambda$ is the intensity parameter.
$$
\E{s} = \lambda
$$
The entropy is not exactly known but can be approximated. 
$$
\E{-\log \mathcal{PO}(s; \lambda)} = -\lambda \log \lambda +\lambda + \E{\log \Gamma(s+1)}
$$
\subsection{Gamma Distribution}

$$
\mathcal{GA}(\lambda; a,b) = \exp((a-1)\log\lambda - b\lambda - \log\Gamma(a) + a\log b)
$$
$a$ is the shape and $b$ is the rate parameter.
Sufficient statistics
$$
\E{\lambda} = a/b, \quad \E{\log \lambda} = \psi(a) - \log(b)
$$
The entropy:
\begin{align*}
\E{-\log \mathcal{GA}(\lambda; a,b)} & =  -(a-1)\E{\log\lambda} + b\E{\lambda} + \log\Gamma(a) - a\log b \\
& = -(a-1)\psi(a) - \log(b) + \log\Gamma(a)  + a
\end{align*}
Here, $\psi(x)$ is the psi function defined as $\psi(x) = \log\Gamma(x)/dx$.

\subsection{Dirichlet Distribution}

Multivariate Beta function:
$$
B(\alpha) = \frac{\prod_n \Gamma(\alpha_n)}{\Gamma(\sum_n \alpha_n)}
$$
Dirichlet density:
$$
\mathcal{D}(\theta; \alpha) = \frac{1}{B(\alpha)} \prod_n \theta_n^{\alpha_n - 1}
$$
$$
\E{\theta_n} = \frac{\alpha_n}{\sum_{m} \alpha_m}, \quad \E{\log \theta_n} = \psi(\alpha_n) - \psi(\sum_m \alpha_m)
$$

\subsection{Categorical and Multinomial Distributions}

Multinomial distribution:
$$
\mathcal{M}(s; \theta, N) = \binom{N}{s_1,\dots, s_K} \prod_{k=1}^K \theta_k^{s_k}
$$
subject to $s_1 + \cdots + s_K = N$.

$$
\E{s_{1:K}} = N \theta_{1:K}
$$
Categorical distribution:
$$
\mathcal{M}(s; \theta, 1) = \prod_{k=1}^K \theta_k^{s_k}
$$
subject to $s_1 + \cdots + s_K = 1$.

\section{Allocation Model}

Let $\Theta \equiv \{\theta_{n|\pa{n}}(:,i_\pa{n}) : \forall(n,i_\pa{n}) \}$ be the set of conditional probability tables implied by a Bayesian Network $\mathcal{G}$ where each $\theta_{n \mid \pa{n}}$ is Dirichlet distributed:
\begin{align}
\theta_{n|\pa{n}}(:, i_{\pa{n}}) & \sim  \mathcal{D}(\alpha_{\fa{n}}(:, i_{\pa{n}})) & & \forall n \in [N], \forall i_{\pa{n}}
\end{align}

An allocation tensor $S$ is constructed via independently allocating the events, that are generated from a Poisson process with the intensity $\lambda$, by using the thinning probabilities $\Theta$:
\begin{align*}
\lambda & \sim  \mathcal{GA}(a, b) & 
\mathbf{S}({i_{1:N}})  & \sim  \mathcal{PO}\left(\lambda \prod_{n=1}^N \theta_{n|\pa{n}}(i_n, i_{\pa{n}})\right)
\end{align*}
Full joint distribution of $S$, $\Theta$ and $\lambda$ is
\begin{eqnarray*}
\pi(\lambda, \Theta, S) & = & \exp\left( (a-1)\log\lambda - b\lambda - \log\Gamma(a) + a\log b \right) \\
& & \prod_n \exp\left\{ \sum_{i_\pa{n}} \log \Gamma( \sum_{i_n} \alpha_{\fa{n}} ) -  \sum_{i_\fa{n}} \log \Gamma( \alpha_{\fa{n}} )\right\} \\
& & \prod_n \exp \left\{ \sum_{i_\fa{n}} (\alpha_{\fa{n}} - 1) \log \theta_{n|\pa{n}} \right\} \\
& & \prod_{i_{1:N}} \exp\left\{ S({i_{1:N}}) \bigl( \log \lambda + \sum_n \log \theta_{n|\pa{n}} \bigr) - \lambda \prod_{n=1}^N \theta_{n|\pa{n}} - \log\Gamma(S({i_{1:N}}) + 1) \right\} \nonumber \\
& = & \exp\left\{ (a-1 + \sum_{i_{1:N}} S({i_{1:N}}) \right\}\log\lambda - (b+1)\lambda - \log\Gamma(a) + a\log b \bigr) \\
& &  \exp \left\{\sum_n \sum_{i_\pa{n}} \log \Gamma( \sum_{i_n} \alpha_{\fa{n}} ) -  \sum_n \sum_{i_\fa{n}}  \log \Gamma( \alpha_{\fa{n}} )\right\} \\
& &  \exp\left\{\sum_n \sum_{i_\fa{n}} (\alpha_{\fa{n}} - 1 +  S_{\fa{n}}) \log \theta_{n|\pa{n}}   \right\} \nonumber \\
& &  \exp\left\{ - \sum_{i_{1:N}} \log\Gamma(S({i_{1:N}}) + 1) \right\} \nonumber \\
\end{eqnarray*}
By completing the terms, we can now integrate out $\lambda$ and $\theta$ 
\begin{eqnarray}
\pi(\lambda, \Theta| S) & = & \mathcal{GA}(\lambda; a + S_+, b+1) 
\prod_n \prod_{i_{\pa{n}}} \mathcal{D}(\alpha_{\fa{n}} + S_{\fa{n}} ) 
\end{eqnarray}
here,
$
S_{\fa{n}}(i_{\fa{n}}) \equiv \sum_{i'_\nfa{n}} S_{i'_{1:N}}
$ and $S_+ \equiv \sum_{i_{1:N}} S({i_{1:N}})$ as above. Note that, surprisingly the posterior remains still factorized as 
$$
\pi(\lambda, \Theta| S) = \pi(\lambda| S) \pi(\Theta| S
)$$ 
which leads to a closed form expression for the marginal likelihood of $S$:
\begin{eqnarray}
\pi(S) & = & 
\frac{\pi(\lambda, \Theta, S)}{\pi(\lambda \mid S) \pi(\Theta \mid S)} \nonumber \\
& = & \exp\left(\log\Gamma(a + S_+) -(a+ S_+)\log(b+1) - \log\Gamma(a) + a\log b  \right) \nonumber \\
& &  \exp\left\{ \sum_n \sum_{i_\pa{n}} \log \Gamma( \sum_{i_n} \alpha_{\fa{n}} ) -  \sum_n \sum_{i_\fa{n}}  \log \Gamma( \alpha_{\fa{n}} )\right\} \nonumber \\
& &  \exp \left\{  -\sum_n \sum_{i_\pa{n}} \log \Gamma( \sum_{i_n} (\alpha_{\fa{n}} + S_{\fa{n}}))) + \sum_n \sum_{i_\fa{n}}  \log \Gamma( \alpha_{\fa{n}} + S_{\fa{n}})) \right\} \nonumber \\
& &  \exp\left\{ - \sum_{i_{1:N}} \log\Gamma(S({i_{1:N}}) + 1) \right\} \nonumber \\
\pi(S) & = & \frac{b^{a}}{(b+1)^{a+ S_+}}\frac{\Gamma(a + S_+)}{\Gamma(a)} \left( \prod_n \frac{B_n( \alpha_{\fa{n}} + S_{\fa{n}} )}{B_n( \alpha_{\fa{n}}  )}  \right) 
\frac{1}{\prod_{i_{1:N}} S({i_{1:N}})!} \label{eq:pS3}
\end{eqnarray}
where, for a tensor $Z$, $B_{n}(Z_{\fa{n}})$ is defined in \eqref{eq: multivariate beta function}.

\section{Variational Bayes}
A mean-field approximation to the allocation model is in the form of
\begin{equation*}
\mathcal{Q}(S, \lambda, \Theta) =  q(S) ~ q(\lambda,\Theta)
\end{equation*}
where the factors are
\begin{eqnarray*}
q(S) & \propto & \exp\bigl(\E[q(\lambda, \Theta)]{\log \pi(\mathbf{S}, \lambda, \Theta) + \log \mathbb{I}(\mathbf{S}_{V} = X)} \bigr) 
\\
q(\lambda, \Theta) & \propto & \exp\bigl(\E[q(S)]{\log \pi(\mathbf{S}, \lambda, \Theta) + \log \mathbb{I}(\mathbf{S}_{V} = X)} \bigr)
\end{eqnarray*}
and explicit evaluation of the equations above implies the following set of marginal variational distributions
\begin{align*}
q(S) & = \prod_{i_\vi} \mathcal{M}\bigl(S(:, i_\vi); X(i_\vi), \Phi_\vi(:, i_\vi) \bigr) \\
q(\lambda, \Theta) & = q( \lambda) \prod_{n = 1}^{N} q(\theta_{n \mid \pa{n}}) 
\end{align*}
where 
\begin{align*}
q(\lambda) & = \mathcal{GA}(\lambda; \hat{a}, b+1) \\
q(\theta_{n \mid \pa{n}}) & = \prod_{i_\pa{n}} \mathcal{D}(\theta_{n \mid \pa{n}}(:,i_\pa{n}); \hat{\alpha}_{\fa{n}}(:,i_\pa{n})) & \forall n \in [N].
\end{align*}
Therefore, $\mathcal{Q}$ is equal to
\begin{eqnarray*}
\mathcal{Q} & = & \exp\left\{ \sum_{i_\vi} \bigl( \log \Gamma(X(i_\vi)+1) - \sum_{i_\vibar} \log \Gamma( S(i_{1:N}) + 1) + \sum_{i_\vibar} {S(i_{1:N})} \log \Phi_{\vi}(i_{1:N})  \bigr) \right\} \\
& & \exp((\hat{a}-1)\log\lambda - (b+1)\lambda - \log\Gamma(\hat{a}) + \hat{a}\log (b+1)) \\
&& \prod_n \exp\left\{\sum_{i_\pa{n}} \log \Gamma\left( \sum_{i_n} \hat{\alpha}_{n|\pa{n}}(i_{n}, i_{\pa{n}}) \right) -  \sum_{i_\pa{n}} \sum_{i_n} \log \Gamma( \hat{\alpha}_{\fa{n}}(i_{n}, i_{\pa{n}}) )\right\} \\
& & \prod_n \exp \left\{\sum_{i_\pa{n}} \sum_{i_n} (\hat{\alpha}_{\fa{n}}(i_{n}, i_{\pa{n}}) - 1) \log \theta_{n|\pa{n}}(i_{n}, i_{\pa{n}}) \right\} 
\end{eqnarray*}
The evidence lower bound is a functional of $\mathcal{Q}$ which can be derived from the $\textup{KL}(\mathcal{Q} \| \mathcal{P})$:
\begin{align}
\mathcal{B}_{\mathcal{P}}[\mathcal{Q}] &\equiv \E[\mathcal{Q}]{\log \{\pi(\mathbf{S}, \lambda, \Theta) ~ \mathbb{I}(\mathbf{S}_{V} = X)} \} - \E[\mathcal{Q}]{\log \mathcal{Q}(\mathbf{S}, \lambda, \Theta)} \nonumber \\
&= F[\mathcal{Q}] + H[\mathcal{Q}]
\label{eq:elbo_terms}
\end{align}
Minimization of $\textup{KL}(\mathcal{Q} \| \mathcal{P})$ w.r.t.\ $\mathcal{Q}$ ends up with the following variational marginal distributions:
\begin{itemize}
\item $q(\lambda) =  \mathcal{GA}(\lambda; \hat{a} , b+1)$ with the variational parameter 
\begin{eqnarray*}
\hat{a} & = & a + \E[\mathcal{Q}]{\mathbf{S}_+}
\end{eqnarray*}
Hence, $\E[\mathcal{Q}]{\log \lambda} = \psi(\hat{a}) - \log(b+1) = \psi(a + \E[\mathcal{Q}]{S_+})) - \log(b+1)$ and $\E[\mathcal{Q}]{\lambda} = \hat{a}/(b+1)$.
\\
\item $q(\theta_{n|\pa{n}}) =  \mathcal{D}(\theta_{\fa{n}}; \hat{\alpha}_{n |\pa{n}})$ with the variational parameter
\begin{eqnarray*}
\hat{\alpha}_{\fa{n}} & = & \alpha_{\fa{n}} + \E[\mathcal{Q}]{\mathbf{S}_{\fa{n}}}
\end{eqnarray*}
Hence, $\E[\mathcal{Q}]{\log \theta_{n|\pa{n}}(i_n, i_{\pa{n}})} = \psi(\hat{\alpha}_{\fa{n}}(i_\fa{n})) - \psi(\hat{\alpha}_{\pa{n}}(i_\pa{n}))$
\\
\item $q(S) = \prod_{i_\vi} \mathcal{M}(S(:, i_\vi); \Phi_{\vi}(:, i_\vi), X(i_\vi))$ with the variational parameter
\begin{eqnarray*}
\Phi_{\vi}(i_{1:N}) & = & \frac{\exp\left(\E[\mathcal{Q}]{\log \lambda} + \sum_n \E[\mathcal{Q}]{\log \theta_{n|\pa{n}}} \right)}{\sum_{i_\vibar} \exp\left(\E[\mathcal{Q}]{\log \lambda} + \sum_n \E[\mathcal{Q}]{\log \theta_{n|\pa{n}}} \right)} \\
& = & \frac{\exp\left(\sum_n \E[\mathcal{Q}]{\log \theta_{n|\pa{n}}} \right)}{\sum_{i_\vibar} \exp\left(\sum_n \E[\mathcal{Q}]{\log \theta_{n|\pa{n}}} \right)} 
\end{eqnarray*}
subject to $X = S_\vi$. Conditioned on $X$, 
the required expectations $\E[\mathcal{Q}]{S}$ can be computed in closed form, as $q(S)$ is a product of multinomial probabilities:
$$
\E[\mathcal{Q}]{\mathbf{S}(i_{1:N})} = X(i_\vi) \Phi(i_{1:N})
$$
\end{itemize}

\subsection{Derivation of ELBO}
The individual terms $F[\mathcal{Q}]$ amd $H[\mathcal{Q}]$ in \eqref{eq:elbo_terms} are
\begin{eqnarray*}
F[\mathcal{Q}] & \equiv & \E[\mathcal{Q}]{\log {\pi}(\mathbf{S}, \lambda, \Theta)} \\
& = &  (a-1 + \E[\mathcal{Q}]{\mathbf{S}_+})\E{\log\lambda} - (b+1)\E[\mathcal{Q}]{\lambda} - \log\Gamma(a) + a\log b  \\
& &  +\sum_n \sum_{i_\pa{n}} \log \Gamma(\alpha_{\pa{n}}(i_{\pa{n}}) ) -  \sum_n \sum_{i_\fa{n}} \log \Gamma( \alpha_{\fa{n}} (i_{\fa{n}})) \\
& &  +\sum_n \sum_{i_\fa{n}} (\alpha_{\fa{n}} - 1 +  \E[\mathcal{Q}]{\mathbf{S}_{\fa{n}}(i_{\fa{n}})}) \E[\mathcal{Q}]{\log \theta_{n|\pa{n}}(i_{n}, i_{\pa{n}}) }  \\
& & -\sum_{i_{1:N}} \E[\mathcal{Q}]{\log\Gamma(\mathbf{S}({i_{1:N}}) + 1) } \\
H[\mathcal{Q}] & \equiv & \E[\mathcal{Q}]{\log \mathcal{Q} (\mathbf{S}, \lambda, \Theta) } \\
& = & \sum_{i_{1:N}} \E[\mathcal{Q}]{\log\Gamma( \mathbf{S}(i_{1:N}) + 1)} -\sum_{i_\vi} \log \Gamma(X(i_\vi)+1) -\sum_{i_{1:N}} \E[\mathcal{Q}]{\mathbf{S}(i_{1:N})} \log \Phi_{\vi}(i_{1:N})  \\
& & -(\hat{a}-1)\E[\mathcal{Q}]{\log\lambda} + (b+1)\E[\mathcal{Q}]{\lambda} + \log\Gamma(\hat{a}) - \hat{a}\log (b+1) \\
&& -\sum_n \sum_{i_\pa{n}} \log \Gamma( \hat{\alpha}_{\pa{n}}(i_{\pa{n}}) ) +  \sum_n \sum_{i_\fa{n}} \log \Gamma( \hat{\alpha}_{\fa{n}}(i_{\fa{n}}) ) \\
& & -\sum_n \sum_{i_\fa{n}} (\hat{\alpha}_{\fa{n}}(i_{\fa{n}}) - 1) \E[\mathcal{Q}]{\log \theta_{n|\pa{n}}(i_{n}, i_{\pa{n}})}  
\end{eqnarray*}
Since evidence lower bound is the difference of the terms $F[\mathcal{Q}]$ and $H[\mathcal{Q}]$, simplification of the common terms yields
\begin{eqnarray*}
\mathcal{B}_\mathcal{P}[\mathcal{Q}] & = &  a\log b - (a + \E[\mathcal{Q}]{\mathbf{S}_+})\log (b+1) + \log\Gamma(a + \E[\mathcal{Q}]{\mathbf{S}_+}) - \log\Gamma(a) \\
& &  +\sum_n \sum_{i_\pa{n}} \log \Gamma( \alpha_{\pa{n}} ) - \sum_n \sum_{i_\fa{n}}  \log \Gamma( \alpha_{\fa{n}}(i_{\fa{n}}) ) \\
&& -\sum_n \sum_{i_\pa{n}} \log \Gamma( \alpha_{\pa{n}}(i_{\pa{n}}) + \E[\mathcal{Q}]{\mathbf{S}_{\pa{n}}}(i_{\pa{n}}) ) \\
&& + \sum_n \sum_{i_\fa{n}} \log \Gamma( \alpha_{\fa{n}}(i_{\fa{n}}) + \E[\mathcal{Q}]{\mathbf{S}_{\fa{n}}}(i_{\fa{n}}) ) \\
&  & - \sum_{i_\vi} \log \Gamma(X(i_\vi)+1)  - \sum_{i_{1:N}} \E[\mathcal{Q}]{\mathbf{S}(i_{1:N})} \log \Phi_{\vi}(i_{1:N})  \\
e^{\mathcal{B}_\mathcal{P}[\mathcal{Q}]} & = & \frac{b^a}{(b+1)^{a + \E[\mathcal{Q}]{\mathbf{S}_+}}}\frac{\Gamma(a + \E[\mathcal{Q}]{\mathbf{S}_+})}{\Gamma(a)} \\
& &\left(\prod_n \frac{B_n\left(\alpha_{\fa{n}} + \E[\mathcal{Q}]{\mathbf{S}_{\fa{n}}}\right)}{B_n(\alpha_{\fa{n}})} \right) 
 \frac{\prod_{i_{1:N}} \Phi_{\vi}(i_{1:N})^{-\E[\mathcal{Q}]{\mathbf{S}(i_{1:N})}}}{\prod_{i_\vi} \Gamma(X(i_\vi)+1)} 
\end{eqnarray*}

\subsection{Calculation of sufficient statistics}
The variational parameter $\Phi_\vi$ of multinomial distribution $q(S)$ also factorizes according to the directed graph $\mathcal{G}$ and have the form
\begin{eqnarray*}
\Phi_{\vi}(i_{1:N}) & \propto & \prod_n \exp\left(\E[\mathcal{Q}]{\log \theta_{n|\pa{n}}(i_n, i_\pa{n}) } \right) 
\end{eqnarray*}
This structure can be exploited to calculate the expectations by an algorithm closely related to belief propagation.  
\begin{eqnarray*}
\E{\mathbf{S}_{\fa{n}}(i_{\fa{n}})} = \sum_{i_{\nfa{n}}} X(i_\vi) \Phi_{\vi}({i_{1:N}})
\end{eqnarray*}

Belief propagation is typically framed as carrying out inference in a Bayesian network by conditioning some of the random variables to their observed values. In our allocation model formalism, this is equivalent to observing
only a single token $\tau$ conditioned on the event that its visible and invisible indices are $i^\tau_\vi$ and $i^\tau_\vibar$ respectively, so the observations can be naturally factorized as
 
\begin{eqnarray*}
s^{\tau} (i_{1:N}) & = & \prod_n \ind{i_n = i^\tau_n} \\
x^\tau(i_\vi) & \equiv & \sum_{i_{\vibar}} s^{\tau}(i_\vi, i_\vibar)
\end{eqnarray*}
A general contraction of the allocation tensor can also be written as  
\begin{eqnarray*}
X(i_\vi) & = & \sum_\tau x^\tau(i_\vi) = \sum_\tau \sum_{i_{\vibar}} \prod_n \ind{i_n = i^\tau_n} \\
& = & \sum_\tau \left(\prod_{n \in V}  \ind{i_n = i^\tau_n} \right) \sum_{i_{\vibar}} \left( \prod_{n\in \vibar} \ind{i_n = i^\tau_n} \right) \\
& = & \sum_\tau \prod_{n \in \vi} \ind{i_n = i^\tau_n}
\end{eqnarray*}
As such, 
\begin{eqnarray*}
\E[\mathcal{Q}]{\mathbf{S}_{\fa{n}}(i_{\fa{n}})} & = & \sum_{i_{\nfa{n}}} \Phi_{\vi}({i_{1:N}}) \sum_\tau \prod_{n \in \vi} \ind{i_n = i^\tau_n}
\end{eqnarray*}
We could run belief propogation for each token $\tau$ separately to estimate the expected sufficient statistics as these are additive
\begin{eqnarray*}
\E[\mathcal{Q}]{\mathbf{S}_{\fa{n}}(i_{\fa{n}})} & = & \sum_\tau  \sum_{i_{\fa{n}}}  \Phi_{\vi}({i_{1:N}}) \prod_{n \in \vi} \ind{i_n = i^\tau_n}
\end{eqnarray*}
However, as the sufficient statistics are only calculated for updating the parameters in bound optimization, this maybe wasteful and it is desirable calculating the statistics in an online or recursive manner.

\section{Parameter tying}

In some applications, it is desirable to constrain the thinning probabilities $\Theta$ further. One common choice is having shared parameters, in order to encode additional structure such as symmetric decompositions $X \approx W W^\top$. Below is the derivation of the SMC algorithm for symmetric CP/PARAFAC model is described. Notice that symmmetric NMF (sNMF) corresponds to the special case of symmetric CP/PARAFAC where $N=2$.

Let $s^{\tau}(r,i_{1:N})$ be the indicator of the event that the token $\tau$ is allocated to cell $(r,i_{1:N})$ and let's define the following indicators in terms of $s^\tau$:

\begin{align*}
s^\tau_0(r) & \equiv \sum_{i_{1:N}} s^\tau(r,i_{1:N}) &
s^\tau_n(i) & \equiv \sum_r \sum_{i_{1:N}:i_n=i} s^\tau(r,i_{1:N}) \\
S_0(r) & \equiv \sum_{\tau=1}^T s_0^\tau(r) &
S_n(i) & \equiv \sum_{\tau=1}^T s_n^\tau(i)
\end{align*}
where the sum $\sum_{i_{1:N}:i_n=i}$ is over all the valuations of indices $i_{1:N}$ satisfying the condition $i_n = i$. Then the conditional probability of the event $s^\tau$ is
\begin{eqnarray*}
\pi(s^{\tau} \mid \Theta) & = & \prod_{r} \theta_r^{s_0^\tau(r)} 
\prod_{n=1}^N \prod_{i_{1:N}} \theta_{i_{1:N} \mid r}^{s^\tau_n (i_{1:N},r)}
= \prod_{r} \theta_r^{s_0^\tau(r)} \prod_{i_{1:N}} \theta_{i_{1:N} \mid r}^{\sum_{n=1}^N s^\tau_n (i_{1:N},r)} \\
\end{eqnarray*}
Since the individual events $s^1, s^2, \dots, s^T$ are conditionally independent given the parameters $\Theta$, the joint distribution of the events $s^{1:T}$ factorizes as follows
\begin{eqnarray*}
\pi(s^{1:T} \mid \Theta) & = & \prod_{\tau=1}^T \pi(s^\tau \mid \Theta) \\
& = & \prod_{\tau=1}^T \prod_{r} \theta_r^{s_0^\tau(r)} 
\prod_{i_{1:N}} \theta_{i_{1:N} \mid r}^{\sum_{n=1}^N s^\tau_n (i_{1:N},r)} \\
& = & \prod_{r} \theta_r^{ \sum_{\tau=1}^T s_0^\tau(r)} 
\prod_{i_{1:N}} \theta_{i_{1:N} \mid r}^{\sum_{n=1}^N \sum_{\tau=1}^T s^\tau_n (i,r)} \\
& = & \prod_{r} \theta_r^{ S_0(r)} 
\prod_{i_{1:N}} \theta_{i_{1:N} \mid r}^{\sum_{n=1}^N S_n (i_{1:N},r)} \\
\end{eqnarray*}
If we assume the following Dirichlet priors on $\Theta$
\begin{align*}
\theta_{:} & \sim \mathcal{D}(\alpha_0) & \theta_{: \mid r} & \sim \mathcal{D}(\alpha(:,r))
\end{align*}
posterior distribution of $\Theta$ turns out to be also Dirichlet due to conjugacy:
\begin{eqnarray*}
\pi(\Theta \mid s^{1:T}) 
& \propto &  
\pi(s^{1:T} \mid \Theta) ~ \pi(\Theta) \\
& \propto &
\Bigl( \prod_{r} \theta_r^{ S_0(r)} 
\prod_{i_{1:N}} \theta_{i_{1:N} \mid r}^{\sum_{n=1}^N S_n (i_{1:N},r)} \Bigr)
\Bigl( \prod_r \theta_r^{\alpha_0(r) - 1} \Bigr) 
\Bigl( \prod_{r} \prod_{i_{1:N}} \theta_{i_{1:N} \mid r} ^{\alpha(i_{1:N},r) - 1} \Bigr) \\
& = &
\Bigl( \prod_{r} \theta_r^{ S_0(r)} 
\prod_{i_{1:N}} \theta_{i_{1:N} \mid r}^{\sum_{n=1}^N S_n (i_{1:N},r)} \Bigr)
\Bigl( \prod_{r} \theta_r^{\alpha_0(r) - 1} \prod_{i_{1:N}} \theta_{i_{1:N} \mid r} ^{\alpha(i_{1:N},r) - 1} \Bigr) \\
& = &
\prod_{r} \theta_r^{\alpha_0(r) + S_0(r) - 1} \prod_{i_{1:N}} \theta_{i_{1:N} \mid r} ^{\alpha(i_{1:N},r) + \sum_{n=1}^N S_n(i_{1:N},r) - 1} \\
& \propto &
\mathcal{D}\bigl(\theta_{:}; \alpha_0 + S_0 \bigr) ~ \prod_r \mathcal{D}\bigl(\theta_{: \mid r}; \alpha(:,r) + \sum_{n=1}^N S_n(:,r) \bigr)
\end{eqnarray*}
Hence, the marginal likelihood of the events $s^{1:T}$ can be found by Bayes theorem:
\begin{eqnarray*}
\pi(s^{1:T}) & = & \frac{\pi(s^{1:T} \mid \Theta) ~ \pi(\Theta)}{\pi( \Theta \mid s^{1:T})} \\
& = & 
\frac{B\bigl(\alpha_0  + S_0 \bigr)}{B\bigl(\alpha_0 \bigr)}
\prod_r \frac{B\bigl(\alpha (:,r) + \sum_n S_n (:,r) \bigr)}{B\bigl(\alpha (:,r)\bigr)}
\end{eqnarray*}
and similarly the \PolyaBayes process probabilities are
\begin{eqnarray*}
\pi(s^\tau \mid S^{\tau-1}) = \pi(s^\tau \mid s^{1:\tau-1}) & = & \frac{\pi(s^{1:\tau})}{ \pi(s^{1:\tau-1})}
\end{eqnarray*}
Then it is straightforward to adapt Algorithm \ref{alg:sisR} to symmetric CP/PARAFAC case by changing only the distributions $p_{\tau}(c_{\bar{V}} \mid c_V, S^{\tau-1})$ and $p_{\tau, \vi}(c_\vi \mid S^{\tau-1})$ as follows
\begin{eqnarray*}
p_{\tau}(r \mid i_{1:N}, S^{\tau-1})
& = & \frac{\sum_{s^\tau} \pi(s^\tau \mid S^{\tau-1}) \ind{s^\tau(r,i_{1:N}) = 1}}{\sum_{s^\tau} \pi(s^\tau \mid S^{\tau-1})\ind{ \sum_{r'} s^\tau(r',i_{1:N}) = 1}} \\
p_{\tau, \vi}(i_{1:N} \mid S^{\tau-1})
& = & {\sum_{s^\tau} \pi(s^\tau \mid S^{\tau-1})\ind{ \sum_{r'} s^\tau(r',i_{1:N}) = 1}}
\end{eqnarray*}

\end{appendices}

\end{document}